%% file: main.tex
\documentclass[11pt]{article}

\usepackage[preprint]{acl}

\usepackage{times}
\usepackage{latexsym}

\usepackage[T1]{fontenc}

\usepackage[utf8]{inputenc}

\usepackage{microtype}

\usepackage{inconsolata}

\usepackage{graphicx}
\usepackage{amsmath}

\usepackage{enumerate}
\usepackage{multirow}
\usepackage{amsthm} 
\usepackage{xspace}
\usepackage{xcolor}
\usepackage{enumitem}
\usepackage{booktabs}
\usepackage[ruled,vlined]{algorithm2e}
\usepackage{tcolorbox}
\usepackage[dvipsnames]{xcolor}
\usepackage{pifont}
\usepackage{colortbl}

\title{Revitalizing Black-Box Interpretability: \\Actionable Interpretability for LLMs via Proxy Models}

\author{Junhao Liu\textsuperscript{1}, Haonan Yu\textsuperscript{2}, Zhenyu Yan\textsuperscript{3}, Xin Zhang\textsuperscript{4} \thanks{Corresponding author.}\\
        Key Lab of High Confidence Software Technologies (Peking University), Ministry of Education\\
    School of Computer Science, Peking University, Beijing, China\\
    \texttt{\{liujunhao\textsuperscript{1}, xin\textsuperscript{4}\}@pku.edu.cn, \{haonanyu\textsuperscript{2}, zhenyuyan\textsuperscript{3}\}@stu.pku.edu.cn}\\
    }
\input{macro.tex}

\input{math_commands.tex}

\makeatletter
\newcommand\blfootnote[1]{%
  \begingroup\renewcommand\thefootnote{}\footnote{#1}%
  \addtocounter{footnote}{-1}\endgroup}
\newcommand{\acceptancenotice}{%
  \ifacl@finalcopy\ifacl@pagenumbers
    \blfootnote{Accepted to ACL 2026 Main Conference.}%
  \fi\fi}
\makeatother

\begin{document}
\maketitle
\acceptancenotice
\input{abstract.tex}

\input{chapters/1.introduction.tex}

\input{chapters/2.pre.tex}
\input{chapters/3.framework.tex}
\input{chapters/4.result.tex}
\input{chapters/4.5.incontext.tex}

\input{chapters/7.conclusion.tex}
\input{chapters/6.limitation.tex}

\section*{Ethics Statement}
While our framework aims to democratize LLM interpretability and enhance model transparency, we acknowledge the dual-use nature of local explanations. Such methods could potentially be misused to facilitate adversarial attacks or generate misleading interpretations of model behavior. We strongly encourage responsible application of our framework and dataset, and emphasize the necessity of robust safeguards when deploying explanation tools in sensitive or high-stakes domains.

\section*{Acknowledgement}

This work was sponsored by the National Natural Science Foundation of China (NSFC) under Grant No. W2411051.

\bibliography{bmsm}

\appendix


\input{chapters/appendix.tex}

\end{document}

%% file: macro.tex
\usepackage{cleveref}

%% file: math_commands.tex
\usepackage{amsmath,amsfonts,bm}

\def\eqref#1{equation~\ref{#1}}

\def\1{\bm{1}}

\def\va{{\bm{a}}}

\def\vx{{\bm{x}}}

\def\vz{{\bm{z}}}

\DeclareMathAlphabet{\mathsfit}{\encodingdefault}{\sfdefault}{m}{sl}
\SetMathAlphabet{\mathsfit}{bold}{\encodingdefault}{\sfdefault}{bx}{n}

\def\gV{{\mathcal{V}}}

\def\sD{{\mathbb{D}}}

\def\sR{{\mathbb{R}}}

\def\sZ{{\mathbb{Z}}}

\newcommand{\E}{\mathbb{E}}



%% file: abstract.tex
\begin{abstract}
Post-hoc explanations provide the transparency and are essential for guiding model optimization, such as prompt engineering and data sanitation.
However, applying model-agnostic techniques to Large Language Models (LLMs) is hindered by prohibitive computational costs, rendering these tools dormant for real-world applications.
To revitalize model-agnostic interpretability, we propose a budget-friendly proxy framework that leverages efficient models to approximate the decision boundaries of expensive LLMs.
We introduce a screen-and-apply mechanism to statistically verify local alignment before deployment.
Our empirical evaluation confirms that proxy explanations achieve over 90\% fidelity with only 11\% of the oracle's cost.
Building on this foundation, we demonstrate the actionable utility of our framework in prompt compression and poisoned example removal.
Results show that reliable proxy explanations effectively guide optimization, transforming interpretability from a passive observation tool into a scalable primitive for LLM development.
Additionally, we open-source code and datasets to facilitate future research\footnote{The code and datasets are available at https://github.com/outerform/Large-Model-Explanation-Benchmark} .
\end{abstract}

%% file: chapters/1.introduction.tex
\section{Introduction}
\label{sec:intro}
\input{figures/bmsm-example.tex}

Post-hoc explanations are not only passive transparency tools, they are also drivers for active model optimization~\citep{autoprompt, tenney2024interactive}.
%
However, the ubiquity of closed-source models like GPT-4o~\citep{GPT4} and Google Gemini~\citep{geminiteam2024geminifamilyhighlycapable} blocks access to internal representations.
This renders model-agnostic methods the only viable option, but their reliance on massive sampling renders them economically impractical for real-world deployment.
In this work, we propose a budget-friendly proxy framework to revitalize model-agnostic interpretability for LLM.

Let us consider an example of interpretability-guided prompt compression as shown in Figure~\ref{fig:example}.
An LLM user aims to reduce the inference cost by removing redundant few-shot examples from a lengthy prompt template. 
Ideally, feature attribution methods like LIME~\citep{LIME} could identify which examples contribute least to the model's predictions.
Specifically, generating a single LIME explanation typically requires querying the target model with thousands of perturbed samples (e.g., 1,000).
Consequently, to evaluate the effectiveness of each perturbed prompt compression strategy, the user would need to generate explanations on a validation set of multiple examples (e.g., 50), leading to an exorbitant number of model invocations.
As detailed in Table~\ref{tab:llm-pricing}, querying GPT-4o for explanations on a small validation set of 50 examples would require $50,000$ queries costing upwards of \$100.
Besides the high cost itself, this also creates a fundamental utility dilemma: the upfront cost of generating explanations dwarfs the potential savings from the optimization task itself, effectively keeping these powerful tools dormant.

To address this limitation, we propose a budget-friendly proxy explanation framework to generate model-agnostic explanations for expensive LLMs.
\input{figures/bmsm-work.tex}
Our approach is motivated by the homogeneity among LLMs~\citep{Jiang_Chai_Li_Liu_Fok_Dziri_Tsvetkov_Sap_Choi_2025, Quantification_of_Distillation}.
This indicates that smaller, cost-efficient models can also potentially have similar local decision boundaries as their larger counterparts, allowing us to \textit{see the big in the small}.
Specifically, instead of querying the expensive oracle, we generate perturbation-based explanations using local, efficient models (e.g., Open-source Qwen~\citep{qwen25} or LLama~\citep{dubey2024llama3herdmodels} models) as budget-friendly proxies.
Crucially, strictly relying on small models introduces risks of misalignment. To ensure reliability, we introduce a \textbf{screen-and-apply} framework: before applying a proxy explanation, a statistical screening step verifies whether the proxy model's local decision boundary aligns with the target model. 
This acts as a safety valve, ensuring we only deploy budget-friendly explanations when they are faithful to the expensive model's behavior.



We validate our framework through a comprehensive empirical study spanning 12 state-of-the-art LLMs—ranging from open-source models like LLaMA 3.1~\citep{dubey2024llama3herdmodels} and DeepSeek V3~\citep{liu2024deepseekv3} to proprietary giants like GPT-4o—across seven diverse tasks including the MMLU benchmark~\citep{mmlu}.
The results confirm that budget-friendly proxies can serve as faithful surrogates, achieving over 90\% fidelity to oracle explanations while reducing economic costs by 88.2\%.
Building on this foundation, we further demonstrate the actionable utility of our framework on downstream optimization tasks: proxy-guided pruning effectively compresses prompts for GPT-4o without performance loss, and proxy explanations accurately identify poisoned examples without accessing the target model.

\paragraph{Contributions.} The main contributions of this work are summarized as follows:

\begin{itemize}[noitemsep,topsep=0pt,parsep=3pt,partopsep=0pt, leftmargin=10pt]
    \item We propose a novel proxy-based framework to \textbf{revitalize} model-agnostic interpretability for LLMs. By introducing a statistical \textbf{screen-and-apply} mechanism, we enable the reliable use of budget-friendly models to approximate the local decision boundaries of expensive, closed-source LLMs.
    
    \item We conduct an extensive two-stage evaluation across 12 state-of-the-art LLMs. Our results demonstrate that proxy explanations achieve over 90\% fidelity with an 88.2\% cost reduction, and also show their actionable utility in downstream applications. We show that proxy explanations effectively guide prompt compression and poisoned example removal.
    
    \item We release \textbf{XLLM-Bench}, a comprehensive dataset containing perturbation samples and model outputs collected from our study, serving as a foundation for future research on efficient LLM interpretability.
\end{itemize}

%% file: figures/bmsm-example.tex
\begin{figure}[t]
    \centering
    \resizebox{\linewidth}{!}{
    \includegraphics{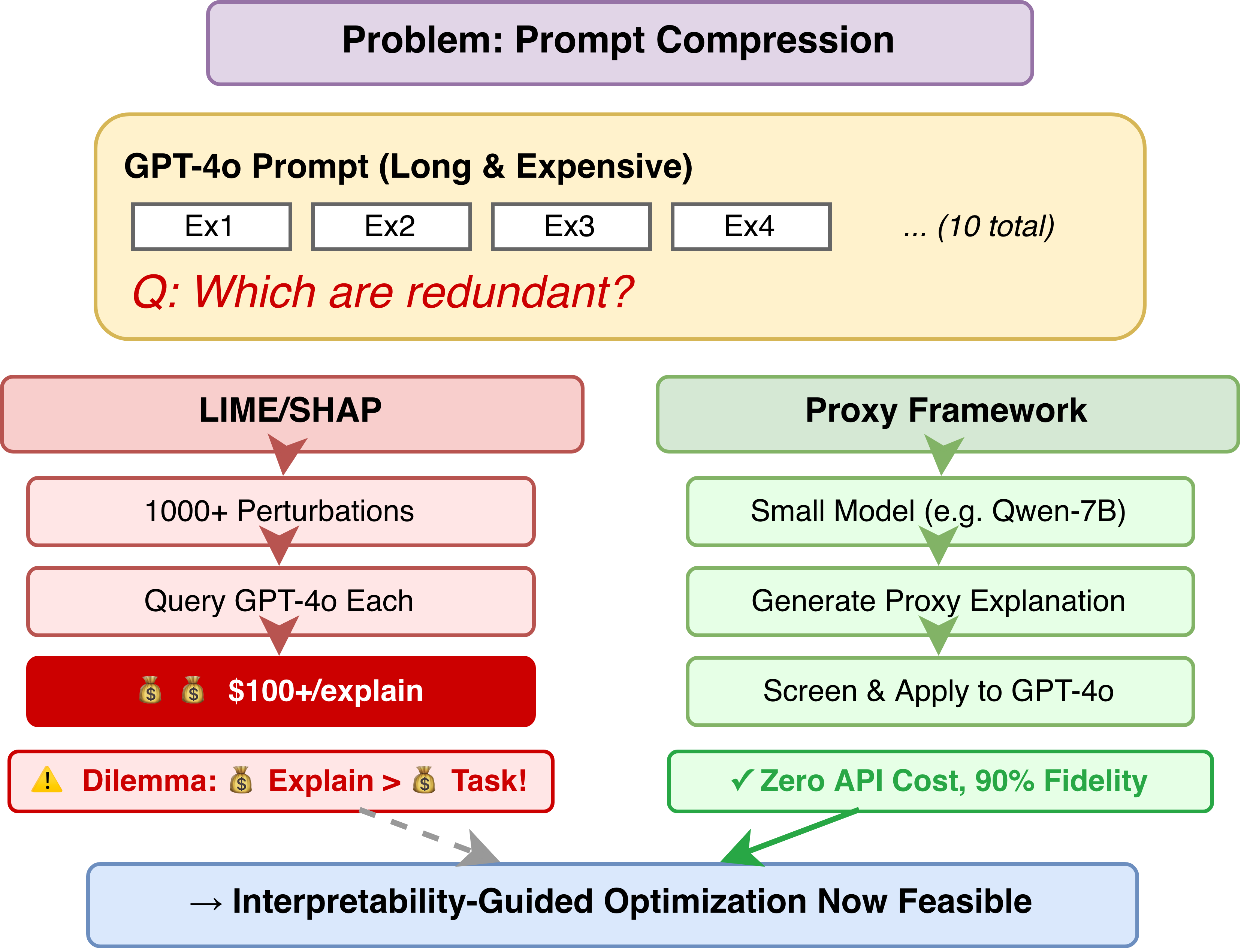}
    }
    \caption{The dilemma of using model-agnostic explanation methods to optimize expensive LLMs.}
    \label{fig:example}
\end{figure} 

%% file: figures/bmsm-work.tex
\begin{figure*}[t]
    \centering
    \includegraphics[scale=0.35]{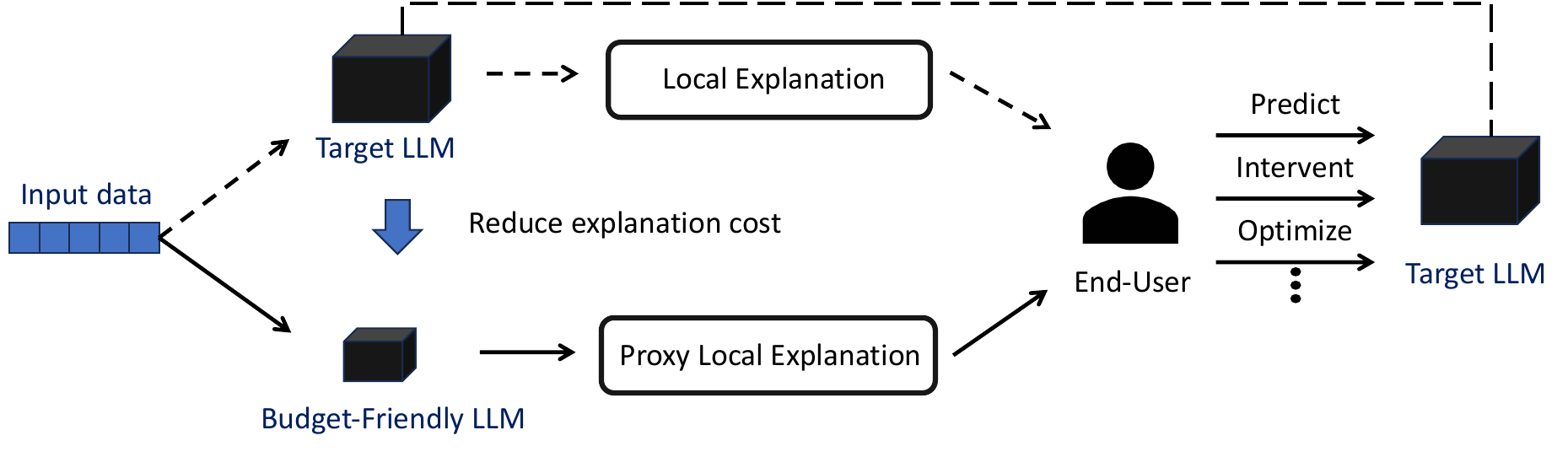}
    \caption{The workflow for leveraging proxy explanations from budget-friendly models to reduce the cost of explaining expensive LLMs. }
    \label{fig:bmsm}
\end{figure*}    

%% file: chapters/2.pre.tex
\section{Background and Related Work}

In this section, we review the background of model-agnostic interpretability and discuss existing limitations that hinder its application for LLM optimization.

\subsection{Large Language Models and Their Homogeneity}
\label{sec:background-llm}

We consider a large language model (LLM) as a probabilistic function \( f \) that maps an input sequence of tokens \( \vx = [x_1, x_2, \dots, x_n] \) to a probability distribution over the vocabulary \( \gV \), denoted as \( f(\vx) \in \sR^{|\gV|} \).
While LLM architectures vary, recent research reveals the homogeneity among state-of-the-art LLMs~\citep{Jiang_Chai_Li_Liu_Fok_Dziri_Tsvetkov_Sap_Choi_2025, Quantification_of_Distillation,Zhou_Xiong_Savarese_Wu, Wright_Masud_Moore_Yadav_Antoniak_Park_Augenstein_2025}, i.e. different LLMs are obtained to behave similarly on similar inputs in many scenarios.
This observation motivated our methodology.

\subsection{Model-Agnostic Explanations}
A local model-agnostic explanation technique \( t \) takes a predictive model \( f \) and an instance \( \vx \) as input, returning an attribution vector \( \va \in \sR^n \) that quantifies the contribution of each input token to the prediction \( f(\vx) \).
Representative methods like LIME~\citep{LIME}, SHAP~\citep{SHAP}, and their variants~\citep{MExGen,enouen-etal-2024-textgenshap,luss2024cell,hackmann2024wordimportanceexplainsprompts, ReX, ConLUX, focuslime, manchor} estimate these attributions by querying \( f \) with a large set of perturbed samples \( \sZ = \{ \vz_1, \dots, \vz_k \} \) sampled from the local neighborhood of \( \vx \).

While effective, Generating a single reliable explanation typically requires thousands of queries. 
For API-based models like GPT-4o, the accumulated cost of these queries renders standard application economically prohibitive.

Several studies have also focused on reducing the cost of generating post-hoc model-agnostic explanations.
As surveyed by~\citet{surveyeffientxai}, some methods amortize explanation generation across inputs by training a unified explainer to approximate the distribution of model explanations~\citep{Covert_Kim_Lee_Zou_Hashimoto_2024, jethani2022fastshap, chuang2023cortxcontrastiveframeworkrealtime, information-theoreticperspective}.
Other approaches remain non-amortized and generate explanations on a per-instance basis, but aim to improve efficiency through various strategies.
These include reducing the number of features in the explanation~\citep{pmlr-v80-chen18j, yoon2018invase, pmlr-v162-wang22b, jullum2021groupshapley}, optimizing the perturbation process~\citep{mitchell2022samplingpermutationsshapleyvalue, dandolo2023acme}, or leveraging global dataset-level information~\citep{yu2025accelerating}.
These methods are orthogonal to our approach and can be integrated with it to further reduce the cost of explanation generation.

\paragraph{Desiderata of Explanations}
\textbf{Fidelity} and \textbf{understandability} are two key desiderata for local explanations aimed at end-users~\citep{survey_XAI,survey_zhang,rojat2021explainable,mahto2025explainable}.

On one hand, explanations should faithfully reflect the model's decision-making process. High fidelity indicates that the explanation accurately captures how the model arrives at its predictions.
On the other hand, explanations should be understandable—that is, they should be presented in a form that humans can easily interpret.

In this paper, we conduct our empirical studies using attribution-based techniques, which produce simple and understandable forms of explanation.
Therefore, we focus our evaluation on the \textbf{fidelity} of proxy explanations generated by budget-friendly models.

\subsection{Explanations for Prompt Optimization}

Beyond transparency, post-hoc explanations are increasingly recognized as powerful drivers for active model optimization, such as prompt debugging or compression.

Some works have explored how explanations can help optimize LLMs. However, existing works either use white-box explanations that require access to model internal representation~\citep{tenney2024interactive,autoprompt}, or only focus on specific task or model settings~\citep{zhao2025leveraging,jiang2023llmlinguacompressingpromptsaccelerated}. This limits their applicability in real-world scenarios. 

In contrast, our method uses proxy model-agnostic explanations as a general and scalable tool to optimize LLMs in a budget-friendly manner.

%% file: chapters/3.framework.tex
\section{Proxy Explanation Framework}

In this section, we introduce the proxy explanations framework.
Our approach is movtivated by the Homogeneity among LLMs as discussed in \ref{sec:background-llm}. While the Homogeneity holds generally, it is not guaranteed for every instance. Blindly applying them introduces risks of misalignment. Thus, we propose a \textbf{screen-and-apply} framework consisting of two key steps: (1) a screening step to determine if proxy explanations can be reliably used for current tasks or instances, and (2) applying the proxy explanations to generate faithful explanations for the target expensive LLMs.

\label{sec:framework}

\subsection{Screening Step}
We use a two-stage screening to ensure proxy explanations from a budget-friendly model \(f'\) are reliable for an expensive LLM \(f\) on a task or dataset with input set \(\sD\) using a local technique \(t\).  For simplicity, we treat both models as deterministic functions mapping an input \(\vx\) to a prediction \(f(\vx)\) or \(f'(\vx)\).
Specifically, the screening procedure includes: an offline task-level screening and an online instance-level screening.
1) The task-level screening is performed once per task. It assesses whether \(f'\) can provide sufficiently faithful proxy explanations for \(f\) over the entire input set \(\sD\), offering a task-level fidelity assessment.
2) The instance-level screening is a lightweight runtime check applied to each input \(\vx\). It verifies whether \(f'\) and \(f\) agree on the prediction for \(\vx\).

\textbf{Task-Level Screening (Offline)\;}
Given a target LLM \(f\), a dataset or task with input set \(\sD\), and an explanation technique \(t\), we run task-level screening once to ensure that proxy explanations from a budget-friendly model \(f'\) are on average sufficiently faithful.
Specifically, we perform statistical hypothesis testing to check whether the proxy explanations from \(f'\) achieve at least a fraction \(\tau\) of the fidelity of oracle explanations from \(f\), with confidence level \(1 - \delta\). Formally, we define the task-level screening decision as a binary function:\(s_{\text{task}}^{\tau,\delta}(\sD; f, f') \in \{0, 1\}.\)
To keep consistent with the instance-level screening, we only consider inputs on which \(f\) and \(f'\) agree:
\(
\sD' = \{ \vx \in \sD \;:\; f(\vx) = f'(\vx) \},
\)
from which we draw samples via rejection sampling.
Let \(q_{\text{proxy}}(\vx)\) and \(q_{\text{oracle}}(\vx)\) denote the (per-instance) fidelities on \(\sD'\) of proxy and oracle explanations, respectively (as defined in Section~\ref{sec:fidelity_metrics}). We conduct a \emph{sequential one-sided paired \(t\)-test} on the paired differences
\[
d_i \,=\, q_{\text{proxy}}(\vx_i) \;-\; \tau\, q_{\text{oracle}}(\vx_i),\qquad i=1,\dots,n,
\]
and test
\[
H_0:\; \mu_d < 0
\quad\text{vs.}\quad
H_1:\; \mu_d \ge 0,
\]
where \(\mu_d = \E[d_i]\) is the population mean difference on \(\sD'\).
At step \(n\) we update the sample mean and variance of the paired differences,
\[
\bar d \,=\, \frac{1}{n}\sum_{i=1}^n d_i,\qquad
s_d^2 \,=\, \frac{1}{n-1}\sum_{i=1}^n (d_i-\bar d)^2.
\]
After each new paired sample, we compute a \((1-\delta)\) confidence interval for \(\mu_d = \overline{q}_{\text{proxy}} - \tau\,\overline{q}_{\text{oracle}}\) as
\[
\left(\,\bar d - t_{\nu,\,1-\delta/2}\,\frac{s_d}{\sqrt{n}}\;,\;\;
\bar d + t_{\nu,\,1-\delta/2}\,\frac{s_d}{\sqrt{n}}\,\right),
\]
where \(t_{\nu,\,1-\delta/2}\) is the \(1-\delta/2\) quantile of the \(t\)-distribution with \(\nu=n-1\) degrees of freedom.

If the entire interval lies above zero, we accept \(H_1\); if it lies entirely below zero, we accept \(H_0\).
Otherwise, we continue sampling until a confident decision is reached or a maximum sample size \(N\) is exhausted.
Finally, if \(H_1\) is accepted, we set \(s_{\text{task}}^{\tau,\delta}(f'; f, \sD) = 1\), indicating that proxy explanations from \(f'\) are sufficiently faithful on average for \(\sD\); otherwise, we set \(s_{\text{task}}^{\tau,\delta}(f'; f, \sD) = 0\).

\textbf{Instance-Level Screening (Online)\;}
If \(f'\) passes the task-level screening, we apply an instance-level check for each input \(\vx\) to filter out cases where the two models disagree.
For a given \(\vx\), the instance-level screening function is
\[
s_{\text{inst}}(\vx; f,f') \;=\; \mathbf{1}\!\left[\, f(\vx) = f'(\vx) \,\right].
\]
The rationale is twofold:
(1) local explanations are designed for the model’s current prediction, so proxy explanations are appropriate only when the two models agree; and
(2) disagreement suggests different local decision behavior around \(\vx\), making proxy explanations more likely to be unfaithful.

\subsection{Applying Proxy Explanations}
If the budget-friendly model \(f'\) passes the task-level screening, and the input instance \(\vx\) passes the instance-level screening, our framework will apply the proxy explanations from \(f'\), i.e., \(t(f',\vx)\), to explain the behavior of the expensive LLM \(f\) around \(\vx\).

For more details, please refer to Appendix~\ref{appendix:framework}.

%% file: chapters/4.result.tex

\section{Fidelity Evaluation}
In this section, we first introduce the experimental setup used in our empirical studies, and then present and analyze results to answer the following research questions:

\begin{enumerate}[noitemsep,topsep=0pt,parsep=0pt,partopsep=0pt, leftmargin=12pt]
    \item \textbf{Cost Reduction:} To what extent can the proposed proxy explanation framework reduce the cost of generating explanations for expensive LLMs? This is the primary focus of our study.
    \item \textbf{Screening Reliability:} How reliable is the screening step in our framework? This checks if the screening step is necessary and sufficient to ensure the fidelity of proxy explanations.
    \item \textbf{Proxy Explanation Generalizability:} Does the transferability of explanations across models hold consistently across different tasks and datasets? This aspect is crucial for demonstrating the generalizability and applicability of our method.
\end{enumerate}

\label{sec:eval}

\subsection{Experimental Setup}
\label{sec:setup}

\input{tables/price.tex}


\subsubsection{Target Models and Explanation Techniques}
We conducted our experiments on 12 popular generative language models, including two from the GPT-4o series, DeepSeek V3, seven Qwen 2.5 models, and two Llama 3.1 models, as listed in Table~\ref{tab:llm-pricing}.
We accessed GPT-4o series and DeepSeek V3 via their official APIs, while running the other models locally.

The models were selected based on their popularity, as well as diversity in architecture, size, and pricing.
They cover both dense and Mixture-of-Experts (MoE) architectures, parameter counts ranging from 0.5 billion (Qwen2.5-0.5B) to 685 billion (DeepSeek V3).
Their associated costs vary significantly: GPT-4o is the most expensive at \$2.50 per million input tokens and \$10.00 per million output tokens, while Qwen2.5-0.5B is the most affordable, whose official APIs are currently free and can also be deployed locally on a single consumer-grade GPU with minimal computational costs.

We use two representative attribution-based explanation techniques: LIME~\citep{LIME} and Kernel SHAP~\citep{SHAP}, to generate local explanations.
For both methods, we set the number of perturbation samples to 1,000 and use default values for all other hyperparameters.
When applying our proxy explanation framework, we set the dataset-level screening threshold \(\tau = 0.9\) and confidence level \(1-\delta = 0.99\), and a maximum sample size \(N=50\). We also conducted sensitivity analysis on hyperparameters in Appendix~\ref{appendix:sensitivity}.

\subsubsection{Tasks and Datasets}
We evaluate our approach on three representative tasks: sentiment analysis, multiple-choice question answering, and text generation, where sentiment analysis is a classic task in studying model explanations, multiple-choice question answering is a common benchmark for evaluating the performance of LLMs, and text generation is a widely used task of LLMs beyond classification.
Considering the budget limitation, we select one dataset from each task to analyze the effectiveness of using proxy explanations. Besides, we conduct our experiments on another three datasets to further validate whether the croess-model explanation transferability holds across different datasets.

\textbf{Sentiment Analysis\;}
We use the Stanford Sentiment Treebank (SST) dataset~\citep{sst} for classification. Following the standard train/validation/test split, we generate explanations for all 2,210 sentences in the test set.
The target model is prompted in a zero-shot setting to predict whether the sentiment of a given sentence is positive or negative.

\textbf{Multiple-Choice Question Answering\;}
We use the MMLU dataset~\citep{mmlu}, which contains 57,000 questions spanning 57 topics. We select 5 topics for evaluation: high school chemistry, high school physics, microeconomics, world history, and computer science.
For each topic, we use the questions in the validation set as the in context examples and the questions in the test set as the target questions.
We generate explanations for all 1321 questions in the test set.

\textbf{Text Generation\;}
We use the Google Natural Questions (NQ) dataset~\citep{nq} for text generation. We randomly select 200 questions from the validation set and generate short answers using the target models.
To apply LIME and Kernel SHAP, we follow prior work~\citep{luss2024cell, hackmann2024wordimportanceexplainsprompts, ReX, MExGen} in using a scoring function \(f_s : \mathcal{X} \rightarrow \mathbb{R}\) that maps the generated sequence to a scalar score, effectively framing the text generation task as a regression problem.
In our experiments, we use \texttt{all-MiniLM-L6-v2}~\citep{wang2020minilm} from the Sentence-Transformers library~\citep{reimers-2019-sentence-bert} as the scoring function. This pre-trained sentence transformer encodes each generated answer into a semantic embedding vector, and we use the cosine similarity between the sample outputs and target outputs as the final scalar score.

\subsection{Fidelity Metrics}

\label{sec:fidelity_metrics}
LIME and Kernel SHAP construct a local surrogate model to approximate the target model's behavior. Following~\citet{balagopalan2022road,yeh2019fidelity,ismail2021improving}, given a target model \(f\), an input \(\vx\), a surrogate explanation model \(g\), a neighborhood distribution \(D(\vx)\), and a performance metric \(L\) (e.g., accuracy, AUROC, or mean squared error (MSE)), the (in)fidelity is defined as:
\(
\mathbb{E}_{\vz \sim D(\vx)} L(f(\vz), g(\vz)).
\)
In our experiments, we use \textbf{accuracy} as the performance metric \(L\).

\subsection{Evaluation Results}
\label{sec:results}

\input{tables/cost.tex}

\subsubsection{RQ1: Cost Reduction of Explaining Expensive LLMs}

We use Cost Reduction Ratio (CRR) to measure the cost reduction achieved by our proxy explanation framework compared to directly generating explanations from expensive LLMs.
Specifically, when explaining an expensive LLM \(f\) with a budget-friendly model \(f'\) on input set \(\sD\) that passes the task-level screening step, we define the CRR as:
\[
\text{CRR} = \frac{C_{\text{oracle}}}{C_{\text{proxy}} + C_{\text{fallback}} + C_{\text{screen}}}
\]
where
\begin{align*}
C_{\text{oracle}} &= \textstyle\sum_{\vx \in \sD} \text{Cost}(f,\vx), \\
C_{\text{proxy}} &= \textstyle\sum_{\vx \in \sD} \text{Cost}(f',\vx) \cdot s_{\text{inst}}(\vx; f,f'), \\
C_{\text{fallback}} &= \textstyle\sum_{\vx \in \sD} \text{Cost}(f,\vx) \cdot (1 \!-\! s_{\text{inst}}(\vx; f,f')), \\
C_{\text{screen}} &= \text{Cost}_{\text{screen}}(f,f',\sD),
\end{align*}
where \(\text{Cost}(f,\vx)\) denotes the cost of generating explanations for model \(f\), \(s(\vx; f,f')\) is the instance-level screening function defined in Section~\ref{sec:framework}, and \(\text{Cost}_{\text{screen}}(f,f',\sD)\) is the cost of performing task-level screening on dataset \(\sD\).
Here, we split the models into two groups based on their costs: all models that can be run locally on a single consumer-grade GPU are considered budget-friendly models, while the rest are classified as target expensive models.

Table~\ref{tab:llm-cost-reduction-shap} shows the CRR achieved by using our proxy explanation framework to explain expensive LLMs with LIME and Kernel SHAP.
We can see that for each expensive model, the use of a budget-friendly proxy model significantly reduces the cost of generating explanations. Especially for the most expensive model GPT-4o, using proxy explanations from budget-friendly models can save at most 88\% of the cost across all these tasks.

\subsubsection{RQ2: Reliability of Screening Step}
\input{tables/screen.tex}

Our task-level screening verifies whether proxy explanations from a budget-friendly model \(f'\) can, on the input set \(\sD\), achieve fidelity comparable to the oracle explanations generated directly from \(f\).
For each task, we validate the reliability of the screening step by checking if the screening results align with the actual fidelity of proxy explanations.
Since the screening decision is a binary classification problem, we assess its reliability using standard classification metrics: Precision, Recall, and F1-score.

Table~\ref{tab:llm-screening} shows that our task-level screening step achieves 98.9\% precision on average, which means that the screening is sufficient to ensure the fidelity of proxy explanations.
For the rare false positives, the realized proxy fidelity still exceeds \(89\%\) of the oracle on average, suggesting that even misclassifications remain reasonably faithful.
On the other hand, although recall is lower, the results of RQ1 demonstrate that for each expensive model there exists at least one budget-friendly model that passes the screening. 
Given that budget-friendly models are inexpensive to run, users can screen multiple candidates in parallel to reliably identify a suitable proxy model.

\subsubsection{RQ3: Generalizability of Proxy Explanations}

To validate if the cross model explanation transferability holds across different tasks and datasets, we conduct experiments on the datasets described in Section~\ref{sec:setup} and three additional datasets: Large Movie Review~\citep{IMDB}, Fake News~\citep{fake-news-dataset}, and Web Question~\citep{berant-etal-2013-semantic} for text generation.
Overall, we observe that the findings from the three main datasets generally hold across these additional datasets, demonstrating the generalizability of our proxy explanation framework.
Due to space limitations, we only show the results of cross-model proxy explanation fidelity between GPT-4o, Qwen2.5-7B, and Qwen2.5-14B on the six tasks in Figure~\ref{fig:rq3_1} and \ref{fig:rq3_2}.
For GPT-4o, Qwen 7B and 14B can both achieve over 90\% fidelity compared to the oracle explanations.

\input{tables/further_experiments.tex}

%

For more detailed results and analysis, please refer to Appendix~\ref{appendix:further_experiments}.

%% file: tables/price.tex
\begin{table}[t]
    \centering
    \caption{Common LLM Official API pricing (USD per million tokens). Specifically, Qwen 2.5 models with 0.5B and 1.5B parameters can be accessed from Alibaba (https://www.aliyun.com/) for \textbf{free}, and all open-source models with 8B or fewer parameters can be run locally on a single consumer-grade GPU.}
    \setlength{\tabcolsep}{1mm} 
    \small
     \resizebox{\linewidth}{!}{ 
    \begin{tabular}{lcccc}
        \toprule
        \textbf{Model} & \textbf{Provider} & \textbf{Input} & \textbf{Output} & \textbf{Open-source} \\
        \midrule
        GPT-4o & OpenAI & \$2.50 & \$10.00 & No \\
        GPT-4o Mini & OpenAI & \$0.15 & \$0.60 & No \\
        DeepSeek V3 & DeepSeek & \$0.27 & \$1.10 & Yes \\
        Qwen 2.5 0.5B & Alibaba & -- & -- & Yes \\
        Qwen 2.5 1.5B & Alibaba & -- & -- & Yes \\
        Qwen 2.5 3B & Alibaba & \$0.04 & \$0.12 & Yes \\
        Qwen 2.5 7B & Alibaba & \$0.07 & \$0.14 &Yes \\
        Qwen 2.5 14B & Alibaba & \$0.14 & \$0.41 & Yes \\
        Qwen 2.5 32B & Alibaba & \$0.28 & \$0.83 & Yes \\
        Qwen 2.5 72B & Alibaba & \$0.56 & \$1.67 & Yes \\
        LLaMA 3.1 8B & Meta & \$0.18 & \$0.18 & Yes \\
        LLaMA 3.1 70B & Meta & \$0.88 & \$0.88 & Yes \\
        \bottomrule
    \end{tabular}
    }
    \label{tab:llm-pricing}
\end{table}

%% file: tables/cost.tex


\begin{table}[t]
    \centering
    \caption{Cost Reduction Ratios (CRR) achieved by using the proxy explanation framework to explain expensive LLMs with LIME and Kernel SHAP. Here, \(\text{CRR}_\text{mean}\) and \(\text{CRR}_\text{max}\) denote the average and maximum CRR obtained from screened budget-friendly models with API access. \(\text{CRR}_\text{local}\) also denotes the maximum CRR, but we run all budget-friendly models locally, thus further reducing the cost.}
    \label{tab:llm-cost-reduction-shap}
    \resizebox{\linewidth}{!}{
    \small
    \setlength{\tabcolsep}{0.5mm} 
    \begin{tabular}{llcccccc}
        \toprule
        & &\multicolumn{3}{c}{LIME} & \multicolumn{3}{c}{Kernel SHAP} \\
        \cmidrule(lr){3-5}\cmidrule(lr){6-8}
        \textbf{Target Model} & CRR & SST & MMLU & NQ & SST & MMLU & NQ \\
        \midrule
        \multirow{3}{*}{GPT-4o} 
            & mean & 8.74 & 3.41 & 5.70 & 8.90 &  3.41 & 7.29 \\
            & max  & 10.33 & 4.84 & 7.41 & 10.33 & 4.84 & 8.20 \\
            & local  & 14.17 & 5.62 & 10.53 & 14.17 & 5.62 & 11.11 \\
            [0.5em]
        \multirow{3}{*}{GPT-4o mini} 
            & mean & 2.50 & 1.83 & 3.65 & 1.97 & 1.96 & 3.43 \\
            & max  & 3.08 & 2.88 & 6.67 & 3.10 & 3.19 & 6.67 \\
            & local  & 13.15 & 4.98 & 6.67 & 14.44 & 5.78 & 10.53 \\
            [0.5em]
        \multirow{3}{*}{DeepSeek V3} 
            & mean & 3.06 & 2.15 & 2.40 & 3.88 & 2.27 & 3.85 \\
            & max  & 4.60 & 3.05 & 4.17 & 6.33 & 3.16 & 7.41 \\
            & local  & 13.31 & 5.32 & 8.33 & 13.64 & 6.10 & 8.33 \\
            [0.5em]
        \multirow{3}{*}{Qwen 2.5 14B} 
            & mean & 2.39 & 1.82 & 2.72 & 1.85 & 1.90 & 3.49 \\
            & max  & 2.90 & 2.99 & 6.06 & 2.89 & 3.20 & 6.25 \\
            & local  & 17.13 & 5.64 & 11.76 & 17.13 & 6.10 & 11.76 \\
            [0.5em]
        \multirow{3}{*}{Qwen 2.5 32B} 
            & mean & 3.18 & 2.06 & 3.74 & 3.19 & 2.16 & 3.86 \\
            & max  & 4.77 & 3.15 & 6.06 & 4.77 & 3.05 & 6.06 \\
            & local  & 15.24 & 5.30 & 9.09 & 15.24 & 5.31 & 9.09 \\
            [0.5em]
        \multirow{3}{*}{Qwen 2.5 72B} 
            & mean & 5.10 & 2.47 & 3.97 & 5.05 & 2.76 & 5.13 \\
            & max  & 7.04 & 3.40 & 6.25 & 6.91 & 3.66 & 7.14 \\
            & local  & 16.25 & 6.07 & 9.09 & 16.25 & 6.31 & 10.53 \\
            [0.5em]
        \multirow{3}{*}{Llama 3.1 70B} 
            & mean & 5.32 & 2.92 & 3.72 & 6.14 & 2.93 & 4.92 \\
            & max  & 6.74 & 3.96 & 5.13 & 8.02 & 3.96 & 6.67 \\
            & local  & 10.33 & 5.77 & 6.90 & 17.13 & 5.77 & 10.00 \\
        \bottomrule
        \end{tabular}
    }
\end{table}

%% file: tables/screen.tex
\begin{table}
    \centering
    \caption{Screening recall, precision, and F1-score of the Proxy Explanation Framework.}
    \label{tab:llm-screening}
    \resizebox{\linewidth}{!}{
    \small
    \setlength{\tabcolsep}{1mm}
    \begin{tabular}{lcccccc}
        \toprule
        \textbf{Method} & \multicolumn{3}{c}{LIME} & \multicolumn{3}{c}{Kernel SHAP} \\
        \cmidrule(lr){2-4}\cmidrule(lr){5-7}
        \textbf{Datasets} & SST & MMLU & NQ & SST & MMLU & NQ \\
        \midrule 
        \textbf{Precision (\%)}  & 100.0 & 99.4 & 94.1 & 100.0& 100.0 & 100.0\\
        \textbf{Recall (\%)}  & 80.2& 77.6 & 76.1 & 96.3 & 97.2 & 96.2\\
        \textbf{F1-score (\%)} & 89.0& 87.2 & 84.2 & 98.1 & 98.5 & 98.0\\
        \bottomrule
    \end{tabular}
    }
\end{table}

%% file: tables/further_experiments.tex
\begin{figure}[t]
\centering
\setlength{\tabcolsep}{2mm}
\small
\resizebox{0.8\linewidth}{!}{
\includegraphics{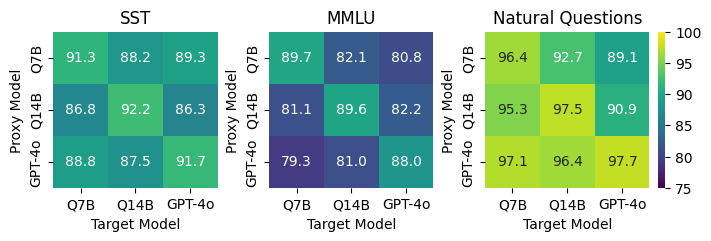}
}
\caption{Accuracy of proxy LIME explanations on SST, MMLU, and Natural Questions datasets.}
\label{fig:rq3_1}
\end{figure}


\begin{figure}[t]
\centering

\setlength{\tabcolsep}{1mm}
\small
\resizebox{0.8\linewidth}{!}{
    \includegraphics{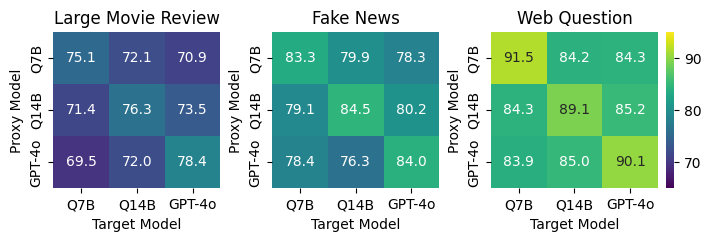}
}
\caption{Accuracy of proxy LIME explanations on Large Movie Review, Fake News, and Web Question datasets.}
\label{fig:rq3_2}
\end{figure}

%% file: chapters/4.5.incontext.tex
\section{Actionable Utility Evaluation}

\label{sec:CaseStudy}

\input{tables/ICL_res.tex}

\input{tables/poison.tex}
Having established the statistical fidelity of proxy explanations in the previous section, we now evaluate the \textbf{actionable utility} of our framework in real-world optimization scenarios.
We focus on \textbf{In-Context Learning (ICL)} settings, where the quality of the input prompt directly dictates model performance.
In this context, explanations serve not only as passive observations, but also as active signals to diagnose redundancy or identify malicious patterns.

Specifically, we use proxy explanations generated from budget-friendly models that have passed the screening step, to guide two distinct optimization tasks, and compare their effectiveness against baselines including using oracle explanations and other state-of-the-art methods.
We demonstrate this utility across two distinct tasks:
(1) \textbf{Prompt Compression} (Efficiency); and
(2) \textbf{Poisoned Example Removal} (Safety).
\subsection{Task 1: Prompt Compression}

Prompt compression aims to help users save costs by reducing the number of examples in the prompt while maintaining the model's performance.


\paragraph{Experiment Setup}
We use explanations to compress the ICL examples in using GPT-4o to answer questions from five subjects of in MMLU~\citep{mmlu}, HellaSwag~\citep{hellaswag}, GSM8K~\citep{gsm8k}, and PIQA~\citep{piqa} datasets.
We compress the prompt by removing the least important examples based on KSHAP explanations.
%
Concretely, we define the prompt compression task as follows:
Given a set of examples \(S\), an explanation \(g\) that attributes the importance of each example, we iteratively remove the least important examples based on \(g\). The goal is to remove as many examples as possible while ensuring that GPT-4o's performance keep above a certain threshold \(\tau\) compared to using all examples (we set \(\tau = 0.9\)).
We use the compression ratio as the metric, which is defined as 
\(\text{CompressionRatio(g)} = 1 - \frac{|S_g^{\tau}|}{|S|},\)
where \(S_g^{\tau}\) is the set of examples retained after applying the explanation \(g\) and ensuring the model's performance remains above the threshold \(\tau\).
A higher compression ratio indicates a more effective explanation.
We verify if the proxy explanations from a budget-friendly model of Qwen 2.5 series can achieve similar performance as oracle explanations from GPT-4o. 
Additionally, we also compare the performance of proxy explanations with a random deletion baseline, and two state-of-the-art prompt compression methods, AttnComp~\citep{zhao2025leveraging} and LLMLingua~\citep{jiang2023llmlinguacompressingpromptsaccelerated}.

%

For each dataset, we repeat the experiment 15 times with different ICL examples and test questions.


\paragraph{Evaluation Results}
Table~\ref{tab:removal_ratios} shows the results. Proxy explanations achieve performance comparable to the oracle explanations from GPT-4o, reaching an average of 91.7\% of the oracle's performance. Moreover, they outperform random deletions and two state-of-the-art prompt compression methods.

\subsection{Task 2: Poisoned Examples Removal}

ICL is useful, while poisoning examples can lead to suboptimal performance~\citep{ranjan2023fooling}. 
Outlier removal focuses on identifying and removing examples that may negatively impact the model's performance, thereby improving the overall quality of the prompt.

\paragraph{Experiment Setup}
We use GPT-4o to perform sentiment analysis on the SST-2 dataset~\citep{sst} as our target task.
We use ICLPoison~\cite{he2025datapoisoningincontextlearning} to add poisoning examples to the original dataset, until the accuracy of GPT-4o drops to lower than 80\%,
and then use explanations to identify and remove these outliers.
We follow the explanation to remove all negatively attributed examples, and evaluate the model's performance after removal.
We compare the performance using oracle explanations from GPT-4o and proxy explanations generated by our methods, along with a random deletion baseline.

\paragraph{Evaluation Results}
Table~\ref{tab:poison} shows the results. After removing the poisoned examples based on explanations, the accuracy of GPT-4o recovers significantly, and proxy explanations achieve comparable performance to oracle explanations, both significantly outperforming random deletion.


%% file: tables/ICL_res.tex
\begin{table*}[t]
\centering
\setlength{\tabcolsep}{1mm}
\small
\caption{Comparison of compression ratios (\%) using oracle explanations, proxy explanations, random deletions and other baselines for in-context learning prompt compression across various tasks.}
\label{tab:removal_ratios}

\resizebox{\linewidth}{!}{
\begin{tabular}{lcccccccc}
\toprule
\textbf{Task} & MMLU-Chem. & MMLU-CS & MMLU-Micro. & MMLU-Psy. & MMLU-WH & HellaSwag & GSM8K  & PIQA\\
\midrule
    Random & 29.0 & 35.6 & 59.8 & 61.0 & 43.4 & 58.8 & 25.3 &  54.3 \\
    AttnComp & 34.5 & 39.1 & 63.2 & 66.5 & 47.1 & 64.3 & 30.2 & 60.2 \\ 
    LLMLingua & 38.7 & 38.3 & 62.5 & 66.1 & 46.3 & 62.7 & 28.9 & 58.7 \\
    \textbf{Proxy Exp.} & \textbf{41.0} & \textbf{43.0} & \textbf{67.6} & \textbf{69.8} & \textbf{52.0} & \textbf{70.1} & \textbf{35.5} & \textbf{64.5} \\[0.2em]
    Oracle Exp. & 49.2 & 50.2 & 72.8 & 72.8 & 57.0 & 75.5 & 37.2 & 69.2 \\
\bottomrule
\end{tabular}
}
\end{table*}



%% file: tables/poison.tex
\begin{table}[t]
\centering
\caption{Comparison of accuracy (\%) of GPT-4o after using different methods to remove poisoned examples.}
\setlength{\tabcolsep}{1mm}
\small
\resizebox{\linewidth}{!}{
\begin{tabular}{lcccc}
\toprule
    \textbf{Task} &  Oracle Exp. & \textbf{Proxy Exp.} & Random deletion \\
\midrule
    SST & 94.2 & 94.0 & 87.1 \\
    HellaSwag & 93.7 & 93.5 & 88.4 \\
    PIQA & 91.5 & 90.7 & 79.6 \\
\bottomrule
\end{tabular}
}
\label{tab:poison}
\end{table}

%% file: chapters/7.conclusion.tex
\section{Conclusion}
In this paper, we introduced a screen-and-apply proxy explanation framework that leverages budget-friendly models to generate proxy explanations for LLMs, reducing the cost of local model-agnostic explanations and enabling their practical use. We demonstrated the effectiveness of our approach through extensive experiments across various tasks and models.
The results indicate that our proxy explanations maintain a high level of fidelity compared to oracle explanations while significantly reducing the cost by 88.2\%. We also show that our proxy explanations can enhance the performance of expensive LLMs in few-shot learning scenarios.

%% file: chapters/6.limitation.tex
\section*{Limitations}
\label{sec:limitations_future}

Our work focuses on perturbation-based feature attribution for black-box LLMs. While our \textit{Screen-and-Apply} mechanism ensures reliability, we acknowledge that in scenarios requiring extreme reasoning capabilities (e.g., complex symbolic logic), the alignment between small proxies and large oracles may weaken. In such cases, our framework prioritizes safety by falling back to the oracle, which guarantees fidelity but may reduce the magnitude of cost savings.
To mitigate this, a potential direction is to align proxy models with oracles through lightweight fine-tuning. We leave this exploration to future work.

%% file: chapters/appendix.tex
\clearpage

\appendix

\section{The Use of Large Language Models}
We use LLMs to refine and polish human writing, and find related work with DeepResearch. We do not use LLMs to generate the main content or ideas of this paper.

\section{Datasets}
To reduce the cost of future research on black-box explanation generation for LLMs, improve accessibility, and facilitate reproducibility, we have open-sourced the datasets used in our experiments.
In particular, since querying LLMs for perturbed samples is the most computationally expensive part of the process, we provide the model outputs for all perturbed samples used in our experiments.
For LIME, perturbations are generated following the original implementation,\footnote{\url{https://github.com/marcotcr/lime}} and for Kernel SHAP, we use the implementation provided by the Captum library.\footnote{\url{https://captum.ai/api/kernel_shap.html}} We generate 1000 perturbed samples for each input instance explained.

As mentioned in Section~\ref{sec:intro}, the datasets cover six tasks: five representative subjects from the MMLU benchmark—High School Chemistry, High School Computer Science, High School Microeconomics, High School Psychology, and High School World History—and the SST-2 sentiment classification dataset.
For each perturbed sample, we collect the model output logits of the first predicted token.

Additionally, as described in Section~\ref{sec:setup}, we select 200 questions from the Natural Questions dataset for the text generation task.
We release both the model  outputs for each perturbed sample of these questions for reproducibility.

As we create the datasets with public datasets and publicly accessible models, there are no ethical or legal issues involved in releasing these datasets.

\section{Proxy Explanation Framework Details}
\label{appendix:framework}
As most budget-friendly models can be run locally, we can perform the task-level screening for multiple budget-friendly models at the same time, as shown in Algorithm~\ref{alg:task_screening}.

\begin{algorithm}[t]
\caption{Task-level Screening for Multiple Proxy Models}
\label{alg:task_screening}
\KwIn{
Target model $f$; candidate proxy models $\{f'_1,\dots,f'_m\}$; 
dataset $\sD$; explanation technique $t$; fidelity threshold $\tau$; 
confidence level $1-\delta$; maximum sample size $N$.
}
\KwOut{
Screening decisions $\{s_{\text{task}}^{\tau,\delta}(f'_j; f, \sD)\}_{j=1}^m$.
}

\ForEach{proxy model $f'_j$}{
    Initialize $n \gets 0$, paired difference set $\mathcal{D}_j \gets \emptyset$\;
    Define $\sD'_j = \{\vx \in \sD: f(\vx) = f'_j(\vx)\}$\;

    \While{decision not reached and $n < N$}{
        Sample $\vx_i$ from $\sD'_j$ via rejection sampling\;
        Compute fidelities $q_{\text{proxy}}(\vx_i)$ and $q_{\text{oracle}}(\vx_i)$ with Buffer $\mathcal{B}$ and update $\mathcal{B}$\;
        Form difference $d_i = q_{\text{proxy}}(\vx_i) - \tau \, q_{\text{oracle}}(\vx_i)$\;
        Update $\hat d_j$ and variance $s_{d,j}^2$\;
        Construct $(1-\delta)$ confidence interval for $\mu_d$\;

        \eIf{interval $> 0$}{
            Accept $H_1$; set $s_{\text{task}}^{\tau,\delta}(f'_j; f, \sD)=1$\;
        }{
            \eIf{interval $< 0$}{
                Accept $H_0$; set $s_{\text{task}}^{\tau,\delta}(f'_j; f, \sD)=0$\;
            }{
                Continue sampling\;
            }
        }
    }
    \If{no decision after $N$ samples}{
        Set $s_{\text{task}}^{\tau,\delta}(f'_j; f, \sD)=0$\;
    }
}
\end{algorithm}

To avoid redundant oracle queries, we introduce a shared buffer that stores 
each input and the corresponding output from the target model $f$. 
Formally, we construct
\[
\mathcal{B} = \{(\vx, f(\vx)) : \vx \in \sD\}.
\]
This buffer is built once and reused across all candidate proxy models 
$\{f'_1,\dots,f'_m\}$.

When each time calculating the fidelity, if the output of $\vx$ and its neighborhood are already in the buffer $\mathcal{B}$, we can directly use the cached values. This avoids redundant calls to the target model $f$ and speeds up the screening process.

\section{Experimental Setup Details}

\subsection{Models}




\subsubsection{Deployment}
We run Qwen2.5 and Llama3.1 models locally on a machine with total 576 GiB VRAM, while GPT-4o and DeepSeekV3 models are accessed via their official APIs.
When locally running the models, we use the default version without additionally quantization or distillation.

\subsection{Sentiment Analysis}
We perform sentiment analysis using LLMs in a zero-shot setting, where the model is prompted to classify the sentiment of a given sentence as either "positive" or "negative." 
The sentiment classification task is defined in the system prompt, while the specific sentence to be classified is provided in the user input. 
The following prompt templates are used:

\begin{quote}
\textbf{system\_prompt}: \\
\begin{flushleft}
\ttfamily
From now on, you should act as a sentiment analysis neural network. \\
You should classify the sentiment of a sentence into positive or negative. \\
The input sentence may be empty. In each task, you will be given the sentences to be classified, which end with \#\#\#\#\#, and then you should reply the sentiment of the sentence by positive or negative.
\end{flushleft}

\vspace{1em}

\textbf{user\_prompt}: \\
\begin{flushleft}
\ttfamily
Perform the following task, your answer should only be positive or negative: \\
Sentence: \\
\{input\_sentence\} \\
\#\#\#\#\# \\
\vspace{0.5em}
Sentiment:
\end{flushleft}
\end{quote}

To obtain the class probabilities, we use the probabilities of the first output token.
Specifically, we extract the logits corresponding to the tokens \texttt{"positive"} and \texttt{"negative"} from the model output and apply the softmax function to compute the probability distribution over the two classes.
For local models, we directly obtain the logits and compute the softmax.
For GPT-4o and DeepSeekV3, we use their official APIs with the temperature set to 1, retrieve the log probabilities of the target tokens, and then apply the softmax function to obtain normalized probabilities.

\subsubsection{Multiple-Choice Question Answering}
\input{figures/few-shot-template.tex}
We perform multiple-choice question answering using LLMs with few-shot prompting, where the model is provided with all examples from the development set using the in-context learning template recommended by OpenAI,\footnote{\url{https://platform.openai.com/docs/guides/prompt-engineering/six-strategies-for-getting-better-results\#tactic-provide-examples}} as illustrated in Figure~\ref{fig:few-shot-example}.
We follow the official instructions provided in OpenAI Evals.\footnote{\url{https://github.com/openai/evals/blob/main/examples/mmlu.ipynb}}

The prompt template is as follows:
\begin{quote}
\textbf{System:} \texttt{The following are multiple choice questions (with answers) about \{subject\}.} \\
\textbf{User:} \texttt{\{example question 1\}} \\
\textbf{Assistant:} \texttt{\{example answer 1\}} \\
\hspace{1em} \vdots \\
\textbf{User:} \texttt{\{question to be answered\}}
\end{quote}

To obtain the class probabilities, we use similar logit extraction methods as in the sentiment analysis task.
Specifically, we extract the logits corresponding to the tokens \texttt{"A"}, \texttt{"B"}, \texttt{"C"}, and \texttt{"D"} from the model output and apply the softmax function to compute the probability distribution over the four classes.

\subsubsection{Text Generation}
As mentioned in Section~\ref{sec:setup}, we treat this task as a regression problem by using a scoring function to evaluate the generated text.
When generating text with LLMs, we set the temperature to \(1\text{e-}5\) or set \texttt{do\_sample = False} to ensure deterministic outputs.
We limit the maximum number of generated tokens to 20 and prompt the model to generate short answers, in line with the short-answer format of the Natural Questions dataset.
Specifically, we use the following prompt template:
\begin{quote}
\begin{verbatim}
[
  {"role": "system", "content": 
   "You are a helpful assistant. 
   Answer the question briefly,
   within 10 words. You will be  
   penalized for overly long answers."
   },
  {"role": "user", 
  "content": "{Question}"}
]
\end{verbatim}
\end{quote}

\
\section{Hyperparameter Study}
\label{appendix:sensitivity}
\begin{figure*}
\centering
\includegraphics[width=0.8\textwidth]{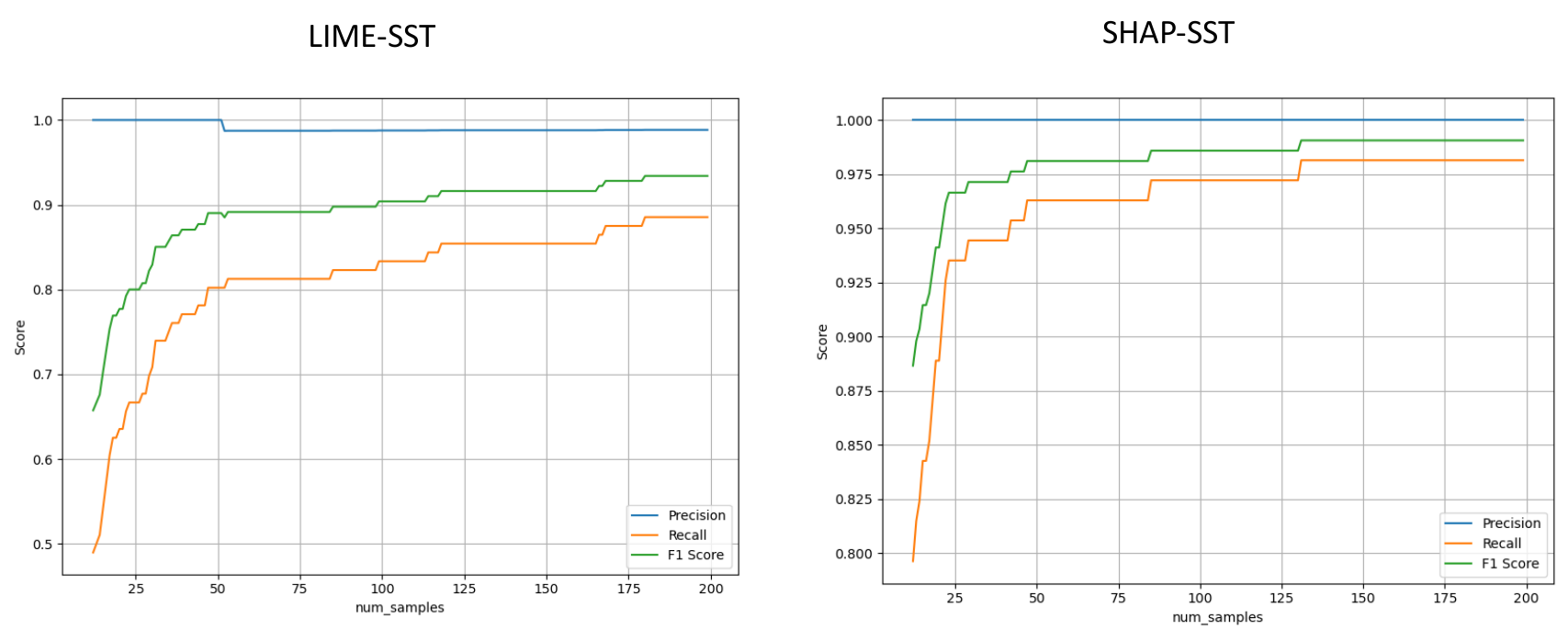}

\caption{Impact of maximum sample size \(N\) on task-level screening results of proxy explanations on the SST dataset.}
\label{fig:Nsensitive}
\end{figure*}

We study the impact of hyperparameters used in our experiments.
Since we have chosen commonly used values for $\tau$ and $\delta$ in statistical testing, we primarily investigate the impact of the maximum sample size on the effectiveness of task-level screening.

Figure~\ref{fig:Nsensitive} shows the results of task-level screening on the SST dataset with different maximum sample sizes \(N\).
We observe that as \(N\) increases, the recall of task-level screening also increases, while the precision remains consistently high.
As we have demonstrated in Section~\ref{sec:eval}, for each expensive LLM, we can find at least one budget-friendly model that passes the task-level screening, indicating the setting of \(N\) does not significantly affect the overall effectiveness of our framework.

\section{RQ3: Generalizability of Proxy Explanation Details}
\label{appendix:further_experiments}
In this section, we further analysis the evaluation results.
\subsection{Sentiment Analysis}
\input{tables/sst_fidelity.tex}
%
Table~\ref{tab:sst_lime_fidelity} and~\ref{tab:sst_shap_fidelity} provides the corresponding detailed results, including 95\% confidence intervals.

%
%
%

%

\subsection{Multiple-Choice Question Answering}

\input{figures/mmlu_fidelity.tex}
\input{figures/mmlu_shap_fidelity.tex}
\input{figures/mmlu_subject_lime_fidelity.tex}
\input{figures/mmlu_subject_shap_fidelity.tex}
\input{tables/mmlu_lime_fidelity.tex}
\input{tables/mmlu_shap_fidelity.tex}

We have provided overall fidelity results in Figure~\ref{fig:mmlu_fidelity} and~\ref{fig:mmlu_shap_fidelity}, and we provide also the explanation fidelity results of Kernel SHAP in Figure~\ref{fig:mmlu_shap_fidelity}, and on each subject in Figure~\ref{fig:mmlu_fidelity_lime_subject} and Figure~\ref{fig:mmlu_fidelity_shap_subject}.

We can see the observation we find in section \ref{sec:eval} also holds for Kernel SHAP and each subject, i.e., filtering out examples where the budget-friendly and expensive models produce different predictions for the same input can significantly improve the fidelity of proxy explanations. Additionally, we also provide the detailed results with 95\% confidence intervals in Table~\cref{tab:mmlu_lime_chemistriy_fidelity}, \ref{tab:mmlu_lime_chemistriy_filtered_fidelity}, \ref{tab:mmlu_lime_computer_science_fidelity}, \ref{tab:mmlu_lime_computer_science_filtered_fidelity}, \ref{tab:mmlu_lime_microeconomics_fidelity}, \ref{tab:mmlu_lime_microeconomics_filtered_fidelity}, \ref{tab:mmlu_lime_psychology_fidelity}, \ref{tab:mmlu_lime_psychology_filtered_fidelity}, \ref{tab:mmlu_lime_world_history_fidelity}, \ref{tab:mmlu_lime_world_history_filtered_fidelity}, \ref{tab:mmlu_shap_chemistriy_fidelity}, \ref{tab:mmlu_shap_chemistriy_filtered_fidelity}, \ref{tab:mmlu_shap_computer_science_fidelity}, \ref{tab:mmlu_shap_computer_science_filtered_fidelity}, \ref{tab:mmlu_shap_microeconomics_fidelity}, \ref{tab:mmlu_shap_microeconomics_filtered_fidelity}, \ref{tab:mmlu_shap_psychology_fidelity}, \ref{tab:mmlu_shap_psychology_filtered_fidelity}, \ref{tab:mmlu_shap_world_history_fidelity}, and \ref{tab:mmlu_shap_world_history_filtered_fidelity}.

Another notable observation is the fidelity of oracle explanations also differ in different subjects.
For \texttt{high school microeconomics}, \texttt{high school psychology}, and \texttt{high school world history}, the oracle LIME explanations generated by the model all achieve a fidelity higher than 90\%, while for \texttt{high school computer science}, \texttt{high school chemistry}, and \texttt{high school physics}, the fidelity is relative lower.
The subjects with higher fidelity are all related to social sciences, while the subjects with lower fidelity are all related to natural sciences.
This may be due to the fact that social science questions often have more diverse and complex answer options, leading performace differences between models.

\section{Licenses of Used Artifacts}
\label{appendix:licenses}
We list the licenses of all the major artifacts used in our work as follows:
\begin{itemize}
    \item LIME: MIT License
    \item SHAP: MIT License
    \item MMLU dataset: MIT License
    \item SST-2 dataset: CC0: Public Domain
    \item Natural Questions dataset: Apache License 2.0
    \item Qwen2.5 models: Apache License 2.0
    \item Llama3.1 models: llama 3.1 license
    \item DeepSeekV3 model: MIT License
    \item Large Movie Review Dataset: CC0 Public Domain
    \item Web Questions: CC0 Public Domain
    \item HellaSwag: CC0 Public Domain
    \item GSM8K: MIT License
\end{itemize}

We use all these artifacts consistent with their intended use and in compliance with their respective licenses.

\section{Resource Usage}

We conducted all experiments on some servers with GPUs that each have 80GB VRAM. The total GPU hours used is around 3000 GPU hours. The total cost of using GPT-4o and DeepSeekV3 APIs is around 20000 USD.

%% file: figures/few-shot-template.tex
\begin{figure}[t]
    \centering
    \resizebox{\linewidth}{!}{
    \includegraphics{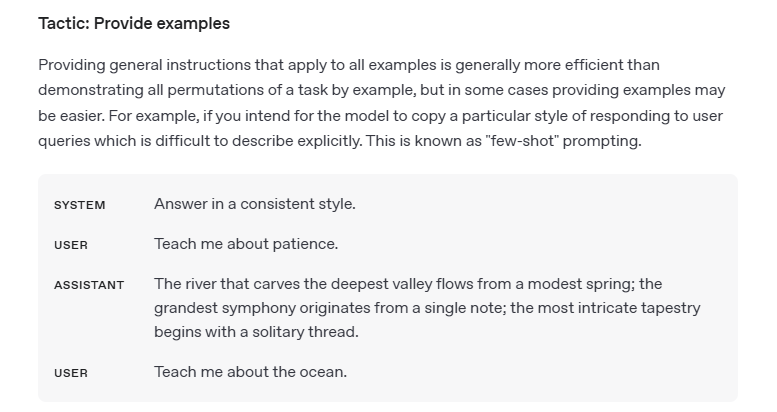}
    }
    \caption{The recommended few-shot template provided by OpenAI.}
    \label{fig:few-shot-example}
\end{figure}    

%% file: tables/sst_fidelity.tex
\begin{table*}
\centering
\caption{Accuracy of proxy LIME explanations on the text classification task: each value shows how well LIME explanations generated by the model on the \textbf{left} serve as surrogates for predicting the behavior of the model on the \textbf{top}.}
\label{tab:sst_lime_fidelity}
\resizebox{\textwidth}{!}{
\begin{tabular}{lcccccc}
\toprule
 & Qwen 0.5B & Qwen 1.5B & Qwen 3B & Qwen 7B & Qwen 14B & Qwen 32B \\
\midrule
Qwen 0.5B & \textbf{98.81\% $\pm$ 0.12} & 41.35\% $\pm$ 1.19 & 19.26\% $\pm$ 0.87 & 24.58\% $\pm$ 1.02 & 18.24\% $\pm$ 0.86 & 26.35\% $\pm$ 1.04 \\
Qwen 1.5B & 41.20\% $\pm$ 1.44 & \textbf{86.22\% $\pm$ 0.27} & 75.85\% $\pm$ 0.78 & 79.48\% $\pm$ 0.67 & 75.41\% $\pm$ 0.83 & 79.74\% $\pm$ 0.62 \\
Qwen 3B & 16.72\% $\pm$ 0.95 & 73.13\% $\pm$ 0.67 & \textbf{91.15\% $\pm$ 0.29} & 86.79\% $\pm$ 0.47 & 89.34\% $\pm$ 0.42 & 85.23\% $\pm$ 0.49 \\
Qwen 7B & 23.30\% $\pm$ 1.15 & 77.50\% $\pm$ 0.56 & 87.17\% $\pm$ 0.48 & \textbf{91.27\% $\pm$ 0.27} & 88.21\% $\pm$ 0.48 & 87.09\% $\pm$ 0.41 \\
Qwen 14B & 16.32\% $\pm$ 0.94 & 72.56\% $\pm$ 0.71 & 88.37\% $\pm$ 0.43 & 86.79\% $\pm$ 0.49 & \textbf{92.22\% $\pm$ 0.28} & 86.11\% $\pm$ 0.46 \\
Qwen 32B & 24.57\% $\pm$ 1.18 & 77.77\% $\pm$ 0.55 & 86.31\% $\pm$ 0.51 & 87.72\% $\pm$ 0.43 & 88.34\% $\pm$ 0.47 & \textbf{90.15\% $\pm$ 0.27} \\
Qwen 72B & \textbf{26.62\% $\pm$ 1.25} & 78.97\% $\pm$ 0.53 & 85.61\% $\pm$ 0.54 & 87.89\% $\pm$ 0.43 & 87.07\% $\pm$ 0.54 & 88.17\% $\pm$ 0.36 \\
Llama 8B & 31.55\% $\pm$ 1.31 & 80.21\% $\pm$ 0.49 & 82.60\% $\pm$ 0.63 & 85.92\% $\pm$ 0.48 & 82.81\% $\pm$ 0.66 & 84.39\% $\pm$ 0.50 \\
Llama 70B & 42.88\% $\pm$ 1.49 & 81.56\% $\pm$ 0.44 & 74.45\% $\pm$ 0.86 & 78.97\% $\pm$ 0.71 & 74.33\% $\pm$ 0.90 & 79.49\% $\pm$ 0.67 \\
DeepSeekV3 & 22.65\% $\pm$ 1.14 & 76.95\% $\pm$ 0.58 & 87.30\% $\pm$ 0.48 & 88.28\% $\pm$ 0.41 & 88.42\% $\pm$ 0.48 & 87.68\% $\pm$ 0.39 \\
GPT-4o Mini & 11.87\% $\pm$ 0.79 & 69.04\% $\pm$ 0.82 & 88.08\% $\pm$ 0.45 & 84.63\% $\pm$ 0.60 & 89.61\% $\pm$ 0.41 & 83.27\% $\pm$ 0.59 \\
GPT-4o & 25.52\% $\pm$ 1.22 & 78.53\% $\pm$ 0.54 & 86.06\% $\pm$ 0.53 & 88.79\% $\pm$ 0.38 & 87.47\% $\pm$ 0.53 & 87.63\% $\pm$ 0.38 \\
\bottomrule
\end{tabular}
}

\quad
\resizebox{\textwidth}{!}{
\begin{tabular}{lcccccc}
\toprule
 & Qwen 72B & Llama 8B & Llama 70B & DeepSeekV3 & GPT-4o Mini & GPT-4o \\
\midrule
Qwen 0.5B & 28.12\% $\pm$ 1.10 & 32.90\% $\pm$ 1.13 & 42.27\% $\pm$ 1.26 & 24.45\% $\pm$ 1.03 & 12.72\% $\pm$ 0.71 & 26.38\% $\pm$ 1.10 \\
Qwen 1.5B & 81.09\% $\pm$ 0.58 & 81.18\% $\pm$ 0.52 & 83.23\% $\pm$ 0.44 & 79.19\% $\pm$ 0.66 & 71.48\% $\pm$ 0.99 & 80.89\% $\pm$ 0.62 \\
Qwen 3B & 84.74\% $\pm$ 0.51 & 80.38\% $\pm$ 0.60 & 72.51\% $\pm$ 0.75 & 86.90\% $\pm$ 0.47 & 90.61\% $\pm$ 0.45 & 86.05\% $\pm$ 0.52 \\
Qwen 7B & 87.62\% $\pm$ 0.39 & 84.15\% $\pm$ 0.45 & 78.15\% $\pm$ 0.59 & 88.35\% $\pm$ 0.40 & 86.90\% $\pm$ 0.64 & 89.32\% $\pm$ 0.36 \\
Qwen 14B & 85.12\% $\pm$ 0.50 & 79.95\% $\pm$ 0.62 & 72.13\% $\pm$ 0.78 & 87.09\% $\pm$ 0.46 & 91.29\% $\pm$ 0.42 & 86.25\% $\pm$ 0.52 \\
Qwen 32B & 88.53\% $\pm$ 0.35 & 83.09\% $\pm$ 0.48 & 78.47\% $\pm$ 0.59 & 88.60\% $\pm$ 0.39 & 86.11\% $\pm$ 0.64 & 88.83\% $\pm$ 0.38 \\
Qwen 72B & \textbf{90.53\% $\pm$ 0.26} & 84.29\% $\pm$ 0.44 & 80.52\% $\pm$ 0.52 & 88.76\% $\pm$ 0.38 & 84.71\% $\pm$ 0.71 & 89.31\% $\pm$ 0.37 \\
Llama 8B & 85.94\% $\pm$ 0.44 & \textbf{88.82\% $\pm$ 0.26} & 83.02\% $\pm$ 0.44 & 85.39\% $\pm$ 0.50 & 80.45\% $\pm$ 0.82 & 86.53\% $\pm$ 0.46 \\
Llama 70B & 81.64\% $\pm$ 0.59 & 82.34\% $\pm$ 0.50 & \textbf{88.31\% $\pm$ 0.25} & 79.32\% $\pm$ 0.69 & 70.17\% $\pm$ 1.07 & 81.10\% $\pm$ 0.65 \\
DeepSeekV3 & 88.17\% $\pm$ 0.36 & 83.49\% $\pm$ 0.47 & 77.81\% $\pm$ 0.60 & \textbf{91.09\% $\pm$ 0.27} & 87.56\% $\pm$ 0.61 & 89.26\% $\pm$ 0.37 \\
GPT-4o Mini & 81.99\% $\pm$ 0.64 & 77.21\% $\pm$ 0.73 & 68.32\% $\pm$ 0.90 & 85.14\% $\pm$ 0.57 & \textbf{94.07\% $\pm$ 0.26} & 83.58\% $\pm$ 0.65 \\
GPT-4o & 88.46\% $\pm$ 0.35 & 84.30\% $\pm$ 0.44 & 79.83\% $\pm$ 0.55 & 88.87\% $\pm$ 0.38 & 85.57\% $\pm$ 0.70 & \textbf{91.67\% $\pm$ 0.26} \\
\bottomrule
\end{tabular}
}
\end{table*}

\begin{table*}
\centering
\caption{Accuracy of proxy Kernel SHAP explanations on the text classification task: each value shows how well Kernel SHAP explanations generated by the model on the \textbf{left} serve as surrogates for predicting the behavior of the model on the \textbf{top}.}
\label{tab:sst_shap_fidelity}
\resizebox{\textwidth}{!}{
\begin{tabular}{lcccccc}
\toprule
 & Qwen 0.5B & Qwen 1.5B & Qwen 3B & Qwen 7B & Qwen 14B & Qwen 32B \\
\midrule
Qwen 0.5B & \textbf{98.55\% $\pm$ 0.16} & 41.38\% $\pm$ 1.18 & 19.29\% $\pm$ 0.87 & 24.61\% $\pm$ 1.03 & 18.27\% $\pm$ 0.87 & 26.38\% $\pm$ 1.04 \\
Qwen 1.5B & 79.40\% $\pm$ 0.90 & \textbf{60.14\% $\pm$ 0.81} & 39.43\% $\pm$ 0.81 & 44.50\% $\pm$ 0.85 & 38.46\% $\pm$ 0.83 & 46.20\% $\pm$ 0.84 \\
Qwen 3B & 28.79\% $\pm$ 1.25 & 75.00\% $\pm$ 0.59 & \textbf{81.69\% $\pm$ 0.77} & 82.12\% $\pm$ 0.69 & 81.05\% $\pm$ 0.82 & 80.75\% $\pm$ 0.69 \\
Qwen 7B & 23.17\% $\pm$ 0.98 & 73.51\% $\pm$ 0.60 & 82.98\% $\pm$ 0.63 & \textbf{84.77\% $\pm$ 0.54} & 83.98\% $\pm$ 0.63 & 82.25\% $\pm$ 0.57 \\
Qwen 14B & 23.21\% $\pm$ 0.98 & 73.13\% $\pm$ 0.60 & 82.83\% $\pm$ 0.64 & 83.51\% $\pm$ 0.58 & \textbf{84.77\% $\pm$ 0.61} & 82.58\% $\pm$ 0.57 \\
Qwen 32B & 24.43\% $\pm$ 0.98 & 73.50\% $\pm$ 0.58 & 82.15\% $\pm$ 0.65 & 83.00\% $\pm$ 0.58 & 83.48\% $\pm$ 0.64 & \textbf{83.15\% $\pm$ 0.55} \\
Qwen 72B & 25.17\% $\pm$ 1.02 & 74.30\% $\pm$ 0.57 & 82.26\% $\pm$ 0.66 & 83.49\% $\pm$ 0.58 & 83.34\% $\pm$ 0.66 & 82.80\% $\pm$ 0.56 \\
Llama 8B & 24.91\% $\pm$ 0.98 & 74.01\% $\pm$ 0.58 & 82.14\% $\pm$ 0.63 & 83.21\% $\pm$ 0.57 & 82.91\% $\pm$ 0.64 & 81.62\% $\pm$ 0.58 \\
Llama 70B & 30.06\% $\pm$ 1.13 & 75.69\% $\pm$ 0.54 & 79.68\% $\pm$ 0.73 & 81.74\% $\pm$ 0.63 & 80.22\% $\pm$ 0.75 & 80.99\% $\pm$ 0.62 \\
DeepSeekV3 & 71.42\% $\pm$ 0.96 & 61.57\% $\pm$ 0.66 & 46.02\% $\pm$ 0.70 & 51.04\% $\pm$ 0.69 & 45.51\% $\pm$ 0.72 & 52.52\% $\pm$ 0.68 \\
GPT-4o Mini & 33.67\% $\pm$ 1.50 & 74.80\% $\pm$ 0.65 & 77.26\% $\pm$ 1.00 & 79.57\% $\pm$ 0.86 & 77.73\% $\pm$ 1.03 & 78.67\% $\pm$ 0.85 \\
GPT-4o & 24.78\% $\pm$ 0.99 & 73.74\% $\pm$ 0.58 & 82.08\% $\pm$ 0.65 & 83.38\% $\pm$ 0.57 & 83.19\% $\pm$ 0.65 & 82.15\% $\pm$ 0.57 \\
\bottomrule
\end{tabular}
}
\resizebox{\textwidth}{!}{
\begin{tabular}{lcccccc}
\toprule
 & Qwen 72B & Llama 8B & Llama 70B & DeepSeekV3 & GPT-4o Mini & GPT-4o \\
\midrule
Qwen 0.5B & 28.16\% $\pm$ 1.11 & 32.93\% $\pm$ 1.13 & 42.30\% $\pm$ 1.26 & 24.48\% $\pm$ 1.03 & 12.76\% $\pm$ 0.72 & 26.42\% $\pm$ 1.10 \\
Qwen 1.5B & 47.97\% $\pm$ 0.88 & 51.83\% $\pm$ 0.85 & 60.81\% $\pm$ 0.87 & 44.47\% $\pm$ 0.85 & 33.12\% $\pm$ 0.83 & 46.36\% $\pm$ 0.86 \\
Qwen 3B & 81.06\% $\pm$ 0.67 & 79.56\% $\pm$ 0.60 & 75.97\% $\pm$ 0.59 & 81.51\% $\pm$ 0.72 & 79.82\% $\pm$ 0.94 & 82.56\% $\pm$ 0.67 \\
Qwen 7B & 82.14\% $\pm$ 0.57 & 79.53\% $\pm$ 0.56 & 74.39\% $\pm$ 0.62 & 83.15\% $\pm$ 0.59 & 83.57\% $\pm$ 0.73 & 84.08\% $\pm$ 0.56 \\
Qwen 14B & 82.17\% $\pm$ 0.57 & 79.10\% $\pm$ 0.57 & 74.03\% $\pm$ 0.62 & 83.07\% $\pm$ 0.59 & 83.63\% $\pm$ 0.73 & 83.75\% $\pm$ 0.57 \\
Qwen 32B & 82.30\% $\pm$ 0.57 & 78.96\% $\pm$ 0.57 & 74.62\% $\pm$ 0.60 & 82.89\% $\pm$ 0.59 & 82.48\% $\pm$ 0.74 & 83.70\% $\pm$ 0.57 \\
Qwen 72B & \textbf{83.44\% $\pm$ 0.54} & 79.96\% $\pm$ 0.54 & 75.86\% $\pm$ 0.56 & 83.53\% $\pm$ 0.58 & 82.33\% $\pm$ 0.76 & 84.29\% $\pm$ 0.55 \\
Llama 8B & 81.87\% $\pm$ 0.57 & \textbf{81.41\% $\pm$ 0.49} & 75.59\% $\pm$ 0.57 & 82.65\% $\pm$ 0.58 & 82.34\% $\pm$ 0.73 & 83.43\% $\pm$ 0.56 \\
Llama 70B & 81.66\% $\pm$ 0.59 & 80.40\% $\pm$ 0.53 & \textbf{78.79\% $\pm$ 0.49} & 81.72\% $\pm$ 0.64 & 78.57\% $\pm$ 0.86 & 83.04\% $\pm$ 0.59 \\
DeepSeekV3 & 54.46\% $\pm$ 0.71 & 56.81\% $\pm$ 0.66 & 64.57\% $\pm$ 0.67 & \textbf{51.67\% $\pm$ 0.71} & 40.64\% $\pm$ 0.76 & 53.12\% $\pm$ 0.69 \\
GPT-4o Mini & 79.51\% $\pm$ 0.82 & 78.58\% $\pm$ 0.71 & 76.89\% $\pm$ 0.64 & 79.11\% $\pm$ 0.89 & \textbf{76.38\% $\pm$ 1.18} & 80.69\% $\pm$ 0.82 \\
GPT-4o & 82.18\% $\pm$ 0.56 & 79.33\% $\pm$ 0.56 & 74.98\% $\pm$ 0.59 & 83.01\% $\pm$ 0.58 & 82.45\% $\pm$ 0.75 & \textbf{84.63\% $\pm$ 0.54} \\
\bottomrule
\end{tabular}
}
\end{table*}

%% file: figures/mmlu_fidelity.tex
\begin{figure*}[ht]
    \centering
    \resizebox{\linewidth}{!}{
        \includegraphics[scale=0.5]{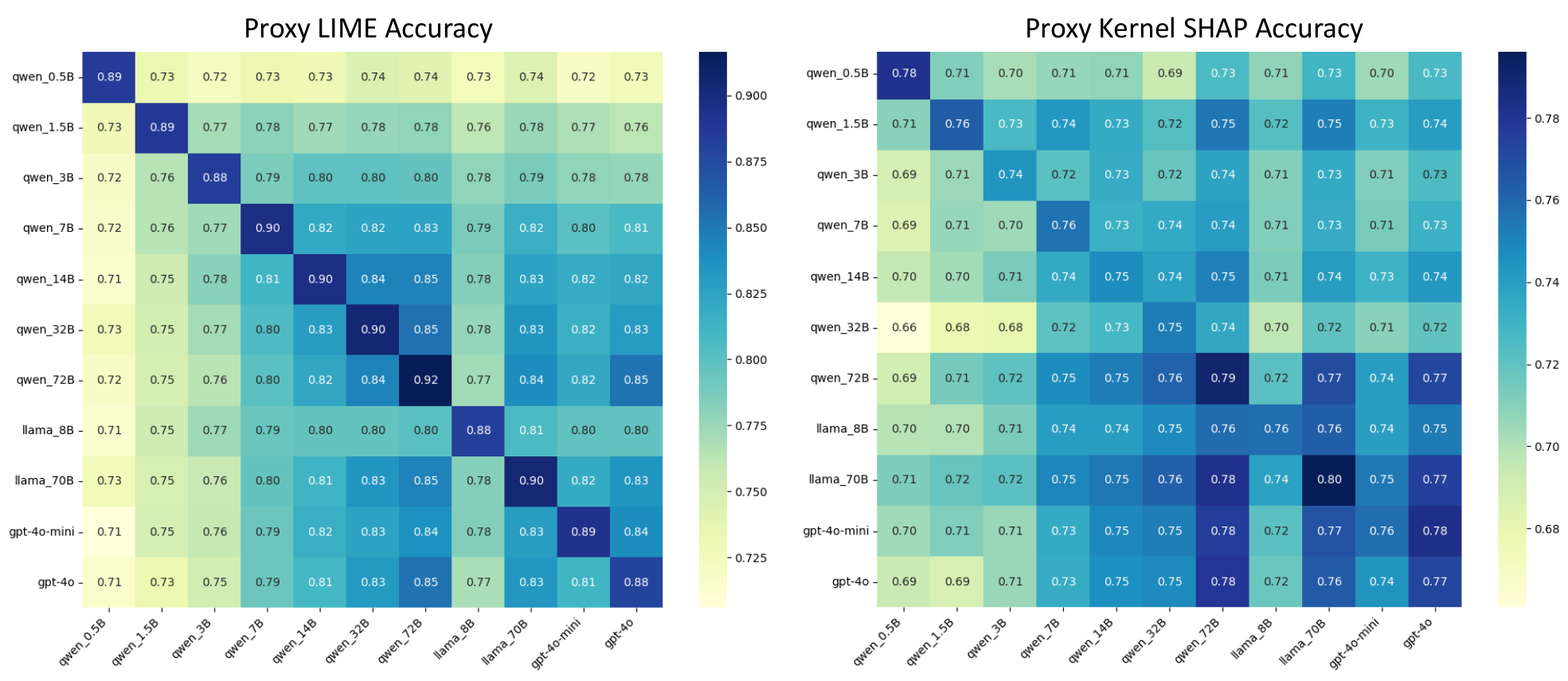}
    }
    \caption{
    Accuracy of LIME proxy explanations on the multiple-choice question answering task. Each cell shows how well explanations generated by the model on the \textbf{y-axis} serve as surrogates for predicting the behavior of the model on the \textbf{x-axis}. 
    The heatmap on the right shows results after filtering out examples where the budget-friendly and expensive models produce different predictions for the input.
    }
    \label{fig:mmlu_fidelity}
\end{figure*}

%% file: figures/mmlu_shap_fidelity.tex
\begin{figure*}[t]
    \centering
    \resizebox{\linewidth}{!}{
        \includegraphics{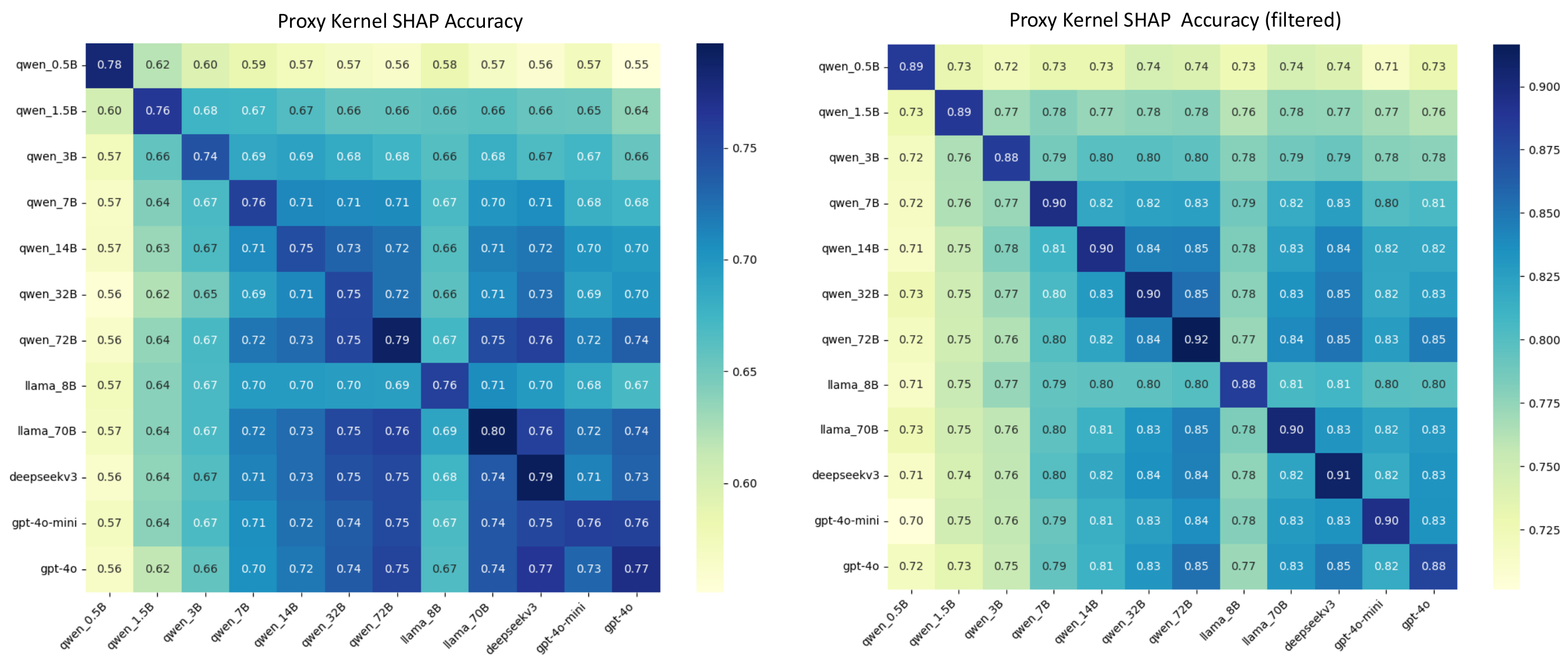}
    }
    \caption{
    Accuracy of Kernel SHAP proxy explanations on the multiple-choice question answering task. Each cell shows how well explanations generated by the model on the \textbf{y-axis} serve as surrogates for predicting the behavior of the model on the \textbf{x-axis}. 
    The heatmap on the right shows results after filtering out examples where the budget-friendly and expensive models produce different predictions for the input.
    }
    \label{fig:mmlu_shap_fidelity}
\end{figure*}

%% file: figures/mmlu_subject_lime_fidelity.tex
\begin{figure*}[t]
    \centering
    \resizebox{0.69\linewidth}{!}{
        \includegraphics{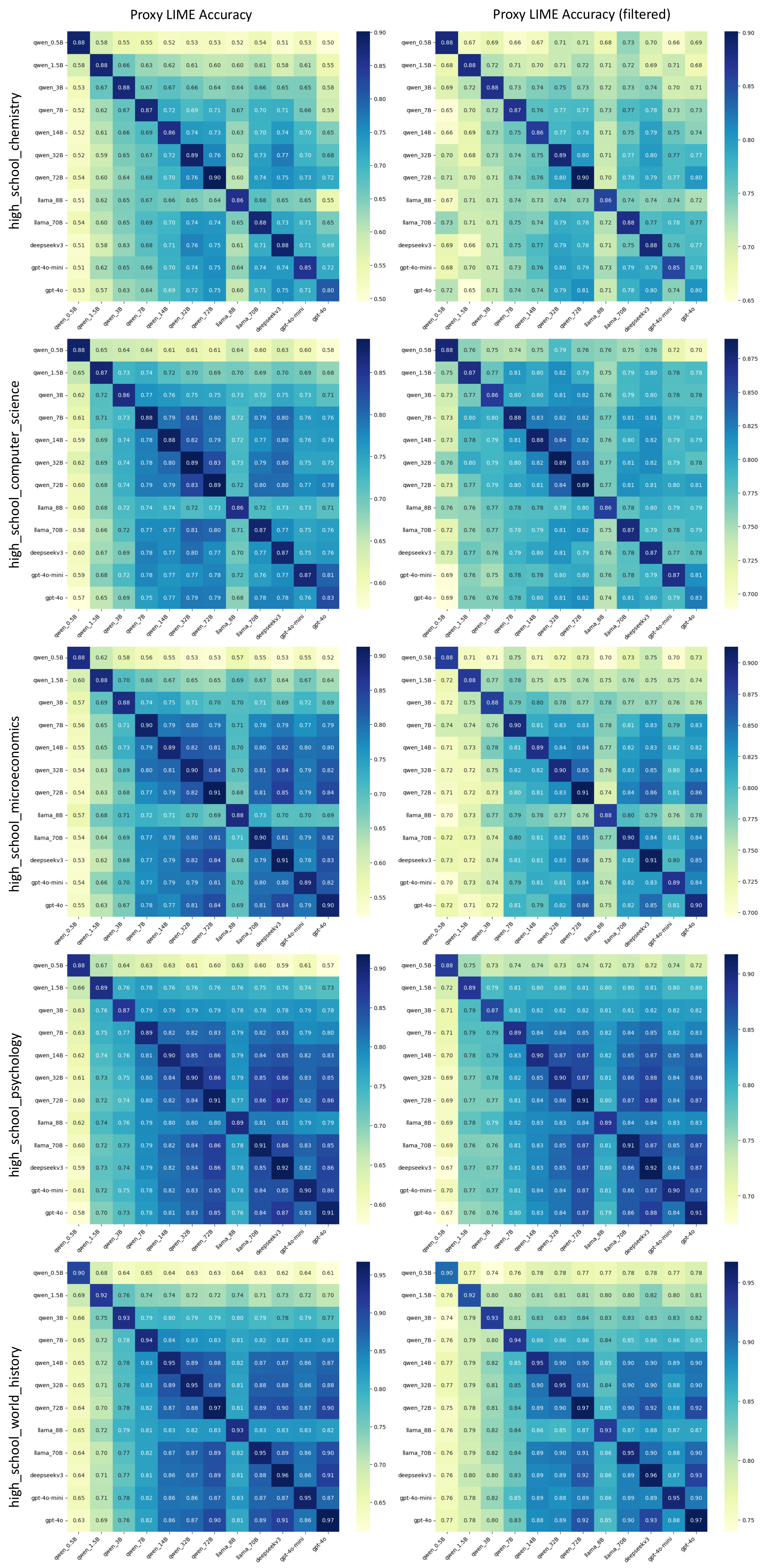}
    }
    \caption{
    Accuracy of LIME proxy explanations on the multiple-choice question answering task on each subject. Each cell shows how well explanations generated by the model on the \textbf{y-axis} serve as surrogates for predicting the behavior of the model on the \textbf{x-axis}. 
    The heatmap on the right shows results after filtering out examples where the budget-friendly and expensive models produce different predictions for the input.
    }
    \label{fig:mmlu_fidelity_lime_subject}
\end{figure*}

%% file: figures/mmlu_subject_shap_fidelity.tex
\begin{figure*}[t]
    \centering
    \resizebox{0.69\linewidth}{!}{
        \includegraphics{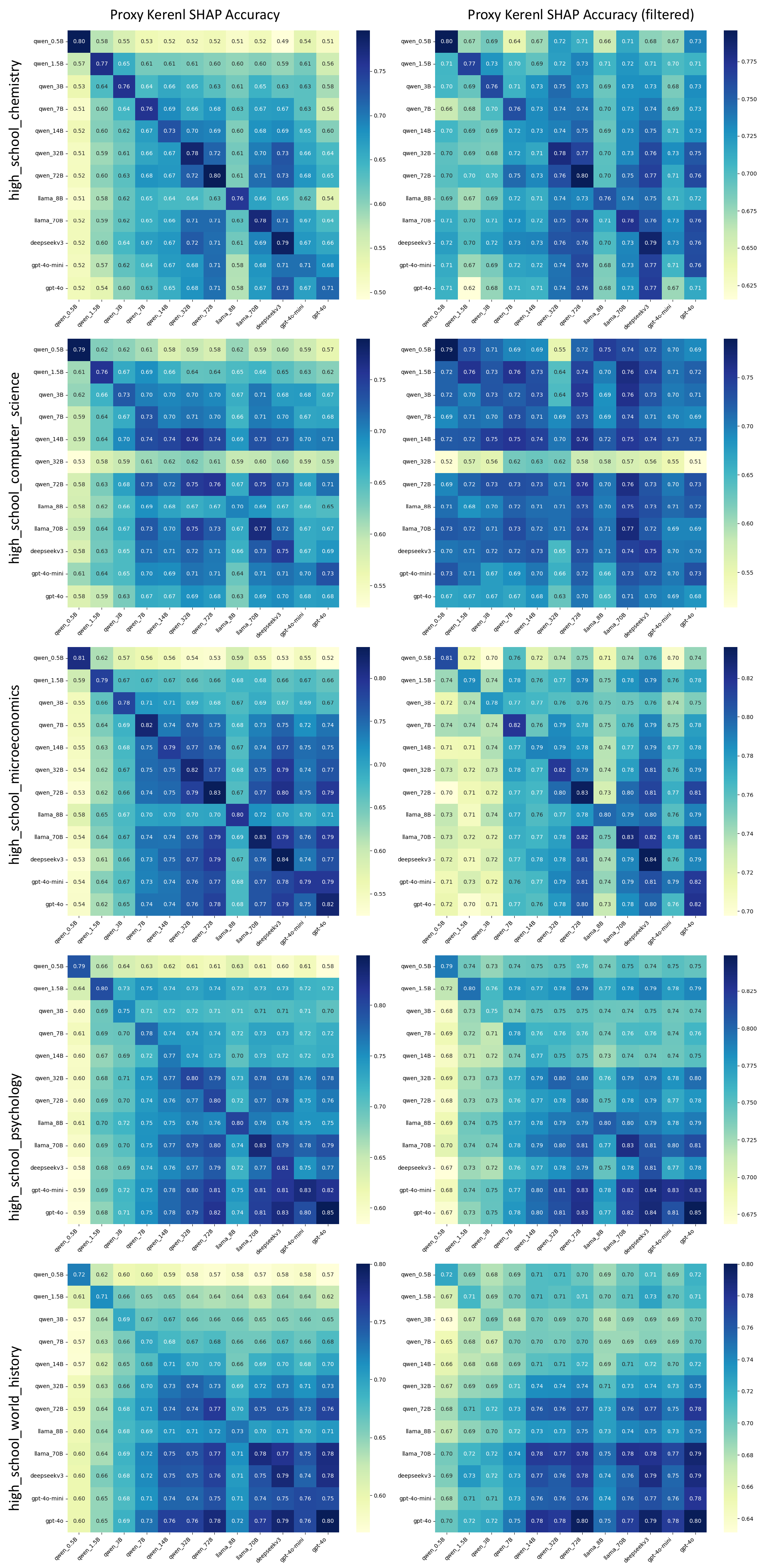}
    }
    \caption{
    Accuracy of Kernel SHAP proxy explanations on the multiple-choice question answering task on each subject. Each cell shows how well explanations generated by the model on the \textbf{y-axis} serve as surrogates for predicting the behavior of the model on the \textbf{x-axis}. 
    The heatmap on the right shows results after filtering out examples where the budget-friendly and expensive models produce different predictions for the input.
    }
    \label{fig:mmlu_fidelity_shap_subject}
\end{figure*}

%% file: tables/mmlu_lime_fidelity.tex
\begin{table*}
\centering
\caption{Accuracy of proxy LIME explanations on high school chemistriy of MMLU datasets: each value shows how well LIME explanations generated by the model on the \textbf{left} serve as surrogates for predicting the behavior of the model on the \textbf{top}.}
\resizebox{\textwidth}{!}{
\begin{tabular}{lcccccc}
\toprule
 & Qwen 0.5B & Qwen 1.5B & Qwen 3B & Qwen 7B & Qwen 14B & Qwen 32B \\
\midrule
Qwen 0.5B & \textbf{88.40\% $\pm$ 1.21} & 57.69\% $\pm$ 3.96 & 54.85\% $\pm$ 4.04 & 54.68\% $\pm$ 4.00 & 52.45\% $\pm$ 3.82 & 52.60\% $\pm$ 4.07 \\
Qwen 1.5B & 57.69\% $\pm$ 3.72 & \textbf{87.85\% $\pm$ 1.19} & 66.16\% $\pm$ 3.41 & 62.99\% $\pm$ 3.57 & 62.17\% $\pm$ 3.46 & 60.66\% $\pm$ 3.84 \\
Qwen 3B & 53.33\% $\pm$ 4.11 & 66.66\% $\pm$ 3.28 & \textbf{87.50\% $\pm$ 1.18} & 66.64\% $\pm$ 3.30 & 67.46\% $\pm$ 3.23 & 66.13\% $\pm$ 3.72 \\
Qwen 7B & 51.98\% $\pm$ 4.20 & 62.08\% $\pm$ 3.75 & 66.84\% $\pm$ 3.50 & \textbf{87.32\% $\pm$ 1.20} & 71.58\% $\pm$ 3.05 & 68.92\% $\pm$ 3.59 \\
Qwen 14B & 51.61\% $\pm$ 4.03 & 60.55\% $\pm$ 3.65 & 66.29\% $\pm$ 3.39 & 68.77\% $\pm$ 3.27 & \textbf{86.26\% $\pm$ 1.28} & 73.70\% $\pm$ 3.08 \\
Qwen 32B & 52.17\% $\pm$ 4.19 & 58.59\% $\pm$ 3.99 & 65.03\% $\pm$ 3.61 & 66.96\% $\pm$ 3.42 & 72.10\% $\pm$ 3.02 & \textbf{88.61\% $\pm$ 1.20} \\
Qwen 72B & 53.53\% $\pm$ 4.19 & 59.96\% $\pm$ 3.79 & 63.63\% $\pm$ 3.65 & 67.55\% $\pm$ 3.39 & 69.91\% $\pm$ 3.09 & 75.94\% $\pm$ 3.11 \\
Llama 8B & 50.80\% $\pm$ 3.93 & 61.74\% $\pm$ 3.72 & 64.80\% $\pm$ 3.29 & 66.87\% $\pm$ 3.31 & 65.83\% $\pm$ 3.20 & 64.98\% $\pm$ 3.59 \\
Llama 70B & 53.71\% $\pm$ 4.05 & 60.43\% $\pm$ 3.89 & 64.73\% $\pm$ 3.62 & 68.63\% $\pm$ 3.25 & 69.92\% $\pm$ 3.13 & 74.35\% $\pm$ 3.07 \\
DeepSeekV3 & 51.13\% $\pm$ 4.21 & 58.08\% $\pm$ 3.92 & 63.28\% $\pm$ 3.56 & 67.76\% $\pm$ 3.37 & 70.82\% $\pm$ 3.05 & 76.40\% $\pm$ 2.83 \\
GPT-4o Mini & 51.46\% $\pm$ 4.05 & 62.03\% $\pm$ 3.75 & 65.46\% $\pm$ 3.45 & 65.75\% $\pm$ 3.47 & 70.48\% $\pm$ 3.03 & 73.60\% $\pm$ 3.10 \\
GPT-4o & 52.82\% $\pm$ 4.21 & 57.45\% $\pm$ 3.77 & 62.72\% $\pm$ 3.68 & 64.12\% $\pm$ 3.53 & 68.97\% $\pm$ 3.08 & 72.44\% $\pm$ 3.33 \\
\bottomrule
\end{tabular}
}
\resizebox{\textwidth}{!}{
\begin{tabular}{lcccccc}
\toprule
 & Qwen 72B & Llama 8B & Llama 70B & DeepSeekV3 & GPT-4o Mini & GPT-4o \\
\midrule
Qwen 0.5B & 52.58\% $\pm$ 4.19 & 51.61\% $\pm$ 3.77 & 53.97\% $\pm$ 3.96 & 50.99\% $\pm$ 4.16 & 52.96\% $\pm$ 3.97 & 49.53\% $\pm$ 4.38 \\
Qwen 1.5B & 59.95\% $\pm$ 3.88 & 60.28\% $\pm$ 3.54 & 61.35\% $\pm$ 3.69 & 58.04\% $\pm$ 3.91 & 60.91\% $\pm$ 3.74 & 54.69\% $\pm$ 4.35 \\
Qwen 3B & 63.84\% $\pm$ 3.84 & 64.41\% $\pm$ 3.27 & 66.42\% $\pm$ 3.48 & 64.77\% $\pm$ 3.63 & 64.92\% $\pm$ 3.57 & 57.94\% $\pm$ 4.43 \\
Qwen 7B & 70.98\% $\pm$ 3.49 & 66.58\% $\pm$ 3.29 & 70.46\% $\pm$ 3.36 & 70.82\% $\pm$ 3.30 & 65.65\% $\pm$ 3.76 & 58.59\% $\pm$ 4.68 \\ 
Qwen 14B & 73.27\% $\pm$ 3.13 & 63.13\% $\pm$ 3.46 & 70.42\% $\pm$ 3.20 & 73.70\% $\pm$ 2.96 & 69.61\% $\pm$ 3.24 & 64.58\% $\pm$ 3.94 \\ 
Qwen 32B & 76.10\% $\pm$ 3.20 & 61.95\% $\pm$ 3.55 & 73.08\% $\pm$ 3.20 & 77.07\% $\pm$ 2.80 & 70.42\% $\pm$ 3.46 & 67.63\% $\pm$ 4.15 \\ 
Qwen 72B & \textbf{90.10\% $\pm$ 1.10} & 60.33\% $\pm$ 3.66 & 73.63\% $\pm$ 3.09 & 75.14\% $\pm$ 3.08 & 73.16\% $\pm$ 3.14 & 71.58\% $\pm$ 3.78 \\ 
Llama 8B & 63.84\% $\pm$ 3.67 & \textbf{85.70\% $\pm$ 1.20} & 67.60\% $\pm$ 3.33 & 65.18\% $\pm$ 3.48 & 64.61\% $\pm$ 3.58 & 55.19\% $\pm$ 4.38 \\ 
Llama 70B & 74.00\% $\pm$ 3.09 & 64.75\% $\pm$ 3.37 & \textbf{87.69\% $\pm$ 1.18} & 72.60\% $\pm$ 3.19 & 70.56\% $\pm$ 3.39 & 65.14\% $\pm$ 4.37 \\ 
DeepSeekV3 & 74.51\% $\pm$ 3.27 & 61.24\% $\pm$ 3.64 & 71.39\% $\pm$ 3.24 & \textbf{88.42\% $\pm$ 1.25} & 70.83\% $\pm$ 3.33 & 68.90\% $\pm$ 3.99 \\ 
GPT-4o Mini & 75.06\% $\pm$ 3.10 & 63.59\% $\pm$ 3.49 & 73.85\% $\pm$ 2.97 & 73.68\% $\pm$ 3.09 & \textbf{85.47\% $\pm$ 1.73} & 71.61\% $\pm$ 3.48 \\ 
GPT-4o & 74.68\% $\pm$ 3.23 & 59.56\% $\pm$ 3.63 & 70.98\% $\pm$ 3.30 & 75.32\% $\pm$ 3.00 & 71.39\% $\pm$ 3.16 & \textbf{80.39\% $\pm$ 2.61} \\
\bottomrule
\end{tabular}
}
\label{tab:mmlu_lime_chemistriy_fidelity}
\end{table*}

\begin{table*}
\centering
\caption{Accuracy of proxy \textbf{filtered} LIME explanations on high school chemistry of MMLU datasets: each value shows how well LIME explanations generated by the model on the \textbf{left} serve as surrogates for predicting the behavior of the model on the \textbf{top}.}
\resizebox{\textwidth}{!}{
\begin{tabular}{lcccccc}
\toprule
 & Qwen 0.5B & Qwen 1.5B & Qwen 3B & Qwen 7B & Qwen 14B & Qwen 32B \\
\midrule
Qwen 0.5B & \textbf{88.40\% $\pm$ 1.21} & 66.96\% $\pm$ 5.07 & 68.54\% $\pm$ 5.25 & 66.02\% $\pm$ 5.54 & 66.96\% $\pm$ 6.10 & 70.93\% $\pm$ 5.80 \\
Qwen 1.5B & 68.20\% $\pm$ 4.93 & \textbf{87.85\% $\pm$ 1.19} & 72.08\% $\pm$ 3.90 & 70.61\% $\pm$ 4.30 & 69.95\% $\pm$ 4.25 & 70.82\% $\pm$ 4.89 \\
Qwen 3B & 69.24\% $\pm$ 5.44 & 71.96\% $\pm$ 4.18 & \textbf{87.50\% $\pm$ 1.18} & 72.88\% $\pm$ 3.84 & 74.31\% $\pm$ 3.90 & 75.15\% $\pm$ 4.18 \\
Qwen 7B & 65.19\% $\pm$ 5.96 & 69.80\% $\pm$ 4.61 & 72.27\% $\pm$ 3.87 & \textbf{87.32\% $\pm$ 1.20} & 76.26\% $\pm$ 3.43 & 76.91\% $\pm$ 3.71 \\
Qwen 14B & 66.06\% $\pm$ 6.58 & 68.60\% $\pm$ 4.57 & 73.32\% $\pm$ 3.84 & 75.29\% $\pm$ 3.47 & \textbf{86.26\% $\pm$ 1.28} & 77.33\% $\pm$ 3.27 \\
Qwen 32B & 70.37\% $\pm$ 6.29 & 68.36\% $\pm$ 5.19 & 72.64\% $\pm$ 4.19 & 74.01\% $\pm$ 3.68 & 75.40\% $\pm$ 3.27 & \textbf{88.61\% $\pm$ 1.20} \\
Qwen 72B & 71.14\% $\pm$ 5.77 & 69.95\% $\pm$ 4.55 & 71.34\% $\pm$ 4.17 & 74.25\% $\pm$ 3.53 & 75.85\% $\pm$ 3.22 & 79.72\% $\pm$ 3.06 \\
Llama 8B & 66.86\% $\pm$ 6.08 & 70.93\% $\pm$ 4.59 & 71.45\% $\pm$ 3.90 & 73.62\% $\pm$ 3.67 & 73.00\% $\pm$ 3.51 & 73.83\% $\pm$ 3.68 \\
Llama 70B & 73.26\% $\pm$ 5.44 & 70.58\% $\pm$ 4.71 & 70.97\% $\pm$ 4.28 & 74.65\% $\pm$ 3.81 & 74.30\% $\pm$ 3.56 & 78.56\% $\pm$ 2.97 \\
DeepSeekV3 & 68.55\% $\pm$ 6.21 & 66.40\% $\pm$ 5.09 & 71.05\% $\pm$ 4.44 & 75.06\% $\pm$ 3.47 & 76.81\% $\pm$ 3.26 & 79.35\% $\pm$ 2.76 \\
GPT-4o Mini & 68.20\% $\pm$ 5.87 & 70.35\% $\pm$ 4.74 & 71.08\% $\pm$ 3.81 & 73.10\% $\pm$ 4.00 & 76.30\% $\pm$ 3.46 & 79.71\% $\pm$ 3.14 \\
GPT-4o & 71.88\% $\pm$ 5.40 & 64.87\% $\pm$ 4.72 & 71.27\% $\pm$ 4.24 & 73.99\% $\pm$ 3.63 & 74.42\% $\pm$ 3.52 & 79.07\% $\pm$ 3.19 \\
\bottomrule
\end{tabular}
}
\resizebox{\textwidth}{!}{
\begin{tabular}{lcccccc}
\toprule
 & Qwen 72B & Llama 8B & Llama 70B & DeepSeekV3 & GPT-4o Mini & GPT-4o \\
\midrule
Qwen 0.5B & 70.60\% $\pm$ 5.61 & 67.99\% $\pm$ 5.86 & 73.40\% $\pm$ 5.29 & 69.92\% $\pm$ 5.95 & 66.09\% $\pm$ 5.76 & 69.36\% $\pm$ 5.97 \\
Qwen 1.5B & 72.49\% $\pm$ 4.45 & 70.68\% $\pm$ 4.27 & 72.49\% $\pm$ 4.44 & 69.08\% $\pm$ 4.80 & 71.08\% $\pm$ 4.65 & 67.79\% $\pm$ 4.84 \\
Qwen 3B & 73.28\% $\pm$ 4.29 & 71.73\% $\pm$ 3.75 & 73.36\% $\pm$ 4.24 & 73.85\% $\pm$ 4.31 & 70.25\% $\pm$ 4.09 & 70.51\% $\pm$ 4.75 \\
Qwen 7B & 77.16\% $\pm$ 3.65 & 72.95\% $\pm$ 3.77 & 76.81\% $\pm$ 3.75 & 77.72\% $\pm$ 3.55 & 72.81\% $\pm$ 4.21 & 73.42\% $\pm$ 4.40 \\
Qwen 14B & 78.43\% $\pm$ 3.17 & 71.18\% $\pm$ 3.60 & 75.27\% $\pm$ 3.53 & 79.44\% $\pm$ 3.17 & 75.43\% $\pm$ 3.74 & 73.76\% $\pm$ 4.00 \\
Qwen 32B & 80.01\% $\pm$ 3.12 & 70.98\% $\pm$ 3.73 & 77.16\% $\pm$ 3.01 & 79.91\% $\pm$ 2.80 & 76.83\% $\pm$ 3.39 & 77.47\% $\pm$ 3.48 \\
Qwen 72B & \textbf{90.10\% $\pm$ 1.10} & 70.20\% $\pm$ 3.66 & 77.73\% $\pm$ 3.11 & 79.11\% $\pm$ 3.07 & 76.91\% $\pm$ 3.34 & 80.03\% $\pm$ 3.01 \\
Llama 8B & 73.49\% $\pm$ 3.76 & \textbf{85.70\% $\pm$ 1.20} & 74.47\% $\pm$ 3.63 & 73.75\% $\pm$ 3.66 & 73.98\% $\pm$ 3.72 & 72.32\% $\pm$ 3.99 \\
Llama 70B & 78.24\% $\pm$ 3.04 & 71.99\% $\pm$ 3.72 & \textbf{87.69\% $\pm$ 1.18} & 76.71\% $\pm$ 3.36 & 77.53\% $\pm$ 3.22 & 76.71\% $\pm$ 3.66 \\
DeepSeekV3 & 78.48\% $\pm$ 3.25 & 70.90\% $\pm$ 3.66 & 75.12\% $\pm$ 3.35 & \textbf{88.42\% $\pm$ 1.25} & 76.31\% $\pm$ 3.43 & 77.03\% $\pm$ 3.40 \\
GPT-4o Mini & 79.15\% $\pm$ 3.22 & 72.64\% $\pm$ 3.70 & 79.25\% $\pm$ 2.92 & 78.83\% $\pm$ 3.34 & \textbf{85.47\% $\pm$ 1.73} & 77.58\% $\pm$ 3.36 \\
GPT-4o & 80.69\% $\pm$ 2.98 & 70.96\% $\pm$ 3.82 & 78.01\% $\pm$ 3.05 & 80.09\% $\pm$ 2.90 & 73.96\% $\pm$ 3.47 & \textbf{80.39\% $\pm$ 2.61} \\
\bottomrule
\end{tabular}
}
\label{tab:mmlu_lime_chemistriy_filtered_fidelity}
\end{table*}

\begin{table*}
\centering
\caption{Accuracy of proxy LIME explanations on high school computer science of MMLU datasets: each value shows how well LIME explanations generated by the model on the \textbf{left} serve as surrogates for predicting the behavior of the model on the \textbf{top}.}
\resizebox{\textwidth}{!}{
\begin{tabular}{lcccccc}
\toprule
 & Qwen 0.5B & Qwen 1.5B & Qwen 3B & Qwen 7B & Qwen 14B & Qwen 32B \\
\midrule
Qwen 0.5B & \textbf{87.75\% $\pm$ 1.77} & 64.64\% $\pm$ 5.28 & 64.23\% $\pm$ 4.75 & 63.54\% $\pm$ 5.23 & 61.44\% $\pm$ 5.29 & 61.47\% $\pm$ 5.26 \\
Qwen 1.5B & 65.08\% $\pm$ 5.40 & \textbf{87.13\% $\pm$ 1.75} & 73.00\% $\pm$ 4.29 & 73.94\% $\pm$ 4.41 & 71.53\% $\pm$ 4.56 & 70.29\% $\pm$ 4.87 \\
Qwen 3B & 62.31\% $\pm$ 5.32 & 71.57\% $\pm$ 4.48 & \textbf{86.11\% $\pm$ 1.81} & 76.87\% $\pm$ 3.56 & 76.13\% $\pm$ 3.51 & 75.05\% $\pm$ 4.09 \\
Qwen 7B & 61.25\% $\pm$ 5.58 & 71.35\% $\pm$ 4.54 & 72.84\% $\pm$ 4.20 & \textbf{88.08\% $\pm$ 1.71} & 79.37\% $\pm$ 3.53 & 81.29\% $\pm$ 3.20 \\
Qwen 14B & 59.28\% $\pm$ 5.70 & 68.95\% $\pm$ 4.72 & 74.12\% $\pm$ 3.50 & 78.41\% $\pm$ 3.68 & \textbf{88.21\% $\pm$ 1.73} & 81.78\% $\pm$ 3.56 \\
Qwen 32B & 62.31\% $\pm$ 5.70 & 69.15\% $\pm$ 5.00 & 73.90\% $\pm$ 3.69 & 78.19\% $\pm$ 3.69 & 80.13\% $\pm$ 3.65 & \textbf{88.88\% $\pm$ 1.76} \\
Qwen 72B & 60.20\% $\pm$ 5.70 & 68.31\% $\pm$ 4.99 & 73.58\% $\pm$ 3.78 & 78.79\% $\pm$ 3.39 & 78.75\% $\pm$ 3.71 & 83.34\% $\pm$ 2.98 \\
Llama 8B & 59.99\% $\pm$ 5.54 & 68.22\% $\pm$ 4.68 & 72.18\% $\pm$ 3.75 & 73.98\% $\pm$ 4.23 & 73.72\% $\pm$ 4.14 & 71.93\% $\pm$ 4.70 \\
Llama 70B & 58.14\% $\pm$ 5.65 & 66.28\% $\pm$ 5.00 & 71.51\% $\pm$ 4.37 & 77.35\% $\pm$ 3.70 & 77.30\% $\pm$ 4.02 & 80.73\% $\pm$ 3.51 \\
DeepSeekV3 & 59.53\% $\pm$ 5.73 & 66.98\% $\pm$ 5.06 & 69.21\% $\pm$ 4.49 & 78.17\% $\pm$ 3.44 & 77.44\% $\pm$ 3.88 & 79.76\% $\pm$ 3.66 \\
GPT-4o Mini & 59.05\% $\pm$ 5.77 & 68.49\% $\pm$ 4.76 & 72.46\% $\pm$ 4.12 & 77.73\% $\pm$ 3.65 & 76.53\% $\pm$ 4.06 & 77.04\% $\pm$ 4.18 \\
GPT-4o & 56.87\% $\pm$ 5.83 & 65.23\% $\pm$ 5.27 & 68.73\% $\pm$ 4.49 & 75.43\% $\pm$ 3.69 & 77.01\% $\pm$ 3.86 & 78.82\% $\pm$ 3.61 \\
\bottomrule
\end{tabular}
}

\quad
\resizebox{\textwidth}{!}{
\begin{tabular}{lcccccc}
\toprule
 & Qwen 72B & Llama 8B & Llama 70B & DeepSeekV3 & GPT-4o Mini & GPT-4o \\
\midrule
Qwen 0.5B & 60.62\% $\pm$ 5.42 & 64.07\% $\pm$ 4.89 & 60.28\% $\pm$ 5.42 & 62.83\% $\pm$ 5.49 & 60.11\% $\pm$ 5.55 & 57.56\% $\pm$ 5.88 \\
Qwen 1.5B & 69.01\% $\pm$ 4.93 & 69.60\% $\pm$ 4.66 & 68.53\% $\pm$ 5.05 & 70.21\% $\pm$ 4.99 & 69.42\% $\pm$ 4.95 & 67.86\% $\pm$ 5.52 \\
Qwen 3B & 74.92\% $\pm$ 3.94 & 73.27\% $\pm$ 3.85 & 72.48\% $\pm$ 4.47 & 74.65\% $\pm$ 4.14 & 73.18\% $\pm$ 4.34 & 71.19\% $\pm$ 4.95 \\
Qwen 7B & 79.54\% $\pm$ 3.71 & 72.38\% $\pm$ 4.34 & 78.66\% $\pm$ 3.73 & 80.10\% $\pm$ 3.56 & 76.07\% $\pm$ 4.16 & 76.38\% $\pm$ 4.24 \\
Qwen 14B & 79.38\% $\pm$ 3.85 & 72.38\% $\pm$ 4.14 & 77.14\% $\pm$ 4.07 & 79.62\% $\pm$ 3.84 & 75.57\% $\pm$ 4.47 & 76.00\% $\pm$ 4.78 \\
Qwen 32B & 82.74\% $\pm$ 3.23 & 72.57\% $\pm$ 4.52 & 78.99\% $\pm$ 3.98 & 80.33\% $\pm$ 3.87 & 74.80\% $\pm$ 4.71 & 74.89\% $\pm$ 4.75 \\
Qwen 72B & \textbf{88.95\% $\pm$ 1.67} & 71.99\% $\pm$ 4.36 & 79.98\% $\pm$ 3.45 & 79.87\% $\pm$ 4.03 & 77.39\% $\pm$ 4.09 & 78.00\% $\pm$ 4.29 \\
Llama 8B & 73.12\% $\pm$ 4.52 & \textbf{85.79\% $\pm$ 1.75} & 72.08\% $\pm$ 4.73 & 73.19\% $\pm$ 4.55 & 73.25\% $\pm$ 4.47 & 71.36\% $\pm$ 5.12 \\
Llama 70B & 80.27\% $\pm$ 3.58 & 70.69\% $\pm$ 4.71 & \textbf{87.45\% $\pm$ 2.10} & 77.40\% $\pm$ 4.43 & 74.81\% $\pm$ 4.53 & 75.62\% $\pm$ 4.76 \\
DeepSeekV3 & 77.38\% $\pm$ 4.29 & 69.87\% $\pm$ 4.60 & 76.71\% $\pm$ 4.28 & \textbf{87.32\% $\pm$ 2.21} & 75.41\% $\pm$ 4.66 & 76.09\% $\pm$ 4.56 \\
GPT-4o Mini & 78.12\% $\pm$ 4.03 & 71.51\% $\pm$ 4.57 & 75.75\% $\pm$ 4.20 & 77.36\% $\pm$ 4.45 & \textbf{87.03\% $\pm$ 2.25} & 80.57\% $\pm$ 4.04 \\
GPT-4o & 78.76\% $\pm$ 3.66 & 68.20\% $\pm$ 4.56 & 78.22\% $\pm$ 3.46 & 78.18\% $\pm$ 3.86 & 76.11\% $\pm$ 4.20 & \textbf{82.58\% $\pm$ 3.51} \\
\bottomrule
\end{tabular}
}
\label{tab:mmlu_lime_computer_science_fidelity}
\end{table*}

\begin{table*}
\centering
\resizebox{\textwidth}{!}{
\begin{tabular}{lcccccc}
\toprule
 & Qwen 0.5B & Qwen 1.5B & Qwen 3B & Qwen 7B & Qwen 14B & Qwen 32B \\
\midrule
Qwen 0.5B & \textbf{87.75\% $\pm$ 1.77} & 75.97\% $\pm$ 5.91 & 74.61\% $\pm$ 5.48 & 74.28\% $\pm$ 6.82 & 75.35\% $\pm$ 6.90 & 78.50\% $\pm$ 6.19 \\
Qwen 1.5B & 75.29\% $\pm$ 5.73 & \textbf{87.13\% $\pm$ 1.75} & 77.29\% $\pm$ 4.67 & 80.88\% $\pm$ 4.18 & 79.97\% $\pm$ 4.45 & 81.58\% $\pm$ 4.05 \\
Qwen 3B & 73.31\% $\pm$ 5.61 & 76.74\% $\pm$ 4.37 & \textbf{86.11\% $\pm$ 1.81} & 80.15\% $\pm$ 3.70 & 80.01\% $\pm$ 3.81 & 81.41\% $\pm$ 3.90 \\
Qwen 7B & 72.53\% $\pm$ 6.77 & 80.23\% $\pm$ 3.89 & 79.53\% $\pm$ 3.88 & \textbf{88.08\% $\pm$ 1.71} & 82.60\% $\pm$ 3.49 & 82.34\% $\pm$ 3.39 \\
Qwen 14B & 72.79\% $\pm$ 7.00 & 78.21\% $\pm$ 4.43 & 78.82\% $\pm$ 3.96 & 81.21\% $\pm$ 3.69 & \textbf{88.21\% $\pm$ 1.73} & 84.08\% $\pm$ 3.26 \\
Qwen 32B & 75.92\% $\pm$ 6.52 & 79.77\% $\pm$ 4.14 & 78.79\% $\pm$ 4.17 & 79.52\% $\pm$ 3.72 & 82.13\% $\pm$ 3.48 & \textbf{88.88\% $\pm$ 1.76} \\
Qwen 72B & 72.53\% $\pm$ 7.12 & 77.10\% $\pm$ 4.61 & 79.00\% $\pm$ 3.99 & 80.20\% $\pm$ 3.59 & 80.74\% $\pm$ 3.73 & 83.79\% $\pm$ 3.00 \\
Llama 8B & 75.51\% $\pm$ 6.51 & 75.91\% $\pm$ 4.53 & 77.24\% $\pm$ 3.88 & 78.13\% $\pm$ 4.38 & 78.28\% $\pm$ 4.35 & 78.50\% $\pm$ 4.63 \\
Llama 70B & 71.92\% $\pm$ 6.93 & 75.73\% $\pm$ 4.71 & 76.56\% $\pm$ 4.73 & 78.46\% $\pm$ 4.00 & 78.79\% $\pm$ 4.19 & 81.30\% $\pm$ 3.65 \\
DeepSeekV3 & 72.91\% $\pm$ 6.81 & 77.23\% $\pm$ 4.76 & 76.21\% $\pm$ 4.67 & 79.15\% $\pm$ 3.70 & 80.15\% $\pm$ 3.63 & 81.15\% $\pm$ 3.50 \\
GPT-4o Mini & 69.10\% $\pm$ 7.67 & 76.07\% $\pm$ 4.74 & 74.98\% $\pm$ 4.58 & 78.29\% $\pm$ 4.04 & 77.98\% $\pm$ 4.68 & 79.63\% $\pm$ 4.57 \\
GPT-4o & 68.86\% $\pm$ 7.57 & 76.25\% $\pm$ 4.76 & 76.19\% $\pm$ 4.36 & 78.01\% $\pm$ 3.76 & 80.03\% $\pm$ 4.01 & 81.12\% $\pm$ 3.56 \\
\bottomrule
\end{tabular}
}

\quad
\resizebox{\textwidth}{!}{
\begin{tabular}{lcccccc}
\toprule
 & Qwen 72B & Llama 8B & Llama 70B & DeepSeekV3 & GPT-4o Mini & GPT-4o \\
\midrule
Qwen 0.5B & 76.28\% $\pm$ 6.99 & 75.93\% $\pm$ 5.83 & 75.21\% $\pm$ 7.04 & 75.81\% $\pm$ 6.80 & 71.57\% $\pm$ 7.71 & 70.18\% $\pm$ 8.21 \\
Qwen 1.5B & 79.39\% $\pm$ 4.55 & 75.18\% $\pm$ 4.75 & 78.71\% $\pm$ 4.74 & 79.69\% $\pm$ 4.79 & 78.06\% $\pm$ 4.97 & 78.48\% $\pm$ 4.95 \\
Qwen 3B & 81.80\% $\pm$ 3.83 & 75.92\% $\pm$ 3.95 & 78.62\% $\pm$ 4.57 & 79.74\% $\pm$ 4.43 & 77.98\% $\pm$ 4.41 & 77.99\% $\pm$ 4.81 \\
Qwen 7B & 82.25\% $\pm$ 3.60 & 76.83\% $\pm$ 4.44 & 81.41\% $\pm$ 3.74 & 81.49\% $\pm$ 3.80 & 79.14\% $\pm$ 4.20 & 78.80\% $\pm$ 4.43 \\
Qwen 14B & 82.50\% $\pm$ 3.53 & 76.25\% $\pm$ 4.29 & 79.61\% $\pm$ 4.06 & 81.77\% $\pm$ 3.77 & 78.62\% $\pm$ 4.79 & 79.35\% $\pm$ 4.72 \\
Qwen 32B & 83.49\% $\pm$ 3.25 & 76.69\% $\pm$ 4.56 & 81.05\% $\pm$ 3.72 & 81.65\% $\pm$ 3.77 & 78.80\% $\pm$ 4.93 & 77.88\% $\pm$ 4.60 \\
Qwen 72B & \textbf{88.95\% $\pm$ 1.67} & 76.58\% $\pm$ 4.39 & 81.47\% $\pm$ 3.34 & 80.65\% $\pm$ 3.98 & 80.23\% $\pm$ 4.12 & 80.59\% $\pm$ 3.97 \\
Llama 8B & 79.71\% $\pm$ 4.31 & \textbf{85.79\% $\pm$ 1.75} & 77.65\% $\pm$ 4.83 & 79.62\% $\pm$ 4.37 & 79.11\% $\pm$ 4.31 & 78.94\% $\pm$ 4.84 \\
Llama 70B & 81.61\% $\pm$ 3.44 & 74.97\% $\pm$ 4.77 & \textbf{87.45\% $\pm$ 2.10} & 78.72\% $\pm$ 4.51 & 78.01\% $\pm$ 4.70 & 78.94\% $\pm$ 4.43 \\
DeepSeekV3 & 79.27\% $\pm$ 4.08 & 75.64\% $\pm$ 4.38 & 78.01\% $\pm$ 4.43 & \textbf{87.32\% $\pm$ 2.21} & 77.47\% $\pm$ 4.98 & 77.82\% $\pm$ 4.52 \\
GPT-4o Mini & 79.87\% $\pm$ 4.31 & 76.12\% $\pm$ 4.51 & 77.77\% $\pm$ 4.42 & 79.18\% $\pm$ 4.67 & \textbf{87.03\% $\pm$ 2.25} & 81.47\% $\pm$ 4.24 \\
GPT-4o & 81.72\% $\pm$ 3.38 & 74.17\% $\pm$ 4.64 & 80.52\% $\pm$ 3.56 & 79.88\% $\pm$ 3.92 & 78.63\% $\pm$ 4.44 & \textbf{82.58\% $\pm$ 3.51} \\
\bottomrule
\end{tabular}
}
\caption{Accuracy of proxy \textbf{filtered} LIME explanations on high school computer science of MMLU datasets: each value shows how well LIME explanations generated by the model on the \textbf{left} serve as surrogates for predicting the behavior of the model on the \textbf{top}.}
\label{tab:mmlu_lime_computer_science_filtered_fidelity}
\end{table*}

\begin{table*}
\centering
\caption{Accuracy of proxy LIME explanations on high school microeconomics of MMLU datasets: each value shows how well LIME explanations generated by the model on the \textbf{left} serve as surrogates for predicting the behavior of the model on the \textbf{top}.}
\resizebox{\textwidth}{!}{
\begin{tabular}{lcccccc}
\toprule
 & Qwen 0.5B & Qwen 1.5B & Qwen 3B & Qwen 7B & Qwen 14B & Qwen 32B \\
\midrule
Qwen 0.5B & \textbf{87.93\% $\pm$ 1.08} & 61.79\% $\pm$ 3.38 & 58.21\% $\pm$ 3.36 & 55.88\% $\pm$ 3.79 & 55.37\% $\pm$ 3.68 & 53.39\% $\pm$ 3.77 \\
Qwen 1.5B & 59.72\% $\pm$ 3.53 & \textbf{88.26\% $\pm$ 1.06} & 70.39\% $\pm$ 2.95 & 68.05\% $\pm$ 3.40 & 66.76\% $\pm$ 3.38 & 65.19\% $\pm$ 3.53 \\
Qwen 3B & 57.43\% $\pm$ 3.60 & 69.15\% $\pm$ 3.10 & \textbf{87.87\% $\pm$ 1.17} & 73.62\% $\pm$ 3.04 & 74.55\% $\pm$ 2.83 & 71.28\% $\pm$ 3.03 \\
Qwen 7B & 56.09\% $\pm$ 3.74 & 65.12\% $\pm$ 3.44 & 71.40\% $\pm$ 3.02 & \textbf{89.77\% $\pm$ 1.11} & 79.03\% $\pm$ 2.51 & 80.10\% $\pm$ 2.36 \\
Qwen 14B & 55.36\% $\pm$ 3.79 & 64.97\% $\pm$ 3.48 & 72.78\% $\pm$ 2.86 & 79.30\% $\pm$ 2.52 & \textbf{89.15\% $\pm$ 1.13} & 82.18\% $\pm$ 2.19 \\
Qwen 32B & 54.31\% $\pm$ 3.81 & 62.69\% $\pm$ 3.57 & 69.43\% $\pm$ 3.03 & 79.95\% $\pm$ 2.34 & 80.79\% $\pm$ 2.28 & \textbf{89.61\% $\pm$ 1.18} \\
Qwen 72B & 54.47\% $\pm$ 3.78 & 62.55\% $\pm$ 3.58 & 67.56\% $\pm$ 3.12 & 77.28\% $\pm$ 2.72 & 79.18\% $\pm$ 2.51 & 82.41\% $\pm$ 2.12 \\
Llama 8B & 57.24\% $\pm$ 3.51 & 67.74\% $\pm$ 3.05 & 70.65\% $\pm$ 2.93 & 72.40\% $\pm$ 3.22 & 71.27\% $\pm$ 3.16 & 70.09\% $\pm$ 3.26 \\
Llama 70B & 54.14\% $\pm$ 3.76 & 64.38\% $\pm$ 3.55 & 68.67\% $\pm$ 3.10 & 77.43\% $\pm$ 2.65 & 78.06\% $\pm$ 2.63 & 80.20\% $\pm$ 2.49 \\
DeepSeekV3 & 53.42\% $\pm$ 3.90 & 61.95\% $\pm$ 3.67 & 67.78\% $\pm$ 3.29 & 77.37\% $\pm$ 2.77 & 78.74\% $\pm$ 2.63 & 82.08\% $\pm$ 2.21 \\
GPT-4o Mini & 53.96\% $\pm$ 3.71 & 65.55\% $\pm$ 3.33 & 70.40\% $\pm$ 2.96 & 76.70\% $\pm$ 2.70 & 78.93\% $\pm$ 2.47 & 79.09\% $\pm$ 2.48 \\
GPT-4o & 54.86\% $\pm$ 3.86 & 62.63\% $\pm$ 3.55 & 67.03\% $\pm$ 3.31 & 78.17\% $\pm$ 2.61 & 77.30\% $\pm$ 2.74 & 80.74\% $\pm$ 2.44 \\
\bottomrule
\end{tabular}
}

\quad
\resizebox{\textwidth}{!}{
\begin{tabular}{lcccccc}
\toprule
 & Qwen 72B & Llama 8B & Llama 70B & DeepSeekV3 & GPT-4o Mini & GPT-4o \\
\midrule
Qwen 0.5B & 52.72\% $\pm$ 3.98 & 57.12\% $\pm$ 3.60 & 54.63\% $\pm$ 3.81 & 52.68\% $\pm$ 4.03 & 55.24\% $\pm$ 3.67 & 52.17\% $\pm$ 3.98 \\
Qwen 1.5B & 64.62\% $\pm$ 3.63 & 69.07\% $\pm$ 3.03 & 67.05\% $\pm$ 3.46 & 63.84\% $\pm$ 3.72 & 67.33\% $\pm$ 3.25 & 64.29\% $\pm$ 3.56 \\
Qwen 3B & 70.11\% $\pm$ 3.30 & 70.33\% $\pm$ 3.11 & 70.82\% $\pm$ 3.24 & 68.82\% $\pm$ 3.47 & 71.80\% $\pm$ 3.07 & 69.00\% $\pm$ 3.39 \\
Qwen 7B & 79.31\% $\pm$ 2.69 & 70.79\% $\pm$ 3.17 & 77.66\% $\pm$ 2.72 & 79.16\% $\pm$ 2.77 & 76.56\% $\pm$ 2.75 & 78.89\% $\pm$ 2.76 \\
Qwen 14B & 81.44\% $\pm$ 2.57 & 70.45\% $\pm$ 3.15 & 79.91\% $\pm$ 2.58 & 81.54\% $\pm$ 2.52 & 79.55\% $\pm$ 2.45 & 79.87\% $\pm$ 2.67 \\
Qwen 32B & 83.54\% $\pm$ 2.29 & 69.52\% $\pm$ 3.13 & 81.12\% $\pm$ 2.51 & 83.73\% $\pm$ 2.23 & 78.92\% $\pm$ 2.48 & 82.13\% $\pm$ 2.38 \\
Qwen 72B & \textbf{91.30\% $\pm$ 1.03} & 67.88\% $\pm$ 3.22 & 80.94\% $\pm$ 2.56 & 85.08\% $\pm$ 2.11 & 78.87\% $\pm$ 2.53 & 84.04\% $\pm$ 2.21 \\
Llama 8B & 68.56\% $\pm$ 3.39 & \textbf{88.01\% $\pm$ 1.17} & 73.17\% $\pm$ 3.05 & 69.80\% $\pm$ 3.51 & 69.99\% $\pm$ 3.23 & 69.48\% $\pm$ 3.39 \\
Llama 70B & 81.23\% $\pm$ 2.64 & 71.47\% $\pm$ 3.04 & \textbf{90.14\% $\pm$ 1.18} & 80.90\% $\pm$ 2.68 & 78.95\% $\pm$ 2.61 & 81.89\% $\pm$ 2.47 \\
DeepSeekV3 & 84.24\% $\pm$ 2.24 & 68.40\% $\pm$ 3.33 & 79.26\% $\pm$ 2.77 & \textbf{91.10\% $\pm$ 1.26} & 78.07\% $\pm$ 2.76 & 83.29\% $\pm$ 2.38 \\
GPT-4o Mini & 80.84\% $\pm$ 2.57 & 69.68\% $\pm$ 3.16 & 80.13\% $\pm$ 2.58 & 79.90\% $\pm$ 2.69 & \textbf{88.90\% $\pm$ 1.20} & 82.35\% $\pm$ 2.31 \\
GPT-4o & 83.78\% $\pm$ 2.33 & 69.32\% $\pm$ 3.23 & 80.61\% $\pm$ 2.55 & 84.00\% $\pm$ 2.27 & 79.43\% $\pm$ 2.49 & \textbf{89.74\% $\pm$ 1.35} \\
\bottomrule
\end{tabular}
}
\label{tab:mmlu_lime_microeconomics_fidelity}
\end{table*}

\begin{table*}
\centering
\caption{Accuracy of proxy \textbf{filtered} LIME explanations on high school microeconomics of MMLU datasets: each value shows how well LIME explanations generated by the model on the \textbf{left} serve as surrogates for predicting the behavior of the model on the \textbf{top}.}
\resizebox{\textwidth}{!}{
\begin{tabular}{lcccccc}
\toprule
 & Qwen 0.5B & Qwen 1.5B & Qwen 3B & Qwen 7B & Qwen 14B & Qwen 32B \\
\midrule
Qwen 0.5B & \textbf{87.93\% $\pm$ 1.08} & 71.42\% $\pm$ 3.82 & 70.66\% $\pm$ 3.94 & 75.31\% $\pm$ 3.72 & 71.32\% $\pm$ 4.38 & 72.11\% $\pm$ 4.26 \\
Qwen 1.5B & 72.28\% $\pm$ 3.80 & \textbf{88.26\% $\pm$ 1.06} & 76.60\% $\pm$ 2.96 & 77.90\% $\pm$ 2.79 & 75.49\% $\pm$ 3.09 & 74.98\% $\pm$ 3.17 \\
Qwen 3B & 71.68\% $\pm$ 4.06 & 75.48\% $\pm$ 3.01 & \textbf{87.87\% $\pm$ 1.17} & 78.61\% $\pm$ 2.86 & 79.81\% $\pm$ 2.63 & 77.66\% $\pm$ 2.90 \\
Qwen 7B & 74.44\% $\pm$ 3.93 & 74.49\% $\pm$ 3.07 & 75.93\% $\pm$ 3.03 & \textbf{89.77\% $\pm$ 1.11} & 81.19\% $\pm$ 2.52 & 82.64\% $\pm$ 2.21 \\
Qwen 14B & 71.16\% $\pm$ 4.62 & 73.29\% $\pm$ 3.36 & 77.97\% $\pm$ 2.79 & 81.01\% $\pm$ 2.56 & \textbf{89.15\% $\pm$ 1.13} & 83.69\% $\pm$ 2.22 \\
Qwen 32B & 71.51\% $\pm$ 4.42 & 71.62\% $\pm$ 3.45 & 74.99\% $\pm$ 3.08 & 82.05\% $\pm$ 2.26 & 82.01\% $\pm$ 2.35 & \textbf{89.61\% $\pm$ 1.18} \\
Qwen 72B & 71.15\% $\pm$ 4.35 & 72.15\% $\pm$ 3.44 & 73.10\% $\pm$ 3.10 & 80.02\% $\pm$ 2.51 & 81.25\% $\pm$ 2.32 & 83.41\% $\pm$ 2.07 \\
Llama 8B & 69.67\% $\pm$ 4.22 & 73.13\% $\pm$ 2.98 & 77.13\% $\pm$ 2.97 & 79.04\% $\pm$ 2.70 & 78.07\% $\pm$ 2.77 & 77.14\% $\pm$ 2.89 \\
Llama 70B & 72.36\% $\pm$ 4.08 & 72.67\% $\pm$ 3.45 & 73.77\% $\pm$ 3.20 & 79.87\% $\pm$ 2.48 & 80.70\% $\pm$ 2.45 & 82.09\% $\pm$ 2.30 \\
DeepSeekV3 & 72.62\% $\pm$ 4.39 & 72.14\% $\pm$ 3.47 & 74.21\% $\pm$ 3.28 & 80.56\% $\pm$ 2.55 & 80.91\% $\pm$ 2.47 & 83.19\% $\pm$ 2.18 \\
GPT-4o Mini & 69.86\% $\pm$ 4.43 & 73.41\% $\pm$ 3.36 & 74.36\% $\pm$ 3.09 & 79.30\% $\pm$ 2.54 & 81.40\% $\pm$ 2.31 & 81.06\% $\pm$ 2.39 \\
GPT-4o & 72.41\% $\pm$ 4.31 & 70.87\% $\pm$ 3.55 & 72.07\% $\pm$ 3.43 & 80.59\% $\pm$ 2.52 & 79.33\% $\pm$ 2.65 & 82.50\% $\pm$ 2.26 \\
\bottomrule
\end{tabular}
}

\quad
\resizebox{\textwidth}{!}{
\begin{tabular}{lcccccc}
\toprule
 & Qwen 72B & Llama 8B & Llama 70B & DeepSeekV3 & GPT-4o Mini & GPT-4o \\
\midrule
Qwen 0.5B & 73.13\% $\pm$ 4.20 & 69.84\% $\pm$ 4.17 & 72.93\% $\pm$ 4.15 & 74.78\% $\pm$ 4.25 & 70.34\% $\pm$ 4.34 & 73.23\% $\pm$ 4.25 \\
Qwen 1.5B & 75.76\% $\pm$ 3.21 & 74.88\% $\pm$ 2.90 & 75.81\% $\pm$ 3.27 & 75.16\% $\pm$ 3.25 & 75.07\% $\pm$ 3.18 & 74.41\% $\pm$ 3.21 \\
Qwen 3B & 77.03\% $\pm$ 3.02 & 77.87\% $\pm$ 2.94 & 76.86\% $\pm$ 3.11 & 76.90\% $\pm$ 3.21 & 76.24\% $\pm$ 3.02 & 75.66\% $\pm$ 3.23 \\
Qwen 7B & 82.62\% $\pm$ 2.34 & 77.60\% $\pm$ 2.78 & 80.77\% $\pm$ 2.51 & 83.14\% $\pm$ 2.33 & 79.40\% $\pm$ 2.57 & 82.99\% $\pm$ 2.30 \\
Qwen 14B & 84.06\% $\pm$ 2.23 & 76.78\% $\pm$ 2.95 & 82.49\% $\pm$ 2.42 & 83.49\% $\pm$ 2.39 & 81.62\% $\pm$ 2.36 & 82.31\% $\pm$ 2.44 \\
Qwen 32B & 85.46\% $\pm$ 1.98 & 75.55\% $\pm$ 3.00 & 82.86\% $\pm$ 2.38 & 84.95\% $\pm$ 2.10 & 80.35\% $\pm$ 2.46 & 83.93\% $\pm$ 2.13 \\
Qwen 72B & \textbf{91.30\% $\pm$ 1.03} & 73.53\% $\pm$ 3.15 & 83.95\% $\pm$ 2.15 & 86.10\% $\pm$ 2.01 & 81.21\% $\pm$ 2.25 & 85.98\% $\pm$ 1.78 \\
Llama 8B & 76.11\% $\pm$ 3.05 & \textbf{88.01\% $\pm$ 1.17} & 79.83\% $\pm$ 2.70 & 78.65\% $\pm$ 2.99 & 76.35\% $\pm$ 3.08 & 77.54\% $\pm$ 2.95 \\
Llama 70B & 84.99\% $\pm$ 2.05 & 77.32\% $\pm$ 2.79 & \textbf{90.14\% $\pm$ 1.18} & 83.70\% $\pm$ 2.30 & 81.08\% $\pm$ 2.46 & 84.05\% $\pm$ 2.21 \\
DeepSeekV3 & 85.85\% $\pm$ 1.99 & 75.30\% $\pm$ 3.15 & 82.38\% $\pm$ 2.42 & \textbf{91.10\% $\pm$ 1.26} & 79.90\% $\pm$ 2.60 & 84.97\% $\pm$ 2.03 \\
GPT-4o Mini & 83.76\% $\pm$ 2.17 & 75.54\% $\pm$ 3.05 & 82.48\% $\pm$ 2.38 & 82.57\% $\pm$ 2.39 & \textbf{88.90\% $\pm$ 1.20} & 84.02\% $\pm$ 2.17 \\
GPT-4o & 85.85\% $\pm$ 1.96 & 75.12\% $\pm$ 3.16 & 82.38\% $\pm$ 2.44 & 85.23\% $\pm$ 2.06 & 80.76\% $\pm$ 2.45 & \textbf{89.74\% $\pm$ 1.35 } \\
\bottomrule
\end{tabular}
}
\label{tab:mmlu_lime_microeconomics_filtered_fidelity}
\end{table*}

\begin{table*}
\centering
\caption{Accuracy of proxy LIME explanations on high school psychology of MMLU datasets: each value shows how well LIME explanations generated by the model on the \textbf{left} serve as surrogates for predicting the behavior of the model on the \textbf{top}.}
\resizebox{\textwidth}{!}{
\begin{tabular}{lcccccc}
\toprule
 & Qwen 0.5B & Qwen 1.5B & Qwen 3B & Qwen 7B & Qwen 14B & Qwen 32B \\
\midrule
Qwen 0.5B & \textbf{88.25\% $\pm$ 0.64} & 66.76\% $\pm$ 2.11 & 64.46\% $\pm$ 2.06 & 63.40\% $\pm$ 2.23 & 62.75\% $\pm$ 2.29 & 61.03\% $\pm$ 2.33 \\
Qwen 1.5B & 65.69\% $\pm$ 2.10 & \textbf{88.57\% $\pm$ 0.68} & 75.99\% $\pm$ 1.57 & 77.51\% $\pm$ 1.59 & 76.13\% $\pm$ 1.75 & 75.68\% $\pm$ 1.80 \\
Qwen 3B & 63.16\% $\pm$ 2.12 & 75.93\% $\pm$ 1.66 & \textbf{87.24\% $\pm$ 0.71} & 78.53\% $\pm$ 1.60 & 79.39\% $\pm$ 1.56 & 78.77\% $\pm$ 1.61 \\
Qwen 7B & 62.57\% $\pm$ 2.21 & 75.22\% $\pm$ 1.72 & 76.68\% $\pm$ 1.60 & \textbf{89.29\% $\pm$ 0.67} & 82.26\% $\pm$ 1.40 & 82.27\% $\pm$ 1.50 \\
Qwen 14B & 61.51\% $\pm$ 2.25 & 73.82\% $\pm$ 1.83 & 76.13\% $\pm$ 1.63 & 81.10\% $\pm$ 1.44 & \textbf{89.63\% $\pm$ 0.68} & 85.26\% $\pm$ 1.18 \\
Qwen 32B & 60.70\% $\pm$ 2.26 & 72.87\% $\pm$ 1.82 & 75.14\% $\pm$ 1.61 & 80.36\% $\pm$ 1.48 & 84.08\% $\pm$ 1.21 & \textbf{89.88\% $\pm$ 0.68} \\
Qwen 72B & 60.12\% $\pm$ 2.34 & 72.42\% $\pm$ 1.95 & 73.94\% $\pm$ 1.72 & 79.51\% $\pm$ 1.57 & 82.40\% $\pm$ 1.39 & 84.50\% $\pm$ 1.25 \\
Llama 8B & 61.59\% $\pm$ 2.21 & 73.76\% $\pm$ 1.83 & 76.16\% $\pm$ 1.60 & 79.36\% $\pm$ 1.58 & 79.93\% $\pm$ 1.57 & 80.45\% $\pm$ 1.51 \\
Llama 70B & 60.03\% $\pm$ 2.33 & 71.82\% $\pm$ 1.98 & 73.37\% $\pm$ 1.75 & 79.26\% $\pm$ 1.59 & 81.78\% $\pm$ 1.45 & 84.20\% $\pm$ 1.27 \\
DeepSeekV3 & 58.92\% $\pm$ 2.33 & 72.55\% $\pm$ 1.99 & 73.64\% $\pm$ 1.76 & 79.44\% $\pm$ 1.61 & 81.52\% $\pm$ 1.50 & 83.68\% $\pm$ 1.36 \\
GPT-4o Mini & 60.51\% $\pm$ 2.26 & 72.04\% $\pm$ 1.98 & 74.61\% $\pm$ 1.69 & 78.25\% $\pm$ 1.63 & 81.76\% $\pm$ 1.43 & 83.07\% $\pm$ 1.33 \\
GPT-4o & 58.06\% $\pm$ 2.37 & 69.99\% $\pm$ 2.12 & 73.06\% $\pm$ 1.79 & 77.77\% $\pm$ 1.71 & 80.69\% $\pm$ 1.52 & 82.78\% $\pm$ 1.42 \\
\bottomrule
\end{tabular}
}

\quad
\resizebox{\textwidth}{!}{
\begin{tabular}{lcccccc}
\toprule
 & Qwen 72B & Llama 8B & Llama 70B & DeepSeekV3 & GPT-4o Mini & GPT-4o \\
\midrule
Qwen 0.5B & 60.22\% $\pm$ 2.51 & 62.79\% $\pm$ 2.21 & 59.91\% $\pm$ 2.50 & 58.71\% $\pm$ 2.55 & 60.59\% $\pm$ 2.42 & 57.47\% $\pm$ 2.59 \\
Qwen 1.5B & 76.11\% $\pm$ 1.90 & 76.19\% $\pm$ 1.69 & 75.19\% $\pm$ 1.95 & 75.78\% $\pm$ 1.99 & 74.21\% $\pm$ 1.96 & 73.23\% $\pm$ 2.15 \\
Qwen 3B & 78.69\% $\pm$ 1.75 & 78.03\% $\pm$ 1.59 & 77.75\% $\pm$ 1.83 & 78.33\% $\pm$ 1.81 & 78.80\% $\pm$ 1.68 & 77.51\% $\pm$ 1.87 \\
Qwen 7B & 82.57\% $\pm$ 1.53 & 79.27\% $\pm$ 1.59 & 81.86\% $\pm$ 1.60 & 82.56\% $\pm$ 1.65 & 79.25\% $\pm$ 1.76 & 79.75\% $\pm$ 1.85 \\
Qwen 14B & 85.50\% $\pm$ 1.30 & 78.79\% $\pm$ 1.60 & 83.63\% $\pm$ 1.45 & 84.63\% $\pm$ 1.48 & 82.42\% $\pm$ 1.53 & 82.89\% $\pm$ 1.64 \\
Qwen 32B & 86.49\% $\pm$ 1.12 & 78.62\% $\pm$ 1.55 & 85.38\% $\pm$ 1.25 & 86.03\% $\pm$ 1.32 & 82.60\% $\pm$ 1.43 & 84.51\% $\pm$ 1.45 \\
Qwen 72B & \textbf{91.35\% $\pm$ 0.61} & 77.26\% $\pm$ 1.72 & 85.66\% $\pm$ 1.30 & 87.48\% $\pm$ 1.13 & 82.10\% $\pm$ 1.58 & 85.98\% $\pm$ 1.40 \\
Llama 8B & 80.32\% $\pm$ 1.70 & \textbf{88.70\% $\pm$ 0.68} & 80.68\% $\pm$ 1.68 & 80.73\% $\pm$ 1.70 & 79.02\% $\pm$ 1.76 & 78.50\% $\pm$ 1.93 \\
Llama 70B & 86.29\% $\pm$ 1.26 & 77.96\% $\pm$ 1.68 & \textbf{91.05\% $\pm$ 0.65} & 85.62\% $\pm$ 1.41 & 82.91\% $\pm$ 1.53 & 85.07\% $\pm$ 1.49 \\
DeepSeekV3 & 86.36\% $\pm$ 1.15 & 77.69\% $\pm$ 1.70 & 84.94\% $\pm$ 1.35 & \textbf{91.77\% $\pm$ 0.68} & 81.58\% $\pm$ 1.67 & 85.76\% $\pm$ 1.44 \\
GPT-4o Mini & 84.76\% $\pm$ 1.40 & 77.64\% $\pm$ 1.68 & 84.43\% $\pm$ 1.40 & 84.59\% $\pm$ 1.51 & \textbf{89.93\% $\pm$ 0.78} & 85.50\% $\pm$ 1.43 \\
GPT-4o & 85.49\% $\pm$ 1.31 & 75.91\% $\pm$ 1.83 & 84.49\% $\pm$ 1.45 & 87.05\% $\pm$ 1.21 & 83.04\% $\pm$ 1.48 & \textbf{90.62\% $\pm$ 0.81} \\
\bottomrule
\end{tabular}
}
\label{tab:mmlu_lime_psychology_fidelity}
\end{table*}

\begin{table*}
\centering
\caption{Accuracy of proxy \textbf{filtered} LIME explanations on high school psychology of MMLU datasets: each value shows how well LIME explanations generated by the model on the \textbf{left} serve as surrogates for predicting the behavior of the model on the \textbf{top}.}
\resizebox{\textwidth}{!}{
\begin{tabular}{lcccccc}
\toprule
 & Qwen 0.5B & Qwen 1.5B & Qwen 3B & Qwen 7B & Qwen 14B & Qwen 32B \\
\midrule
Qwen 0.5B & \textbf{88.25\% $\pm$ 0.64} & 75.04\% $\pm$ 2.10 & 73.04\% $\pm$ 2.18 & 73.87\% $\pm$ 2.18 & 73.82\% $\pm$ 2.35 & 72.81\% $\pm$ 2.41 \\
Qwen 1.5B & 72.43\% $\pm$ 2.23 & \textbf{88.57\% $\pm$ 0.68} & 79.09\% $\pm$ 1.48 & 80.55\% $\pm$ 1.52 & 80.12\% $\pm$ 1.67 & 80.10\% $\pm$ 1.70 \\
Qwen 3B & 70.61\% $\pm$ 2.28 & 79.03\% $\pm$ 1.55 & \textbf{87.24\% $\pm$ 0.71} & 80.59\% $\pm$ 1.54 & 82.01\% $\pm$ 1.47 & 81.97\% $\pm$ 1.51 \\
Qwen 7B & 70.64\% $\pm$ 2.30 & 78.60\% $\pm$ 1.64 & 78.69\% $\pm$ 1.56 & \textbf{89.29\% $\pm$ 0.67} & 83.83\% $\pm$ 1.28 & 84.24\% $\pm$ 1.37 \\
Qwen 14B & 70.09\% $\pm$ 2.46 & 77.96\% $\pm$ 1.74 & 78.79\% $\pm$ 1.58 & 82.83\% $\pm$ 1.34 & \textbf{89.63\% $\pm$ 0.68} & 86.68\% $\pm$ 1.09 \\
Qwen 32B & 68.77\% $\pm$ 2.50 & 76.73\% $\pm$ 1.79 & 77.90\% $\pm$ 1.59 & 82.17\% $\pm$ 1.41 & 85.24\% $\pm$ 1.16 & \textbf{89.88\% $\pm$ 0.68} \\
Qwen 72B & 69.04\% $\pm$ 2.52 & 76.67\% $\pm$ 1.87 & 76.67\% $\pm$ 1.71 & 81.19\% $\pm$ 1.51 & 83.87\% $\pm$ 1.34 & 85.51\% $\pm$ 1.16 \\
Llama 8B & 69.15\% $\pm$ 2.43 & 77.93\% $\pm$ 1.73 & 79.20\% $\pm$ 1.55 & 82.30\% $\pm$ 1.41 & 83.04\% $\pm$ 1.41 & 83.39\% $\pm$ 1.40 \\
Llama 70B & 69.17\% $\pm$ 2.59 & 76.41\% $\pm$ 1.87 & 76.37\% $\pm$ 1.72 & 81.32\% $\pm$ 1.51 & 83.14\% $\pm$ 1.40 & 85.05\% $\pm$ 1.24 \\
DeepSeekV3 & 67.48\% $\pm$ 2.54 & 76.95\% $\pm$ 1.89 & 76.78\% $\pm$ 1.74 & 81.31\% $\pm$ 1.51 & 83.27\% $\pm$ 1.38 & 85.16\% $\pm$ 1.21 \\
GPT-4o Mini & 69.71\% $\pm$ 2.45 & 76.70\% $\pm$ 1.87 & 77.26\% $\pm$ 1.65 & 80.63\% $\pm$ 1.50 & 83.59\% $\pm$ 1.32 & 84.32\% $\pm$ 1.26 \\
GPT-4o & 67.44\% $\pm$ 2.63 & 75.65\% $\pm$ 1.96 & 76.24\% $\pm$ 1.78 & 80.41\% $\pm$ 1.60 & 82.65\% $\pm$ 1.45 & 84.33\% $\pm$ 1.31 \\
\bottomrule
\end{tabular}
}

\quad
\resizebox{\textwidth}{!}{
\begin{tabular}{lcccccc}
\toprule
 & Qwen 72B & Llama 8B & Llama 70B & DeepSeekV3 & GPT-4o Mini & GPT-4o \\
\midrule
Qwen 0.5B & 73.54\% $\pm$ 2.44 & 72.37\% $\pm$ 2.35 & 73.02\% $\pm$ 2.53 & 71.91\% $\pm$ 2.51 & 73.67\% $\pm$ 2.36 & 72.08\% $\pm$ 2.58 \\
Qwen 1.5B & 80.93\% $\pm$ 1.73 & 79.69\% $\pm$ 1.66 & 80.31\% $\pm$ 1.75 & 81.04\% $\pm$ 1.79 & 79.70\% $\pm$ 1.76 & 79.79\% $\pm$ 1.87 \\
Qwen 3B & 82.04\% $\pm$ 1.61 & 81.16\% $\pm$ 1.51 & 81.33\% $\pm$ 1.68 & 82.42\% $\pm$ 1.65 & 81.85\% $\pm$ 1.56 & 81.78\% $\pm$ 1.71 \\
Qwen 7B & 84.59\% $\pm$ 1.37 & 82.06\% $\pm$ 1.46 & 84.30\% $\pm$ 1.44 & 85.05\% $\pm$ 1.45 & 82.25\% $\pm$ 1.50 & 83.44\% $\pm$ 1.56 \\
Qwen 14B & 87.18\% $\pm$ 1.17 & 81.97\% $\pm$ 1.46 & 85.33\% $\pm$ 1.35 & 86.72\% $\pm$ 1.28 & 84.89\% $\pm$ 1.29 & 85.86\% $\pm$ 1.38 \\
Qwen 32B & 87.42\% $\pm$ 1.04 & 81.18\% $\pm$ 1.51 & 86.17\% $\pm$ 1.23 & 87.56\% $\pm$ 1.12 & 84.28\% $\pm$ 1.29 & 86.38\% $\pm$ 1.25 \\
Qwen 72B & \textbf{91.35\% $\pm$ 0.61} & 80.19\% $\pm$ 1.68 & 86.53\% $\pm$ 1.23 & 88.09\% $\pm$ 1.10 & 84.42\% $\pm$ 1.30 & 87.40\% $\pm$ 1.19 \\
Llama 8B & 83.76\% $\pm$ 1.52 & \textbf{88.70\% $\pm$ 0.68} & 83.95\% $\pm$ 1.49 & 84.12\% $\pm$ 1.50 & 82.57\% $\pm$ 1.54 & 82.98\% $\pm$ 1.63 \\
Llama 70B & 87.20\% $\pm$ 1.18 & 80.96\% $\pm$ 1.61 & \textbf{91.05\% $\pm$ 0.65} & 87.22\% $\pm$ 1.25 & 84.82\% $\pm$ 1.30 & 86.94\% $\pm$ 1.24 \\
DeepSeekV3 & 86.97\% $\pm$ 1.10 & 80.37\% $\pm$ 1.64 & 86.26\% $\pm$ 1.25 & \textbf{91.77\% $\pm$ 0.68} & 84.10\% $\pm$ 1.39 & 87.40\% $\pm$ 1.20 \\
GPT-4o Mini & 86.67\% $\pm$ 1.17 & 80.70\% $\pm$ 1.59 & 86.04\% $\pm$ 1.22 & 86.66\% $\pm$ 1.30 & \textbf{89.93\% $\pm$ 0.78} & 86.88\% $\pm$ 1.29 \\
GPT-4o & 86.68\% $\pm$ 1.14 & 79.34\% $\pm$ 1.73 & 86.06\% $\pm$ 1.25 & 87.99\% $\pm$ 1.12 & 84.46\% $\pm$ 1.36 & \textbf{90.62\% $\pm$ 0.81} \\
\bottomrule
\end{tabular}
}
\label{tab:mmlu_lime_psychology_filtered_fidelity}
\end{table*}

\begin{table*}
\centering
\caption{Accuracy of proxy LIME explanations on high school world history of MMLU datasets: each value shows how well LIME explanations generated by the model on the \textbf{left} serve as surrogates for predicting the behavior of the model on the \textbf{top}.}
\resizebox{\textwidth}{!}{
\begin{tabular}{lcccccc}
\toprule
 & Qwen 0.5B & Qwen 1.5B & Qwen 3B & Qwen 7B & Qwen 14B & Qwen 32B \\
\midrule
Qwen 0.5B & \textbf{90.36\% $\pm$ 0.91} & 68.25\% $\pm$ 3.34 & 63.86\% $\pm$ 3.73 & 65.07\% $\pm$ 3.85 & 64.21\% $\pm$ 3.95 & 63.09\% $\pm$ 4.10 \\
Qwen 1.5B & 69.49\% $\pm$ 3.18 & \textbf{91.99\% $\pm$ 0.83} & 76.31\% $\pm$ 2.98 & 73.84\% $\pm$ 3.29 & 73.58\% $\pm$ 3.46 & 72.11\% $\pm$ 3.70 \\
Qwen 3B & 65.82\% $\pm$ 3.52 & 74.68\% $\pm$ 3.05 & \textbf{93.28\% $\pm$ 0.78} & 79.03\% $\pm$ 2.87 & 79.62\% $\pm$ 2.87 & 79.05\% $\pm$ 3.08 \\
Qwen 7B & 65.24\% $\pm$ 3.67 & 72.01\% $\pm$ 3.37 & 78.18\% $\pm$ 2.93 & \textbf{94.01\% $\pm$ 0.77} & 84.16\% $\pm$ 2.38 & 83.29\% $\pm$ 2.60 \\
Qwen 14B & 64.82\% $\pm$ 3.70 & 71.56\% $\pm$ 3.44 & 77.93\% $\pm$ 2.98 & 83.48\% $\pm$ 2.36 & \textbf{94.83\% $\pm$ 0.74} & 89.08\% $\pm$ 1.84 \\
Qwen 32B & 64.62\% $\pm$ 3.75 & 71.06\% $\pm$ 3.56 & 77.67\% $\pm$ 3.03 & 82.73\% $\pm$ 2.52 & 88.93\% $\pm$ 1.80 & \textbf{95.39\% $\pm$ 0.75} \\
Qwen 72B & 64.12\% $\pm$ 3.82 & 70.28\% $\pm$ 3.66 & 77.60\% $\pm$ 3.18 & 81.59\% $\pm$ 2.80 & 87.11\% $\pm$ 2.16 & 87.85\% $\pm$ 2.24 \\
Llama 8B & 65.05\% $\pm$ 3.64 & 72.43\% $\pm$ 3.34 & 79.46\% $\pm$ 2.81 & 81.47\% $\pm$ 2.60 & 82.76\% $\pm$ 2.62 & 82.49\% $\pm$ 2.75 \\
Llama 70B & 64.27\% $\pm$ 3.83 & 70.04\% $\pm$ 3.65 & 77.32\% $\pm$ 3.10 & 81.53\% $\pm$ 2.71 & 86.75\% $\pm$ 2.16 & 87.39\% $\pm$ 2.27 \\
DeepSeekV3 & 63.64\% $\pm$ 4.01 & 70.65\% $\pm$ 3.73 & 76.50\% $\pm$ 3.36 & 81.45\% $\pm$ 2.83 & 86.32\% $\pm$ 2.35 & 86.76\% $\pm$ 2.45 \\
GPT-4o Mini & 64.55\% $\pm$ 3.79 & 70.86\% $\pm$ 3.61 & 77.83\% $\pm$ 3.06 & 82.42\% $\pm$ 2.65 & 86.04\% $\pm$ 2.29 & 85.71\% $\pm$ 2.56 \\
GPT-4o & 63.23\% $\pm$ 3.98 & 69.48\% $\pm$ 3.79 & 75.88\% $\pm$ 3.42 & 81.55\% $\pm$ 2.80 & 86.39\% $\pm$ 2.30 & 87.06\% $\pm$ 2.40 \\
\bottomrule
\end{tabular}
}

\quad
\resizebox{\textwidth}{!}{
\begin{tabular}{lcccccc}
\toprule
 & Qwen 72B & Llama 8B & Llama 70B & DeepSeekV3 & GPT-4o Mini & GPT-4o \\
\midrule
Qwen 0.5B & 62.71\% $\pm$ 4.21 & 63.99\% $\pm$ 3.88 & 62.61\% $\pm$ 4.18 & 62.32\% $\pm$ 4.38 & 63.97\% $\pm$ 4.04 & 61.13\% $\pm$ 4.41 \\
Qwen 1.5B & 71.75\% $\pm$ 3.84 & 73.66\% $\pm$ 3.35 & 71.11\% $\pm$ 3.73 & 72.88\% $\pm$ 3.83 & 72.11\% $\pm$ 3.70 & 70.35\% $\pm$ 3.99 \\
Qwen 3B & 79.03\% $\pm$ 3.28 & 79.90\% $\pm$ 2.84 & 78.50\% $\pm$ 3.23 & 78.03\% $\pm$ 3.49 & 79.14\% $\pm$ 3.09 & 77.49\% $\pm$ 3.56 \\
Qwen 7B & 83.03\% $\pm$ 2.84 & 81.11\% $\pm$ 2.71 & 82.05\% $\pm$ 2.85 & 82.64\% $\pm$ 2.96 & 83.01\% $\pm$ 2.78 & 82.70\% $\pm$ 2.93 \\
Qwen 14B & 87.92\% $\pm$ 2.20 & 81.78\% $\pm$ 2.72 & 87.24\% $\pm$ 2.22 & 87.44\% $\pm$ 2.38 & 86.08\% $\pm$ 2.29 & 87.39\% $\pm$ 2.36 \\
Qwen 32B & 88.52\% $\pm$ 2.23 & 81.44\% $\pm$ 2.79 & 87.95\% $\pm$ 2.19 & 87.76\% $\pm$ 2.42 & 85.73\% $\pm$ 2.45 & 88.09\% $\pm$ 2.37 \\
Qwen 72B & \textbf{96.52\% $\pm$ 0.59} & 81.28\% $\pm$ 3.01 & 89.10\% $\pm$ 2.09 & 89.58\% $\pm$ 2.24 & 86.52\% $\pm$ 2.49 & 90.36\% $\pm$ 2.15 \\
Llama 8B & 82.54\% $\pm$ 3.03 & \textbf{93.42\% $\pm$ 0.84} & 82.66\% $\pm$ 2.82 & 82.56\% $\pm$ 3.08 & 82.79\% $\pm$ 2.76 & 82.05\% $\pm$ 3.15 \\
Llama 70B & 89.32\% $\pm$ 2.13 & 81.60\% $\pm$ 2.82 & \textbf{95.32\% $\pm$ 0.77} & 88.75\% $\pm$ 2.26 & 85.94\% $\pm$ 2.48 & 89.57\% $\pm$ 2.19 \\
DeepSeekV3 & 89.23\% $\pm$ 2.24 & 81.05\% $\pm$ 3.03 & 87.85\% $\pm$ 2.36 & \textbf{96.40\% $\pm$ 0.70} & 85.64\% $\pm$ 2.57 & 91.06\% $\pm$ 1.98 \\
GPT-4o Mini & 87.32\% $\pm$ 2.54 & 82.60\% $\pm$ 2.71 & 87.26\% $\pm$ 2.32 & 87.26\% $\pm$ 2.43 & \textbf{95.49\% $\pm$ 0.70} & 87.23\% $\pm$ 2.54 \\
GPT-4o & 90.03\% $\pm$ 2.10 & 80.74\% $\pm$ 3.09 & 88.63\% $\pm$ 2.27 & 90.94\% $\pm$ 1.97 & 86.06\% $\pm$ 2.53 & \textbf{96.75\% $\pm$ 0.61 } \\
\bottomrule
\end{tabular}
}
\label{tab:mmlu_lime_world_history_fidelity}
\end{table*}

\begin{table*}
\centering
\caption{Accuracy of proxy \textbf{filtered} LIME explanations on high school world history of MMLU datasets: each value shows how well LIME explanations generated by the model on the \textbf{left} serve as surrogates for predicting the behavior of the model on the \textbf{top}.}
\resizebox{\textwidth}{!}{
\begin{tabular}{lcccccc}
\toprule
 & Qwen 0.5B & Qwen 1.5B & Qwen 3B & Qwen 7B & Qwen 14B & Qwen 32B \\
\midrule
Qwen 0.5B & \textbf{90.36\% $\pm$ 0.91} & 76.59\% $\pm$ 3.51 & 74.15\% $\pm$ 3.92 & 76.08\% $\pm$ 3.77 & 77.79\% $\pm$ 3.81 & 78.14\% $\pm$ 3.76 \\
Qwen 1.5B & 76.23\% $\pm$ 3.48 & \textbf{91.99\% $\pm$ 0.83} & 79.96\% $\pm$ 2.99 & 80.04\% $\pm$ 3.27 & 80.87\% $\pm$ 3.28 & 80.63\% $\pm$ 3.34 \\
Qwen 3B & 73.98\% $\pm$ 3.87 & 79.24\% $\pm$ 2.99 & \textbf{93.28\% $\pm$ 0.78} & 80.66\% $\pm$ 3.03 & 83.21\% $\pm$ 2.86 & 83.12\% $\pm$ 2.91 \\
Qwen 7B & 75.56\% $\pm$ 3.68 & 78.94\% $\pm$ 3.24 & 79.94\% $\pm$ 3.03 & \textbf{94.01\% $\pm$ 0.77} & 86.40\% $\pm$ 2.26 & 86.03\% $\pm$ 2.39 \\
Qwen 14B & 76.69\% $\pm$ 3.72 & 79.12\% $\pm$ 3.26 & 81.81\% $\pm$ 2.87 & 85.34\% $\pm$ 2.31 & \textbf{94.83\% $\pm$ 0.74} & 90.40\% $\pm$ 1.76 \\
Qwen 32B & 76.53\% $\pm$ 3.59 & 78.90\% $\pm$ 3.25 & 81.25\% $\pm$ 2.93 & 84.70\% $\pm$ 2.42 & 89.95\% $\pm$ 1.77 & \textbf{95.39\% $\pm$ 0.75} \\
Qwen 72B & 75.24\% $\pm$ 3.82 & 78.44\% $\pm$ 3.40 & 81.38\% $\pm$ 3.06 & 84.27\% $\pm$ 2.67 & 89.40\% $\pm$ 1.92 & 89.90\% $\pm$ 2.06 \\
Llama 8B & 75.64\% $\pm$ 3.70 & 78.88\% $\pm$ 3.29 & 81.76\% $\pm$ 2.81 & 83.92\% $\pm$ 2.59 & 85.68\% $\pm$ 2.55 & 85.21\% $\pm$ 2.64 \\
Llama 70B & 76.14\% $\pm$ 3.90 & 78.74\% $\pm$ 3.36 & 81.53\% $\pm$ 2.96 & 83.56\% $\pm$ 2.68 & 89.29\% $\pm$ 1.96 & 89.69\% $\pm$ 2.03 \\
DeepSeekV3 & 75.74\% $\pm$ 3.92 & 79.76\% $\pm$ 3.37 & 80.39\% $\pm$ 3.29 & 83.34\% $\pm$ 2.82 & 88.67\% $\pm$ 2.08 & 88.98\% $\pm$ 2.11 \\
GPT-4o Mini & 76.04\% $\pm$ 3.63 & 78.45\% $\pm$ 3.49 & 81.58\% $\pm$ 2.91 & 84.88\% $\pm$ 2.53 & 88.84\% $\pm$ 2.03 & 88.13\% $\pm$ 2.35 \\
GPT-4o & 76.52\% $\pm$ 3.87 & 78.35\% $\pm$ 3.43 & 79.97\% $\pm$ 3.30 & 83.27\% $\pm$ 2.86 & 88.40\% $\pm$ 2.20 & 88.81\% $\pm$ 2.23 \\
\bottomrule
\end{tabular}
}

\quad
\resizebox{\textwidth}{!}{
\begin{tabular}{lcccccc}
\toprule
 & Qwen 72B & Llama 8B & Llama 70B & DeepSeekV3 & GPT-4o Mini & GPT-4o \\
\midrule
Qwen 0.5B & 76.94\% $\pm$ 3.95 & 76.53\% $\pm$ 3.79 & 77.58\% $\pm$ 3.98 & 77.58\% $\pm$ 4.04 & 77.23\% $\pm$ 3.77 & 78.31\% $\pm$ 4.00 \\
Qwen 1.5B & 80.67\% $\pm$ 3.47 & 80.07\% $\pm$ 3.28 & 80.49\% $\pm$ 3.42 & 82.30\% $\pm$ 3.44 & 80.00\% $\pm$ 3.54 & 80.51\% $\pm$ 3.56 \\
Qwen 3B & 83.64\% $\pm$ 3.08 & 82.65\% $\pm$ 2.75 & 83.40\% $\pm$ 2.95 & 82.79\% $\pm$ 3.28 & 83.13\% $\pm$ 2.93 & 82.49\% $\pm$ 3.32 \\
Qwen 7B & 86.09\% $\pm$ 2.63 & 84.18\% $\pm$ 2.63 & 85.05\% $\pm$ 2.67 & 85.67\% $\pm$ 2.70 & 85.94\% $\pm$ 2.54 & 85.18\% $\pm$ 2.81 \\
Qwen 14B & 90.45\% $\pm$ 1.86 & 85.09\% $\pm$ 2.56 & 89.95\% $\pm$ 1.98 & 89.98\% $\pm$ 2.04 & 88.94\% $\pm$ 2.05 & 89.87\% $\pm$ 2.12 \\
Qwen 32B & 90.58\% $\pm$ 2.03 & 84.23\% $\pm$ 2.65 & 89.91\% $\pm$ 2.03 & 89.82\% $\pm$ 2.08 & 87.65\% $\pm$ 2.35 & 89.77\% $\pm$ 2.18 \\
Qwen 72B & \textbf{96.52\% $\pm$ 0.59} & 85.49\% $\pm$ 2.69 & 90.20\% $\pm$ 2.06 & 91.99\% $\pm$ 1.81 & 88.03\% $\pm$ 2.38 & 92.37\% $\pm$ 1.80 \\
Llama 8B & 86.96\% $\pm$ 2.67 & \textbf{93.42\% $\pm$ 0.84} & 87.26\% $\pm$ 2.47 & 88.00\% $\pm$ 2.59 & 86.85\% $\pm$ 2.48 & 86.80\% $\pm$ 2.69 \\
Llama 70B & 90.72\% $\pm$ 2.03 & 85.99\% $\pm$ 2.50 & \textbf{95.32\% $\pm$ 0.77} & 90.08\% $\pm$ 2.12 & 88.27\% $\pm$ 2.24 & 90.49\% $\pm$ 2.13 \\
DeepSeekV3 & 91.51\% $\pm$ 1.89 & 85.79\% $\pm$ 2.72 & 89.16\% $\pm$ 2.23 & \textbf{96.40\% $\pm$ 0.70} & 87.36\% $\pm$ 2.46 & 92.85\% $\pm$ 1.61 \\
GPT-4o Mini & 89.12\% $\pm$ 2.34 & 86.28\% $\pm$ 2.45 & 89.12\% $\pm$ 2.21 & 88.85\% $\pm$ 2.35 & \textbf{95.49\% $\pm$ 0.70} & 89.63\% $\pm$ 2.33 \\
GPT-4o & 91.73\% $\pm$ 1.90 & 84.95\% $\pm$ 2.73 & 89.56\% $\pm$ 2.22 & 92.57\% $\pm$ 1.65 & 88.03\% $\pm$ 2.43 & \textbf{96.75\% $\pm$ 0.61} \\
\bottomrule
\end{tabular}
}
\label{tab:mmlu_lime_world_history_filtered_fidelity}
\end{table*}

%% file: tables/mmlu_shap_fidelity.tex
\begin{table*}
\centering
\caption{Accuracy of proxy Kernel SHAP explanations on high school chemistriy of MMLU datasets: each value shows how well Kernel SHAP explanations generated by the model on the \textbf{left} serve as surrogates for predicting the behavior of the model on the \textbf{top}.}
\label{tab:mmlu_shap_chemistriy_fidelity}
\resizebox{\textwidth}{!}{
\begin{tabular}{lcccccc}
\toprule
 & Qwen 0.5B & Qwen 1.5B & Qwen 3B & Qwen 7B & Qwen 14B & Qwen 32B \\
\midrule
Qwen 0.5B & \textbf{79.57\% $\pm$ 2.29} & 57.58\% $\pm$ 3.83 & 55.11\% $\pm$ 4.00 & 52.69\% $\pm$ 3.94 & 52.11\% $\pm$ 3.89 & 51.86\% $\pm$ 4.08 \\
Qwen 1.5B & 56.67\% $\pm$ 3.75 & \textbf{76.57\% $\pm$ 2.48} & 65.22\% $\pm$ 3.40 & 61.13\% $\pm$ 3.58 & 61.17\% $\pm$ 3.44 & 60.88\% $\pm$ 3.62 \\
Qwen 3B & 53.17\% $\pm$ 4.02 & 63.62\% $\pm$ 3.28 & \textbf{76.13\% $\pm$ 2.43} & 64.01\% $\pm$ 3.36 & 65.89\% $\pm$ 3.19 & 65.00\% $\pm$ 3.51 \\
Qwen 7B & 50.78\% $\pm$ 4.08 & 59.90\% $\pm$ 3.71 & 64.21\% $\pm$ 3.38 & \textbf{75.91\% $\pm$ 2.62} & 68.60\% $\pm$ 2.87 & 65.86\% $\pm$ 3.49 \\
Qwen 14B & 52.47\% $\pm$ 3.86 & 60.29\% $\pm$ 3.51 & 62.05\% $\pm$ 3.49 & 66.96\% $\pm$ 3.20 & \textbf{72.97\% $\pm$ 2.73} & 69.70\% $\pm$ 3.09 \\
Qwen 32B & 51.49\% $\pm$ 4.08 & 59.04\% $\pm$ 3.87 & 61.37\% $\pm$ 3.68 & 66.05\% $\pm$ 3.30 & 67.24\% $\pm$ 3.25 & \textbf{78.10\% $\pm$ 2.47} \\
Qwen 72B & 51.61\% $\pm$ 4.14 & 59.96\% $\pm$ 3.83 & 62.77\% $\pm$ 3.57 & 67.94\% $\pm$ 3.28 & 67.42\% $\pm$ 3.23 & 71.90\% $\pm$ 3.21 \\
Llama 8B & 50.51\% $\pm$ 3.97 & 58.38\% $\pm$ 3.80 & 62.09\% $\pm$ 3.44 & 64.93\% $\pm$ 3.28 & 63.94\% $\pm$ 3.23 & 63.82\% $\pm$ 3.58 \\
Llama 70B & 52.08\% $\pm$ 3.96 & 59.40\% $\pm$ 3.81 & 62.38\% $\pm$ 3.68 & 65.09\% $\pm$ 3.41 & 66.48\% $\pm$ 3.31 & 71.06\% $\pm$ 2.96 \\
DeepSeekV3 & 52.15\% $\pm$ 4.19 & 59.74\% $\pm$ 3.90 & 64.14\% $\pm$ 3.43 & 66.82\% $\pm$ 3.33 & 67.48\% $\pm$ 3.20 & 72.30\% $\pm$ 3.19 \\
GPT-4o Mini & 51.72\% $\pm$ 4.16 & 57.42\% $\pm$ 3.92 & 62.46\% $\pm$ 3.57 & 64.06\% $\pm$ 3.52 & 67.18\% $\pm$ 3.27 & 67.59\% $\pm$ 3.40 \\
GPT-4o & 52.27\% $\pm$ 4.18 & 53.93\% $\pm$ 3.98 & 59.94\% $\pm$ 3.79 & 62.91\% $\pm$ 3.62 & 65.29\% $\pm$ 3.46 & 68.00\% $\pm$ 3.54 \\
\bottomrule
\end{tabular}
}

\quad
\resizebox{\textwidth}{!}{
\begin{tabular}{lcccccc}
\toprule
 & Qwen 72B & Llama 8B & Llama 70B & DeepSeekV3 & GPT-4o Mini & GPT-4o \\
\midrule
Qwen 0.5B & 52.37\% $\pm$ 4.14 & 50.76\% $\pm$ 3.66 & 52.13\% $\pm$ 3.91 & 49.19\% $\pm$ 4.10 & 53.86\% $\pm$ 3.98 & 51.17\% $\pm$ 4.32 \\
Qwen 1.5B & 60.18\% $\pm$ 3.82 & 59.53\% $\pm$ 3.49 & 60.26\% $\pm$ 3.66 & 59.45\% $\pm$ 3.63 & 61.10\% $\pm$ 3.63 & 55.91\% $\pm$ 4.23 \\
Qwen 3B & 62.91\% $\pm$ 3.69 & 61.13\% $\pm$ 3.42 & 64.79\% $\pm$ 3.36 & 62.55\% $\pm$ 3.59 & 62.76\% $\pm$ 3.48 & 58.50\% $\pm$ 4.30 \\
Qwen 7B & 67.59\% $\pm$ 3.45 & 62.75\% $\pm$ 3.30 & 66.82\% $\pm$ 3.26 & 67.06\% $\pm$ 3.32 & 62.73\% $\pm$ 3.58 & 55.86\% $\pm$ 4.45 \\
Qwen 14B & 69.30\% $\pm$ 3.31 & 60.49\% $\pm$ 3.47 & 67.95\% $\pm$ 3.17 & 68.56\% $\pm$ 3.20 & 65.43\% $\pm$ 3.31 & 60.26\% $\pm$ 4.17 \\
Qwen 32B & 71.60\% $\pm$ 3.26 & 61.20\% $\pm$ 3.37 & 69.65\% $\pm$ 3.18 & 72.82\% $\pm$ 2.94 & 66.38\% $\pm$ 3.51 & 63.95\% $\pm$ 4.15 \\
Qwen 72B & \textbf{79.57\% $\pm$ 2.54} & 60.77\% $\pm$ 3.58 & 70.85\% $\pm$ 3.22 & 72.52\% $\pm$ 3.09 & 67.61\% $\pm$ 3.48 & 64.93\% $\pm$ 4.24 \\
Llama 8B & 62.59\% $\pm$ 3.74 & \textbf{75.51\% $\pm$ 2.39} & 66.46\% $\pm$ 3.34 & 64.92\% $\pm$ 3.53 & 62.24\% $\pm$ 3.64 & 54.18\% $\pm$ 4.47 \\
Llama 70B & 71.09\% $\pm$ 3.28 & 62.62\% $\pm$ 3.44 & \textbf{77.52\% $\pm$ 2.58} & 71.37\% $\pm$ 3.03 & 66.51\% $\pm$ 3.57 & 63.95\% $\pm$ 4.28 \\
DeepSeekV3 & 70.80\% $\pm$ 3.51 & 61.44\% $\pm$ 3.55 & 69.22\% $\pm$ 3.35 & \textbf{78.82\% $\pm$ 2.52} & 67.41\% $\pm$ 3.42 & 65.96\% $\pm$ 4.14 \\
GPT-4o Mini & 70.99\% $\pm$ 3.38 & 57.86\% $\pm$ 3.68 & 67.77\% $\pm$ 3.43 & 70.57\% $\pm$ 3.27 & \textbf{70.62\% $\pm$ 3.26} & 68.27\% $\pm$ 3.72 \\
GPT-4o & 70.64\% $\pm$ 3.51 & 57.77\% $\pm$ 3.61 & 67.25\% $\pm$ 3.43 & 72.61\% $\pm$ 3.18 & 66.68\% $\pm$ 3.46 & \textbf{71.38\% $\pm$ 3.50 } \\
\bottomrule
\end{tabular}
}
\end{table*}

\begin{table*}
\centering
\caption{Accuracy of \textbf{filtered} proxy Kernel SHAP explanations on high school chemistriy of MMLU datasets: each value shows how well Kernel SHAP explanations generated by the model on the \textbf{left} serve as surrogates for predicting the behavior of the model on the \textbf{top}.}
\label{tab:mmlu_shap_chemistriy_filtered_fidelity}
\resizebox{\textwidth}{!}{
\begin{tabular}{lcccccc}
\toprule
 & Qwen 0.5B & Qwen 1.5B & Qwen 3B & Qwen 7B & Qwen 14B & Qwen 32B \\
\midrule
Qwen 0.5B & \textbf{79.57\% $\pm$ 2.29} & 67.42\% $\pm$ 4.66 & 68.75\% $\pm$ 5.00 & 64.16\% $\pm$ 5.44 & 67.35\% $\pm$ 6.09 & 72.14\% $\pm$ 5.57 \\
Qwen 1.5B & 70.51\% $\pm$ 4.43 & \textbf{76.57\% $\pm$ 2.48} & 72.57\% $\pm$ 3.55 & 70.46\% $\pm$ 4.17 & 69.40\% $\pm$ 4.06 & 71.97\% $\pm$ 4.24 \\
Qwen 3B & 70.46\% $\pm$ 5.07 & 69.16\% $\pm$ 4.09 & \textbf{76.13\% $\pm$ 2.43} & 70.79\% $\pm$ 3.82 & 72.63\% $\pm$ 3.75 & 74.54\% $\pm$ 3.63 \\
Qwen 7B & 65.90\% $\pm$ 5.45 & 67.53\% $\pm$ 4.51 & 70.18\% $\pm$ 3.63 & \textbf{75.91\% $\pm$ 2.62} & 73.07\% $\pm$ 3.32 & 74.09\% $\pm$ 3.54 \\
Qwen 14B & 70.50\% $\pm$ 5.14 & 69.12\% $\pm$ 3.96 & 68.63\% $\pm$ 4.02 & 72.42\% $\pm$ 3.56 & \textbf{72.97\% $\pm$ 2.73} & 73.58\% $\pm$ 3.15 \\
Qwen 32B & 69.56\% $\pm$ 5.99 & 68.68\% $\pm$ 4.63 & 68.40\% $\pm$ 4.41 & 72.28\% $\pm$ 3.74 & 71.45\% $\pm$ 3.55 & \textbf{78.10\% $\pm$ 2.47} \\
Qwen 72B & 69.70\% $\pm$ 5.71 & 70.44\% $\pm$ 4.73 & 70.33\% $\pm$ 4.11 & 74.52\% $\pm$ 3.21 & 73.21\% $\pm$ 3.53 & 75.92\% $\pm$ 3.11 \\
Llama 8B & 68.94\% $\pm$ 5.82 & 67.18\% $\pm$ 4.75 & 68.81\% $\pm$ 4.26 & 71.85\% $\pm$ 3.63 & 71.10\% $\pm$ 3.61 & 73.71\% $\pm$ 3.50 \\
Llama 70B & 71.21\% $\pm$ 5.34 & 69.64\% $\pm$ 4.36 & 71.16\% $\pm$ 4.09 & 72.81\% $\pm$ 3.65 & 71.75\% $\pm$ 3.61 & 75.07\% $\pm$ 2.93 \\
DeepSeekV3 & 71.98\% $\pm$ 5.65 & 69.87\% $\pm$ 4.53 & 72.20\% $\pm$ 3.88 & 73.35\% $\pm$ 3.59 & 73.42\% $\pm$ 3.57 & 75.62\% $\pm$ 3.13 \\
GPT-4o Mini & 71.25\% $\pm$ 5.45 & 67.38\% $\pm$ 4.84 & 68.52\% $\pm$ 3.98 & 71.75\% $\pm$ 4.04 & 71.94\% $\pm$ 3.97 & 74.18\% $\pm$ 3.69 \\
GPT-4o & 70.86\% $\pm$ 5.54 & 61.56\% $\pm$ 5.10 & 67.62\% $\pm$ 4.67 & 70.94\% $\pm$ 4.29 & 70.84\% $\pm$ 4.16 & 74.26\% $\pm$ 3.76 \\
\bottomrule
\end{tabular}
}

\quad
\resizebox{\textwidth}{!}{
\begin{tabular}{lcccccc}
\toprule
 & Qwen 72B & Llama 8B & Llama 70B & DeepSeekV3 & GPT-4o Mini & GPT-4o \\
\midrule
Qwen 0.5B & 71.04\% $\pm$ 5.32 & 65.76\% $\pm$ 5.67 & 71.43\% $\pm$ 5.32 & 67.91\% $\pm$ 6.00 & 67.45\% $\pm$ 5.58 & 73.38\% $\pm$ 5.06 \\
Qwen 1.5B & 73.40\% $\pm$ 4.23 & 70.67\% $\pm$ 4.05 & 72.48\% $\pm$ 4.16 & 71.46\% $\pm$ 4.12 & 70.20\% $\pm$ 4.24 & 70.77\% $\pm$ 4.27 \\
Qwen 3B & 73.17\% $\pm$ 3.83 & 69.04\% $\pm$ 3.95 & 72.75\% $\pm$ 3.81 & 72.87\% $\pm$ 3.98 & 68.48\% $\pm$ 3.89 & 72.95\% $\pm$ 3.87 \\
Qwen 7B & 74.22\% $\pm$ 3.52 & 69.55\% $\pm$ 3.78 & 72.96\% $\pm$ 3.69 & 73.94\% $\pm$ 3.69 & 69.22\% $\pm$ 4.08 & 72.63\% $\pm$ 3.72 \\
Qwen 14B & 74.86\% $\pm$ 3.27 & 69.10\% $\pm$ 3.68 & 73.18\% $\pm$ 3.38 & 74.73\% $\pm$ 3.37 & 71.11\% $\pm$ 3.62 & 72.94\% $\pm$ 3.48 \\
Qwen 32B & 76.50\% $\pm$ 3.07 & 69.62\% $\pm$ 3.64 & 73.22\% $\pm$ 3.28 & 76.26\% $\pm$ 2.95 & 72.94\% $\pm$ 3.44 & 75.47\% $\pm$ 3.15 \\
Qwen 72B & \textbf{79.57\% $\pm$ 2.54} & 69.68\% $\pm$ 3.66 & 74.78\% $\pm$ 3.36 & 76.62\% $\pm$ 3.06 & 71.35\% $\pm$ 3.71 & 76.18\% $\pm$ 3.36 \\
Llama 8B & 73.27\% $\pm$ 3.69 & \textbf{75.51\% $\pm$ 2.39} & 74.03\% $\pm$ 3.56 & 74.51\% $\pm$ 3.55 & 70.98\% $\pm$ 4.02 & 71.60\% $\pm$ 4.17 \\
Llama 70B & 76.37\% $\pm$ 3.22 & 71.22\% $\pm$ 3.52 & \textbf{77.52\% $\pm$ 2.58} & 75.51\% $\pm$ 3.05 & 72.90\% $\pm$ 3.51 & 76.19\% $\pm$ 3.28 \\
DeepSeekV3 & 75.76\% $\pm$ 3.54 & 70.13\% $\pm$ 3.62 & 73.24\% $\pm$ 3.45 & \textbf{78.82\% $\pm$ 2.52} & 73.35\% $\pm$ 3.62 & 75.97\% $\pm$ 3.22 \\
GPT-4o Mini & 76.20\% $\pm$ 3.44 & 68.34\% $\pm$ 4.09 & 73.39\% $\pm$ 3.71 & 76.63\% $\pm$ 3.48 & \textbf{70.62\% $\pm$ 3.26} & 76.11\% $\pm$ 3.36 \\
GPT-4o & 76.07\% $\pm$ 3.70 & 68.22\% $\pm$ 4.21 & 72.60\% $\pm$ 3.84 & 76.87\% $\pm$ 3.42 & 67.41\% $\pm$ 4.05 & \textbf{71.38\% $\pm$ 3.50 } \\
\bottomrule
\end{tabular}
}
\end{table*}

\begin{table*}
\centering
\caption{Accuracy of proxy Kernel SHAP explanations on high school computer science of MMLU datasets: each value shows how well Kernel SHAP explanations generated by the model on the \textbf{left} serve as surrogates for predicting the behavior of the model on the \textbf{top}.}
\label{tab:mmlu_shap_computer_science_fidelity}
\resizebox{\textwidth}{!}{
\begin{tabular}{lcccccc}
\toprule
 & Qwen 0.5B & Qwen 1.5B & Qwen 3B & Qwen 7B & Qwen 14B & Qwen 32B \\
\midrule
Qwen 0.5B & \textbf{78.94\% $\pm$ 3.20} & 62.47\% $\pm$ 5.17 & 61.97\% $\pm$ 4.79 & 60.71\% $\pm$ 5.16 & 57.75\% $\pm$ 5.15 & 59.20\% $\pm$ 5.11 \\
Qwen 1.5B & 61.14\% $\pm$ 5.59 & \textbf{75.69\% $\pm$ 3.63} & 67.03\% $\pm$ 4.79 & 69.18\% $\pm$ 4.81 & 66.45\% $\pm$ 4.79 & 64.11\% $\pm$ 5.07 \\
Qwen 3B & 61.63\% $\pm$ 5.22 & 66.26\% $\pm$ 4.49 & \textbf{72.73\% $\pm$ 3.81} & 70.34\% $\pm$ 4.33 & 69.87\% $\pm$ 4.28 & 70.31\% $\pm$ 4.45 \\
Qwen 7B & 59.00\% $\pm$ 5.40 & 64.09\% $\pm$ 4.69 & 66.83\% $\pm$ 4.46 & \textbf{73.20\% $\pm$ 3.74} & 70.19\% $\pm$ 4.11 & 71.18\% $\pm$ 3.85 \\
Qwen 14B & 59.04\% $\pm$ 5.47 & 64.16\% $\pm$ 4.84 & 69.60\% $\pm$ 4.20 & 73.69\% $\pm$ 3.97 & \textbf{74.34\% $\pm$ 4.09} & 75.84\% $\pm$ 3.90 \\
Qwen 32B & 52.86\% $\pm$ 3.93 & 57.94\% $\pm$ 3.56 & 59.09\% $\pm$ 3.25 & 61.35\% $\pm$ 3.16 & 62.02\% $\pm$ 3.14 & \textbf{62.40\% $\pm$ 3.19} \\
Qwen 72B & 58.06\% $\pm$ 5.49 & 62.79\% $\pm$ 5.17 & 67.77\% $\pm$ 4.36 & 72.77\% $\pm$ 3.81 & 71.57\% $\pm$ 4.26 & 75.13\% $\pm$ 3.71 \\
Llama 8B & 58.28\% $\pm$ 5.67 & 61.59\% $\pm$ 5.23 & 65.50\% $\pm$ 4.66 & 69.26\% $\pm$ 4.88 & 67.55\% $\pm$ 4.79 & 66.63\% $\pm$ 4.97 \\
Llama 70B & 59.01\% $\pm$ 5.86 & 63.51\% $\pm$ 5.09 & 66.50\% $\pm$ 4.82 & 72.90\% $\pm$ 4.03 & 70.47\% $\pm$ 4.80 & 74.90\% $\pm$ 3.85 \\
DeepSeekV3 & 57.81\% $\pm$ 5.77 & 63.22\% $\pm$ 5.06 & 65.30\% $\pm$ 4.63 & 71.31\% $\pm$ 4.37 & 70.73\% $\pm$ 4.64 & 72.04\% $\pm$ 4.52 \\
GPT-4o Mini & 61.37\% $\pm$ 5.48 & 64.37\% $\pm$ 5.01 & 65.37\% $\pm$ 4.69 & 69.66\% $\pm$ 4.45 & 69.46\% $\pm$ 4.59 & 71.12\% $\pm$ 4.35 \\
GPT-4o & 57.59\% $\pm$ 5.84 & 59.38\% $\pm$ 5.61 & 62.93\% $\pm$ 5.11 & 66.82\% $\pm$ 4.81 & 66.55\% $\pm$ 5.28 & 69.07\% $\pm$ 4.86 \\
\bottomrule
\end{tabular}
}

\quad
\resizebox{\textwidth}{!}{
\begin{tabular}{lcccccc}
\toprule
 & Qwen 72B & Llama 8B & Llama 70B & DeepSeekV3 & GPT-4o Mini & GPT-4o \\
\midrule
Qwen 0.5B & 58.02\% $\pm$ 5.28 & 61.61\% $\pm$ 4.81 & 59.14\% $\pm$ 5.15 & 59.99\% $\pm$ 5.43 & 58.67\% $\pm$ 5.54 & 56.69\% $\pm$ 5.79 \\
Qwen 1.5B & 64.27\% $\pm$ 5.10 & 65.30\% $\pm$ 4.85 & 66.07\% $\pm$ 5.04 & 64.62\% $\pm$ 5.30 & 63.27\% $\pm$ 5.14 & 62.28\% $\pm$ 5.79 \\
Qwen 3B & 69.72\% $\pm$ 4.47 & 66.63\% $\pm$ 4.25 & 70.84\% $\pm$ 4.17 & 68.22\% $\pm$ 4.68 & 68.25\% $\pm$ 4.33 & 66.74\% $\pm$ 4.94 \\
Qwen 7B & 70.50\% $\pm$ 4.10 & 66.03\% $\pm$ 4.51 & 70.67\% $\pm$ 3.91 & 69.92\% $\pm$ 4.22 & 67.35\% $\pm$ 4.24 & 67.61\% $\pm$ 4.51 \\
Qwen 14B & 74.06\% $\pm$ 4.09 & 68.61\% $\pm$ 4.54 & 73.41\% $\pm$ 4.03 & 72.76\% $\pm$ 4.58 & 70.20\% $\pm$ 4.68 & 70.67\% $\pm$ 5.02 \\
Qwen 32B & 61.44\% $\pm$ 3.21 & 59.41\% $\pm$ 3.39 & 59.63\% $\pm$ 3.34 & 60.41\% $\pm$ 3.30 & 58.72\% $\pm$ 3.43 & 59.49\% $\pm$ 3.61 \\
Qwen 72B & \textbf{75.60\% $\pm$ 3.57} & 66.92\% $\pm$ 4.47 & 74.79\% $\pm$ 3.85 & 72.75\% $\pm$ 4.19 & 67.75\% $\pm$ 4.81 & 71.22\% $\pm$ 4.47 \\
Llama 8B & 67.39\% $\pm$ 4.82 & \textbf{70.03\% $\pm$ 4.27} & 68.70\% $\pm$ 4.71 & 66.92\% $\pm$ 5.15 & 65.75\% $\pm$ 5.04 & 65.16\% $\pm$ 5.42 \\
Llama 70B & 72.76\% $\pm$ 4.30 & 67.10\% $\pm$ 4.77 & \textbf{76.91\% $\pm$ 3.65} & 71.63\% $\pm$ 4.53 & 66.94\% $\pm$ 4.91 & 67.30\% $\pm$ 5.31 \\
DeepSeekV3 & 70.78\% $\pm$ 4.60 & 66.43\% $\pm$ 4.68 & 72.68\% $\pm$ 4.19 & \textbf{75.02\% $\pm$ 4.14} & 67.05\% $\pm$ 5.06 & 68.70\% $\pm$ 4.94 \\
GPT-4o Mini & 70.65\% $\pm$ 4.48 & 63.73\% $\pm$ 4.97 & 70.58\% $\pm$ 4.52 & 71.24\% $\pm$ 4.72 & \textbf{70.29\% $\pm$ 4.94} & 73.27\% $\pm$ 4.72 \\
GPT-4o & 68.38\% $\pm$ 4.88 & 62.70\% $\pm$ 5.13 & 68.84\% $\pm$ 4.76 & 69.57\% $\pm$ 4.81 & 67.85\% $\pm$ 5.19 & \textbf{68.19\% $\pm$ 5.62 } \\
\bottomrule
\end{tabular}
}
\end{table*}

\begin{table*}
\centering
\caption{Accuracy of \textbf{filtered} proxy Kernel SHAP explanations on high school computer science of MMLU datasets: each value shows how well Kernel SHAP explanations generated by the model on the \textbf{left} serve as surrogates for predicting the behavior of the model on the \textbf{top}.}
\label{tab:mmlu_shap_computer_science_filtered_fidelity}
\resizebox{\textwidth}{!}{
\begin{tabular}{lcccccc}
\toprule
 & Qwen 0.5B & Qwen 1.5B & Qwen 3B & Qwen 7B & Qwen 14B & Qwen 32B \\
\midrule
Qwen 0.5B & \textbf{78.94\% $\pm$ 3.20} & 73.30\% $\pm$ 5.79 & 71.40\% $\pm$ 5.84 & 69.41\% $\pm$ 7.09 & 68.97\% $\pm$ 7.31 & 55.29\% $\pm$ 7.38 \\
Qwen 1.5B & 71.83\% $\pm$ 6.37 & \textbf{75.69\% $\pm$ 3.63} & 72.60\% $\pm$ 5.37 & 75.67\% $\pm$ 5.13 & 73.31\% $\pm$ 5.45 & 64.04\% $\pm$ 8.61 \\
Qwen 3B & 71.80\% $\pm$ 5.76 & 69.96\% $\pm$ 4.74 & \textbf{72.73\% $\pm$ 3.81} & 71.95\% $\pm$ 5.08 & 73.40\% $\pm$ 5.04 & 64.50\% $\pm$ 7.61 \\
Qwen 7B & 69.11\% $\pm$ 6.31 & 70.92\% $\pm$ 5.00 & 70.42\% $\pm$ 5.09 & \textbf{73.20\% $\pm$ 3.74} & 71.46\% $\pm$ 4.58 & 69.11\% $\pm$ 6.32 \\
Qwen 14B & 72.27\% $\pm$ 6.55 & 72.50\% $\pm$ 4.98 & 74.73\% $\pm$ 4.58 & 75.39\% $\pm$ 4.35 & \textbf{74.34\% $\pm$ 4.09} & 70.47\% $\pm$ 7.47 \\
Qwen 32B & 51.87\% $\pm$ 5.21 & 56.86\% $\pm$ 5.96 & 55.80\% $\pm$ 5.37 & 62.33\% $\pm$ 5.40 & 62.89\% $\pm$ 5.26 & \textbf{62.40\% $\pm$ 3.19} \\
Qwen 72B & 69.31\% $\pm$ 6.79 & 71.64\% $\pm$ 4.91 & 73.27\% $\pm$ 4.65 & 73.26\% $\pm$ 4.22 & 72.96\% $\pm$ 4.36 & 70.74\% $\pm$ 7.97 \\
Llama 8B & 70.88\% $\pm$ 7.60 & 68.47\% $\pm$ 5.71 & 69.63\% $\pm$ 5.30 & 71.86\% $\pm$ 5.62 & 71.08\% $\pm$ 5.52 & 70.90\% $\pm$ 9.64 \\
Llama 70B & 72.58\% $\pm$ 7.02 & 72.12\% $\pm$ 5.19 & 71.33\% $\pm$ 5.39 & 73.48\% $\pm$ 4.43 & 71.71\% $\pm$ 5.06 & 70.92\% $\pm$ 7.83 \\
DeepSeekV3 & 70.49\% $\pm$ 6.87 & 70.88\% $\pm$ 5.81 & 71.81\% $\pm$ 5.28 & 72.24\% $\pm$ 4.78 & 72.83\% $\pm$ 4.88 & 64.75\% $\pm$ 10.47 \\
GPT-4o Mini & 72.76\% $\pm$ 6.06 & 70.58\% $\pm$ 5.45 & 67.45\% $\pm$ 5.54 & 68.70\% $\pm$ 4.94 & 70.37\% $\pm$ 5.31 & 65.80\% $\pm$ 8.57 \\
GPT-4o & 66.78\% $\pm$ 7.81 & 67.19\% $\pm$ 6.41 & 67.48\% $\pm$ 5.96 & 66.75\% $\pm$ 5.38 & 68.45\% $\pm$ 5.80 & 63.02\% $\pm$ 10.25 \\
\bottomrule
\end{tabular}
}

\quad
\resizebox{\textwidth}{!}{
\begin{tabular}{lcccccc}
\toprule
 & Qwen 72B & Llama 8B & Llama 70B & DeepSeekV3 & GPT-4o Mini & GPT-4o \\
\midrule
Qwen 0.5B & 71.58\% $\pm$ 7.33 & 74.66\% $\pm$ 5.72 & 73.57\% $\pm$ 6.62 & 72.41\% $\pm$ 6.97 & 70.12\% $\pm$ 7.96 & 69.07\% $\pm$ 8.17 \\
Qwen 1.5B & 74.25\% $\pm$ 5.43 & 70.00\% $\pm$ 5.28 & 76.48\% $\pm$ 4.87 & 74.20\% $\pm$ 5.79 & 70.73\% $\pm$ 5.87 & 71.90\% $\pm$ 5.99 \\
Qwen 3B & 75.19\% $\pm$ 4.98 & 68.55\% $\pm$ 4.55 & 75.98\% $\pm$ 4.51 & 72.74\% $\pm$ 5.35 & 70.15\% $\pm$ 5.05 & 70.90\% $\pm$ 5.56 \\
Qwen 7B & 72.87\% $\pm$ 4.21 & 68.78\% $\pm$ 4.96 & 73.73\% $\pm$ 3.91 & 70.59\% $\pm$ 4.58 & 69.56\% $\pm$ 4.42 & 69.18\% $\pm$ 4.71 \\
Qwen 14B & 76.00\% $\pm$ 4.26 & 72.38\% $\pm$ 4.76 & 75.07\% $\pm$ 4.29 & 74.14\% $\pm$ 4.85 & 72.73\% $\pm$ 5.03 & 72.81\% $\pm$ 5.28 \\
Qwen 32B & 57.79\% $\pm$ 5.32 & 58.29\% $\pm$ 6.01 & 56.93\% $\pm$ 5.67 & 56.43\% $\pm$ 5.26 & 54.72\% $\pm$ 5.83 & 51.39\% $\pm$ 6.52 \\
Qwen 72B & \textbf{75.60\% $\pm$ 3.57} & 70.47\% $\pm$ 4.84 & 76.41\% $\pm$ 3.73 & 72.85\% $\pm$ 4.30 & 70.26\% $\pm$ 5.10 & 73.12\% $\pm$ 4.38 \\
Llama 8B & 73.38\% $\pm$ 5.18 & \textbf{70.03\% $\pm$ 4.27} & 74.60\% $\pm$ 4.82 & 73.29\% $\pm$ 5.45 & 71.20\% $\pm$ 5.49 & 72.36\% $\pm$ 5.78 \\
Llama 70B & 74.03\% $\pm$ 4.32 & 70.60\% $\pm$ 5.08 & \textbf{76.91\% $\pm$ 3.65} & 72.43\% $\pm$ 4.70 & 69.47\% $\pm$ 5.37 & 69.39\% $\pm$ 5.41 \\
DeepSeekV3 & 73.10\% $\pm$ 4.47 & 70.67\% $\pm$ 4.93 & 74.34\% $\pm$ 4.31 & \textbf{75.02\% $\pm$ 4.14} & 69.91\% $\pm$ 5.42 & 70.21\% $\pm$ 5.05 \\
GPT-4o Mini & 72.01\% $\pm$ 4.89 & 66.35\% $\pm$ 5.47 & 72.84\% $\pm$ 4.79 & 72.13\% $\pm$ 5.04 & \textbf{70.29\% $\pm$ 4.94} & 73.04\% $\pm$ 5.12 \\
GPT-4o & 70.42\% $\pm$ 5.11 & 64.97\% $\pm$ 6.24 & 71.37\% $\pm$ 4.94 & 69.83\% $\pm$ 5.19 & 69.41\% $\pm$ 5.72 & \textbf{68.19\% $\pm$ 5.62} \\
\bottomrule
\end{tabular}
}
\end{table*}

\begin{table*}
\centering
\caption{Accuracy of \textbf{filtered} proxy Kernel SHAP explanations on high school computer science of MMLU datasets: each value shows how well Kernel SHAP explanations generated by the model on the \textbf{left} serve as surrogates for predicting the behavior of the model on the \textbf{top}.}
\label{tab:mmlu_shap_computer_science_filtered_fidelity}
\resizebox{\textwidth}{!}{
\begin{tabular}{lcccccc}
\toprule
 & Qwen 0.5B & Qwen 1.5B & Qwen 3B & Qwen 7B & Qwen 14B & Qwen 32B \\
\midrule
Qwen 0.5B & \textbf{80.88\% $\pm$ 1.98} & 61.67\% $\pm$ 3.43 & 57.45\% $\pm$ 3.41 & 56.32\% $\pm$ 3.82 & 55.58\% $\pm$ 3.69 & 53.73\% $\pm$ 3.81 \\
Qwen 1.5B & 59.47\% $\pm$ 3.52 & \textbf{78.61\% $\pm$ 2.28} & 67.09\% $\pm$ 3.10 & 67.26\% $\pm$ 3.49 & 66.60\% $\pm$ 3.40 & 65.98\% $\pm$ 3.54 \\
Qwen 3B & 55.26\% $\pm$ 3.61 & 66.02\% $\pm$ 3.26 & \textbf{78.10\% $\pm$ 2.23} & 71.28\% $\pm$ 3.05 & 70.98\% $\pm$ 2.92 & 69.09\% $\pm$ 3.14 \\
Qwen 7B & 55.17\% $\pm$ 3.72 & 64.23\% $\pm$ 3.41 & 68.50\% $\pm$ 3.05 & \textbf{81.79\% $\pm$ 2.08} & 73.68\% $\pm$ 2.80 & 76.07\% $\pm$ 2.62 \\
Qwen 14B & 54.66\% $\pm$ 3.65 & 62.56\% $\pm$ 3.40 & 67.67\% $\pm$ 3.11 & 74.66\% $\pm$ 2.69 & \textbf{78.87\% $\pm$ 2.32} & 76.76\% $\pm$ 2.54 \\
Qwen 32B & 54.25\% $\pm$ 3.72 & 62.40\% $\pm$ 3.54 & 67.24\% $\pm$ 3.11 & 74.79\% $\pm$ 2.79 & 75.49\% $\pm$ 2.60 & \textbf{81.67\% $\pm$ 2.20} \\
Qwen 72B & 53.16\% $\pm$ 3.66 & 62.17\% $\pm$ 3.39 & 66.47\% $\pm$ 3.11 & 74.49\% $\pm$ 2.73 & 75.05\% $\pm$ 2.64 & 78.53\% $\pm$ 2.39 \\
Llama 8B & 57.76\% $\pm$ 3.53 & 65.09\% $\pm$ 3.13 & 67.47\% $\pm$ 3.08 & 70.17\% $\pm$ 3.25 & 70.40\% $\pm$ 3.11 & 69.81\% $\pm$ 3.24 \\
Llama 70B & 54.42\% $\pm$ 3.74 & 63.87\% $\pm$ 3.35 & 67.13\% $\pm$ 3.13 & 74.33\% $\pm$ 2.80 & 74.19\% $\pm$ 2.84 & 75.99\% $\pm$ 2.65 \\
DeepSeekV3 & 52.79\% $\pm$ 3.83 & 60.88\% $\pm$ 3.69 & 65.88\% $\pm$ 3.38 & 73.14\% $\pm$ 3.13 & 75.08\% $\pm$ 2.89 & 77.05\% $\pm$ 2.60 \\
GPT-4o Mini & 53.87\% $\pm$ 3.73 & 63.73\% $\pm$ 3.39 & 66.83\% $\pm$ 3.12 & 73.03\% $\pm$ 2.92 & 74.01\% $\pm$ 2.84 & 76.38\% $\pm$ 2.66 \\
GPT-4o & 54.03\% $\pm$ 3.75 & 61.93\% $\pm$ 3.45 & 65.49\% $\pm$ 3.25 & 74.17\% $\pm$ 2.77 & 74.12\% $\pm$ 2.84 & 76.34\% $\pm$ 2.73 \\
\bottomrule
\end{tabular}
}

\quad
\resizebox{\textwidth}{!}{
\begin{tabular}{lcccccc}
\toprule
 & Qwen 72B & Llama 8B & Llama 70B & DeepSeekV3 & GPT-4o Mini & GPT-4o \\
\midrule
Qwen 0.5B & 53.40\% $\pm$ 4.00 & 58.72\% $\pm$ 3.55 & 55.39\% $\pm$ 3.84 & 53.12\% $\pm$ 4.06 & 55.12\% $\pm$ 3.69 & 52.46\% $\pm$ 4.02 \\
Qwen 1.5B & 66.36\% $\pm$ 3.71 & 68.09\% $\pm$ 3.14 & 68.17\% $\pm$ 3.39 & 66.08\% $\pm$ 3.75 & 66.93\% $\pm$ 3.39 & 65.83\% $\pm$ 3.65 \\
Qwen 3B & 68.05\% $\pm$ 3.31 & 67.25\% $\pm$ 3.17 & 69.07\% $\pm$ 3.11 & 67.31\% $\pm$ 3.44 & 68.98\% $\pm$ 3.05 & 66.96\% $\pm$ 3.34 \\
Qwen 7B & 74.55\% $\pm$ 2.79 & 68.18\% $\pm$ 3.26 & 73.19\% $\pm$ 2.83 & 75.48\% $\pm$ 2.80 & 72.01\% $\pm$ 2.88 & 74.11\% $\pm$ 2.79 \\
Qwen 14B & 75.63\% $\pm$ 2.75 & 66.99\% $\pm$ 3.29 & 74.03\% $\pm$ 2.78 & 76.79\% $\pm$ 2.74 & 74.61\% $\pm$ 2.74 & 75.21\% $\pm$ 2.78 \\
Qwen 32B & 77.35\% $\pm$ 2.73 & 67.79\% $\pm$ 3.21 & 75.45\% $\pm$ 2.78 & 79.40\% $\pm$ 2.52 & 74.20\% $\pm$ 2.80 & 76.72\% $\pm$ 2.74 \\
Qwen 72B & \textbf{83.40\% $\pm$ 2.03} & 66.96\% $\pm$ 3.16 & 76.92\% $\pm$ 2.69 & 80.37\% $\pm$ 2.41 & 74.94\% $\pm$ 2.78 & 79.23\% $\pm$ 2.60 \\
Llama 8B & 69.87\% $\pm$ 3.30 & \textbf{79.76\% $\pm$ 2.22} & 71.90\% $\pm$ 3.08 & 70.37\% $\pm$ 3.42 & 69.59\% $\pm$ 3.23 & 70.53\% $\pm$ 3.33 \\
Llama 70B & 78.62\% $\pm$ 2.61 & 69.43\% $\pm$ 3.08 & \textbf{82.66\% $\pm$ 2.07} & 79.18\% $\pm$ 2.59 & 75.73\% $\pm$ 2.73 & 79.34\% $\pm$ 2.41 \\
DeepSeekV3 & 79.11\% $\pm$ 2.66 & 66.97\% $\pm$ 3.38 & 75.74\% $\pm$ 2.83 & \textbf{83.59\% $\pm$ 2.14} & 73.88\% $\pm$ 3.01 & 77.08\% $\pm$ 2.83 \\
GPT-4o Mini & 77.49\% $\pm$ 2.75 & 67.53\% $\pm$ 3.24 & 76.78\% $\pm$ 2.75 & 77.97\% $\pm$ 2.73 & \textbf{79.43\% $\pm$ 2.52} & 79.36\% $\pm$ 2.53 \\
GPT-4o & 78.36\% $\pm$ 2.74 & 67.54\% $\pm$ 3.14 & 76.75\% $\pm$ 2.66 & 78.82\% $\pm$ 2.70 & 75.37\% $\pm$ 2.78 & \textbf{82.00\% $\pm$ 2.28 } \\
\bottomrule
\end{tabular}
}
\caption{Accuracy of proxy Kernel SHAP explanations on high school microeconomics of MMLU datasets: each value shows how well Kernel SHAP explanations generated by the model on the \textbf{left} serve as surrogates for predicting the behavior of the model on the \textbf{top}.}
\label{tab:mmlu_shap_microeconomics_fidelity}
\end{table*}

\begin{table*}
\centering
\resizebox{\textwidth}{!}{
\begin{tabular}{lcccccc}
\toprule
 & Qwen 0.5B & Qwen 1.5B & Qwen 3B & Qwen 7B & Qwen 14B & Qwen 32B \\
\midrule
Qwen 0.5B & \textbf{80.88\% $\pm$ 1.98} & 71.51\% $\pm$ 3.83 & 70.03\% $\pm$ 3.99 & 76.19\% $\pm$ 3.64 & 72.03\% $\pm$ 4.20 & 73.52\% $\pm$ 3.93 \\
Qwen 1.5B & 74.25\% $\pm$ 3.36 & \textbf{78.61\% $\pm$ 2.28} & 73.88\% $\pm$ 3.15 & 77.99\% $\pm$ 2.82 & 76.05\% $\pm$ 2.98 & 76.95\% $\pm$ 2.92 \\
Qwen 3B & 71.52\% $\pm$ 3.73 & 73.73\% $\pm$ 2.98 & \textbf{78.10\% $\pm$ 2.23} & 76.84\% $\pm$ 2.73 & 76.58\% $\pm$ 2.71 & 76.12\% $\pm$ 2.89 \\
Qwen 7B & 73.92\% $\pm$ 3.80 & 74.14\% $\pm$ 2.88 & 73.50\% $\pm$ 3.05 & \textbf{81.79\% $\pm$ 2.08} & 75.55\% $\pm$ 2.89 & 78.63\% $\pm$ 2.53 \\
Qwen 14B & 71.25\% $\pm$ 4.05 & 71.35\% $\pm$ 3.12 & 73.77\% $\pm$ 3.05 & 77.18\% $\pm$ 2.59 & \textbf{78.87\% $\pm$ 2.32} & 78.78\% $\pm$ 2.49 \\
Qwen 32B & 72.83\% $\pm$ 3.91 & 71.94\% $\pm$ 3.23 & 73.06\% $\pm$ 3.15 & 77.82\% $\pm$ 2.60 & 76.94\% $\pm$ 2.69 & \textbf{81.67\% $\pm$ 2.20} \\
Qwen 72B & 69.76\% $\pm$ 4.10 & 71.24\% $\pm$ 3.30 & 71.84\% $\pm$ 3.19 & 76.97\% $\pm$ 2.63 & 76.58\% $\pm$ 2.63 & 79.73\% $\pm$ 2.36 \\
Llama 8B & 73.29\% $\pm$ 3.71 & 70.83\% $\pm$ 3.17 & 73.58\% $\pm$ 3.25 & 77.27\% $\pm$ 2.80 & 76.30\% $\pm$ 2.86 & 77.08\% $\pm$ 2.83 \\
Llama 70B & 72.80\% $\pm$ 4.00 & 72.09\% $\pm$ 3.18 & 72.05\% $\pm$ 3.20 & 77.04\% $\pm$ 2.60 & 76.71\% $\pm$ 2.69 & 77.80\% $\pm$ 2.54 \\
DeepSeekV3 & 72.04\% $\pm$ 4.14 & 71.02\% $\pm$ 3.52 & 72.49\% $\pm$ 3.37 & 76.66\% $\pm$ 2.95 & 77.72\% $\pm$ 2.72 & 78.28\% $\pm$ 2.62 \\
GPT-4o Mini & 71.23\% $\pm$ 4.08 & 72.52\% $\pm$ 3.26 & 71.57\% $\pm$ 3.24 & 76.25\% $\pm$ 2.72 & 76.56\% $\pm$ 2.78 & 78.77\% $\pm$ 2.53 \\
GPT-4o & 71.55\% $\pm$ 4.17 & 70.04\% $\pm$ 3.38 & 70.92\% $\pm$ 3.39 & 76.52\% $\pm$ 2.80 & 76.10\% $\pm$ 2.79 & 78.07\% $\pm$ 2.61 \\
\bottomrule
\end{tabular}
}

\quad
\resizebox{\textwidth}{!}{
\begin{tabular}{lcccccc}
\toprule
 & Qwen 72B & Llama 8B & Llama 70B & DeepSeekV3 & GPT-4o Mini & GPT-4o \\
\midrule
Qwen 0.5B & 74.58\% $\pm$ 3.94 & 71.47\% $\pm$ 3.96 & 74.22\% $\pm$ 3.98 & 75.53\% $\pm$ 4.09 & 70.43\% $\pm$ 4.34 & 74.36\% $\pm$ 4.17 \\
Qwen 1.5B & 78.92\% $\pm$ 2.95 & 74.80\% $\pm$ 2.91 & 78.25\% $\pm$ 2.90 & 79.15\% $\pm$ 2.79 & 76.23\% $\pm$ 3.13 & 77.51\% $\pm$ 2.93 \\
Qwen 3B & 76.03\% $\pm$ 2.77 & 75.34\% $\pm$ 2.82 & 75.43\% $\pm$ 2.83 & 76.00\% $\pm$ 2.93 & 73.94\% $\pm$ 2.94 & 74.93\% $\pm$ 2.87 \\
Qwen 7B & 77.57\% $\pm$ 2.55 & 74.54\% $\pm$ 3.00 & 76.10\% $\pm$ 2.73 & 78.93\% $\pm$ 2.54 & 74.64\% $\pm$ 2.76 & 77.81\% $\pm$ 2.50 \\
Qwen 14B & 78.32\% $\pm$ 2.48 & 73.55\% $\pm$ 3.06 & 76.81\% $\pm$ 2.64 & 79.37\% $\pm$ 2.53 & 76.87\% $\pm$ 2.63 & 78.06\% $\pm$ 2.50 \\
Qwen 32B & 79.46\% $\pm$ 2.52 & 73.53\% $\pm$ 3.12 & 77.78\% $\pm$ 2.62 & 80.74\% $\pm$ 2.46 & 76.27\% $\pm$ 2.80 & 78.95\% $\pm$ 2.50 \\
Qwen 72B & \textbf{83.40\% $\pm$ 2.03} & 72.58\% $\pm$ 3.05 & 79.72\% $\pm$ 2.43 & 81.16\% $\pm$ 2.36 & 77.04\% $\pm$ 2.60 & 81.20\% $\pm$ 2.28 \\
Llama 8B & 78.18\% $\pm$ 2.73 & \textbf{79.76\% $\pm$ 2.22} & 79.26\% $\pm$ 2.65 & 79.66\% $\pm$ 2.78 & 76.04\% $\pm$ 3.07 & 78.93\% $\pm$ 2.73 \\
Llama 70B & 81.90\% $\pm$ 2.15 & 75.09\% $\pm$ 2.88 & \textbf{82.66\% $\pm$ 2.07} & 81.63\% $\pm$ 2.29 & 77.56\% $\pm$ 2.67 & 81.42\% $\pm$ 2.17 \\
DeepSeekV3 & 80.79\% $\pm$ 2.46 & 73.96\% $\pm$ 3.22 & 78.81\% $\pm$ 2.57 & \textbf{83.59\% $\pm$ 2.14} & 76.16\% $\pm$ 2.89 & 78.81\% $\pm$ 2.57 \\
GPT-4o Mini & 80.57\% $\pm$ 2.39 & 74.43\% $\pm$ 3.02 & 79.33\% $\pm$ 2.52 & 80.78\% $\pm$ 2.41 & \textbf{79.43\% $\pm$ 2.52} & 81.61\% $\pm$ 2.29 \\
GPT-4o & 80.38\% $\pm$ 2.54 & 73.15\% $\pm$ 3.07 & 78.49\% $\pm$ 2.60 & 79.92\% $\pm$ 2.59 & 76.48\% $\pm$ 2.83 & \textbf{82.00\% $\pm$ 2.28} \\
\bottomrule
\end{tabular}
}
\caption{Accuracy of \textbf{filtered} proxy Kernel SHAP explanations on high school microeconomics of MMLU datasets: each value shows how well Kernel SHAP explanations generated by the model on the \textbf{left} serve as surrogates for predicting the behavior of the model on the \textbf{top}.}
\label{tab:mmlu_shap_microeconomics_filtered_fidelity}
\end{table*}

\begin{table*}
\centering
\resizebox{\textwidth}{!}{
\begin{tabular}{lcccccc}
\toprule
 & Qwen 0.5B & Qwen 1.5B & Qwen 3B & Qwen 7B & Qwen 14B & Qwen 32B \\
\midrule
Qwen 0.5B & \textbf{79.08\% $\pm$ 1.31} & 65.63\% $\pm$ 2.05 & 63.79\% $\pm$ 1.97 & 63.02\% $\pm$ 2.13 & 62.30\% $\pm$ 2.21 & 61.41\% $\pm$ 2.22 \\
Qwen 1.5B & 63.83\% $\pm$ 2.05 & \textbf{79.61\% $\pm$ 1.33} & 72.62\% $\pm$ 1.61 & 74.60\% $\pm$ 1.57 & 73.54\% $\pm$ 1.69 & 73.37\% $\pm$ 1.72 \\
Qwen 3B & 59.66\% $\pm$ 2.05 & 69.23\% $\pm$ 1.78 & \textbf{75.48\% $\pm$ 1.39} & 71.30\% $\pm$ 1.65 & 71.97\% $\pm$ 1.67 & 71.63\% $\pm$ 1.67 \\
Qwen 7B & 60.72\% $\pm$ 2.05 & 69.35\% $\pm$ 1.68 & 69.59\% $\pm$ 1.62 & \textbf{77.64\% $\pm$ 1.38} & 74.26\% $\pm$ 1.56 & 74.20\% $\pm$ 1.57 \\
Qwen 14B & 60.05\% $\pm$ 2.02 & 67.38\% $\pm$ 1.77 & 69.23\% $\pm$ 1.61 & 71.54\% $\pm$ 1.63 & \textbf{76.59\% $\pm$ 1.47} & 74.28\% $\pm$ 1.56 \\
Qwen 32B & 60.12\% $\pm$ 2.11 & 68.44\% $\pm$ 1.87 & 70.53\% $\pm$ 1.68 & 74.64\% $\pm$ 1.63 & 77.50\% $\pm$ 1.52 & \textbf{80.23\% $\pm$ 1.32} \\
Qwen 72B & 59.83\% $\pm$ 2.10 & 68.97\% $\pm$ 1.82 & 70.05\% $\pm$ 1.65 & 74.23\% $\pm$ 1.55 & 76.04\% $\pm$ 1.51 & 76.85\% $\pm$ 1.44 \\
Llama 8B & 60.67\% $\pm$ 2.12 & 70.39\% $\pm$ 1.84 & 71.71\% $\pm$ 1.66 & 74.68\% $\pm$ 1.68 & 75.20\% $\pm$ 1.66 & 76.18\% $\pm$ 1.54 \\
Llama 70B & 59.80\% $\pm$ 2.24 & 68.78\% $\pm$ 1.98 & 70.36\% $\pm$ 1.77 & 74.87\% $\pm$ 1.68 & 77.40\% $\pm$ 1.59 & 78.88\% $\pm$ 1.45 \\
DeepSeekV3 & 58.41\% $\pm$ 2.18 & 68.47\% $\pm$ 1.86 & 69.13\% $\pm$ 1.71 & 73.91\% $\pm$ 1.67 & 75.53\% $\pm$ 1.58 & 76.85\% $\pm$ 1.55 \\
GPT-4o Mini & 58.81\% $\pm$ 2.25 & 68.85\% $\pm$ 2.07 & 71.60\% $\pm$ 1.84 & 74.71\% $\pm$ 1.83 & 77.66\% $\pm$ 1.70 & 79.52\% $\pm$ 1.53 \\
GPT-4o & 58.61\% $\pm$ 2.30 & 67.57\% $\pm$ 2.18 & 71.37\% $\pm$ 1.85 & 75.03\% $\pm$ 1.83 & 77.62\% $\pm$ 1.70 & 79.46\% $\pm$ 1.60 \\
\bottomrule
\end{tabular}
}

\quad
\resizebox{\textwidth}{!}{
\begin{tabular}{lcccccc}
\toprule
 & Qwen 72B & Llama 8B & Llama 70B & DeepSeekV3 & GPT-4o Mini & GPT-4o \\
\midrule
Qwen 0.5B & 60.79\% $\pm$ 2.38 & 62.77\% $\pm$ 2.13 & 60.61\% $\pm$ 2.37 & 59.51\% $\pm$ 2.44 & 60.85\% $\pm$ 2.32 & 58.50\% $\pm$ 2.48 \\
Qwen 1.5B & 74.23\% $\pm$ 1.81 & 72.89\% $\pm$ 1.69 & 73.18\% $\pm$ 1.85 & 73.22\% $\pm$ 1.94 & 71.95\% $\pm$ 1.89 & 71.72\% $\pm$ 2.04 \\
Qwen 3B & 71.15\% $\pm$ 1.77 & 71.08\% $\pm$ 1.69 & 70.79\% $\pm$ 1.79 & 70.66\% $\pm$ 1.83 & 71.15\% $\pm$ 1.70 & 70.03\% $\pm$ 1.82 \\
Qwen 7B & 73.62\% $\pm$ 1.66 & 71.53\% $\pm$ 1.69 & 73.36\% $\pm$ 1.67 & 73.32\% $\pm$ 1.75 & 71.54\% $\pm$ 1.75 & 72.46\% $\pm$ 1.80 \\
Qwen 14B & 73.32\% $\pm$ 1.64 & 70.07\% $\pm$ 1.71 & 72.94\% $\pm$ 1.69 & 72.48\% $\pm$ 1.75 & 72.09\% $\pm$ 1.73 & 72.53\% $\pm$ 1.78 \\
Qwen 32B & 78.74\% $\pm$ 1.46 & 72.96\% $\pm$ 1.69 & 78.28\% $\pm$ 1.50 & 77.95\% $\pm$ 1.59 & 75.94\% $\pm$ 1.65 & 77.78\% $\pm$ 1.64 \\
Qwen 72B & \textbf{80.22\% $\pm$ 1.36} & 72.10\% $\pm$ 1.71 & 77.50\% $\pm$ 1.53 & 78.20\% $\pm$ 1.51 & 74.52\% $\pm$ 1.73 & 76.43\% $\pm$ 1.65 \\
Llama 8B & 75.64\% $\pm$ 1.71 & \textbf{80.42\% $\pm$ 1.30} & 76.43\% $\pm$ 1.65 & 75.65\% $\pm$ 1.74 & 74.70\% $\pm$ 1.75 & 74.73\% $\pm$ 1.85 \\
Llama 70B & 79.82\% $\pm$ 1.49 & 74.16\% $\pm$ 1.70 & \textbf{83.22\% $\pm$ 1.29} & 79.27\% $\pm$ 1.60 & 78.07\% $\pm$ 1.62 & 79.35\% $\pm$ 1.61 \\
DeepSeekV3 & 78.67\% $\pm$ 1.49 & 72.02\% $\pm$ 1.70 & 77.18\% $\pm$ 1.63 & \textbf{81.20\% $\pm$ 1.44} & 74.70\% $\pm$ 1.76 & 77.06\% $\pm$ 1.71 \\
GPT-4o Mini & 80.95\% $\pm$ 1.59 & 74.53\% $\pm$ 1.84 & 80.55\% $\pm$ 1.64 & 81.44\% $\pm$ 1.63 & \textbf{82.88\% $\pm$ 1.50} & 81.75\% $\pm$ 1.56 \\
GPT-4o & 81.79\% $\pm$ 1.54 & 73.65\% $\pm$ 1.92 & 80.76\% $\pm$ 1.69 & 82.81\% $\pm$ 1.51 & 79.56\% $\pm$ 1.67 & \textbf{84.90\% $\pm$ 1.39} \\
\bottomrule
\end{tabular}
}
\caption{Accuracy of proxy Kernel SHAP explanations on high school psychology of MMLU datasets: each value shows how well Kernel SHAP explanations generated by the model on the \textbf{left} serve as surrogates for predicting the behavior of the model on the \textbf{top}.}
\label{tab:mmlu_shap_psychology_fidelity}
\end{table*}

\begin{table*}
\centering
\resizebox{\textwidth}{!}{
%
}
\caption{Accuracy of \textbf{filtered} proxy Kernel SHAP explanations on high school psychology of MMLU datasets: each value shows how well Kernel SHAP explanations generated by the model on the \textbf{left} serve as surrogates for predicting the behavior of the model on the \textbf{top}.}
\label{tab:mmlu_shap_psychology_filtered_fidelity}
\end{table*}

\begin{table*}
\centering
\resizebox{\textwidth}{!}{
%
}
\caption{Accuracy of proxy Kernel SHAP explanations on high school world history of MMLU datasets: each value shows how well Kernel SHAP explanations generated by the model on the \textbf{left} serve as surrogates for predicting the behavior of the model on the \textbf{top}.}
\label{tab:mmlu_shap_world_history_fidelity}
\end{table*}

\begin{table*}
\centering
\resizebox{\textwidth}{!}{
%
}
\caption{Accuracy of \textbf{filtered} proxy Kernel SHAP explanations on high school world history of MMLU datasets: each value shows how well Kernel SHAP explanations generated by the model on the \textbf{left} serve as surrogates for predicting the behavior of the model on the \textbf{top}.}
\label{tab:mmlu_shap_world_history_filtered_fidelity}
\end{table*}